%% file: dp_Bayes_UAI_Arxiv.tex
\documentclass[]{article}
\usepackage{proceed2e}
\usepackage{times}

\title{On the Theory and Practice of \\Privacy-Preserving Bayesian Data Analysis}

\author{
{\bf James Foulds}\\
Calit2 \& CSE Department \\ UC San Diego\\
jfoulds@ucsd.edu
\And {\bf Joseph Geumlek}\\
 CSE Department \\ UC San Diego\\
 jgeumlek@cs.ucsd.edu
\And
{\bf Max Welling}\\
Informatics Institute \& QUVA Lab\\
 University of Amsterdam\\
m.welling@uva.nl
\And
{\bf Kamalika Chaudhuri}\\
CSE Department \\ UC San Diego\\
kamalika@cs.ucsd.edu
}

\usepackage{natbib}
\usepackage{amsmath}
\usepackage{amssymb}
\usepackage{amsfonts}
\usepackage{amsthm}
\usepackage{algorithm}
\usepackage{color}

\newtheorem{Thm}{Theorem}
\newtheorem{Lem}[Thm]{Lemma}
\newtheorem{Cor}[Thm]{Corollary}
\newtheorem{Prop}[Thm]{Proposition}

\newtheorem{theorem}{Theorem}
\newenvironment{proof2}{\noindent {\sc Proof:}}{$\Box$ \medskip} 

\newtheorem{Rem}{Remark}

\theoremstyle{definition}
\newtheorem{Defn}{Definition}
\theoremstyle{plain}

\def\argmax{\mathrm{argmax}}

\usepackage{mathtools}

\newif\ifarxiv 

\usepackage{enumitem}
\setlistdepth{9}

\usepackage{chngcntr}
\usepackage{apptools}
\AtAppendix{\counterwithin{theorem}{section}}

\begin{document}

\maketitle

\begin{abstract}
Bayesian inference has great promise for the privacy-preserving analysis of sensitive data, as posterior sampling automatically preserves differential privacy, an algorithmic notion of data privacy, under certain conditions \citep{dimitrakakis2014robust, wang2015privacy}.  While this \emph{one posterior sample} (OPS) approach elegantly provides privacy ``for free,'' it is data inefficient in the sense of asymptotic relative efficiency (ARE).  We show that a simple alternative based on the Laplace mechanism, the workhorse of differential privacy, is as asymptotically efficient as non-private posterior inference, under general assumptions.  This technique also has practical advantages including efficient use of the privacy budget for MCMC.  We demonstrate the practicality of our approach on a time-series analysis of sensitive military records from the Afghanistan and Iraq wars disclosed by the Wikileaks organization.  
\end{abstract}
\input{introduction.tex}

\input{background.tex}

\input{exponentialFamilies}

\input{exponentialFamiliesLaplace}

\input{laplaceTheory}

\input{privateMCMC}

\input{caseStudy}

\input{conclusion}

%
\subsubsection*{Acknowledgements}
The work of K. Chaudhuri and J. Geumlek was supported in part by NSF under IIS 1253942, and the work of M. Welling was supported in part by Qualcomm, Google and Facebook. We also thank Mijung Park, Eric Nalisnick, and Babak Shahbaba for helpful discussions.



\bibliographystyle{apalike}
\bibliography{references}

\newpage
\appendix

\setlist[itemize,1]{label=$ $}
\setlist[itemize,2]{label=$ $}
\setlist[itemize,3]{label=$ $}
\setlist[itemize,4]{label=$ $}
\setlist[itemize,5]{label=$ $}
\setlist[itemize,6]{label=$ $}
\setlist[itemize,7]{label=$ $}
\setlist[itemize,8]{label=$ $}
\setlist[itemize,9]{label=$ $}

\renewlist{itemize}{itemize}{9}

\twocolumn[
\hrule height4pt 

\vskip .25in
\begin{center}
\Large\bf Supplementary Material
\end{center}
\vskip .18in 

\hrule height1pt 

\vskip .25in

\vskip 1.377in
]

\input{supplementaryContent}

\end{document}

%% file: introduction.tex
\section{INTRODUCTION}
Probabilistic models trained via Bayesian inference are widely and successfully used in application domains where privacy is invaluable, from text analysis \citep{blei2003latent, goldwater2007fully}, to personalization \citep{salakhutdinov2008bayesian}, to medical informatics \citep{husmeier2006probabilistic}, to MOOCs \citep{piech2013tuning}.  In these applications, data scientists must carefully balance the benefits and potential insights from data analysis against the privacy concerns of the individuals whose data are being studied \citep{daries2014privacy}.

\citet{dwork2006calibrating} placed the notion of privacy-preserving data analysis on a solid foundation by introducing \emph{differential privacy} \citep{dwork2013algorithmic}, an algorithmic formulation of privacy which is a gold standard for privacy-preserving data-driven algorithms. Differential privacy measures the privacy ``cost'' of an algorithm.  When designing privacy-preserving methods, the goal is to achieve a good trade-off between privacy and utility, which ideally improves with the amount of available data.
 
As observed by \citet{dimitrakakis2014robust} and \citet{wang2015privacy}, Bayesian posterior sampling behaves synergistically with differential privacy because it automatically provides a degree of differential privacy under certain conditions.  However, there are substantial gaps between this elegant theory and the practical reality of Bayesian data analysis.  Privacy-preserving posterior sampling is hampered by data inefficiency, as measured by asymptotic relative efficiency (ARE).  In practice, it generally requires artificially selected constraints on the spaces of parameters as well as data points. Its privacy properties are also not typically guaranteed for approximate inference.

This paper identifies these gaps between theory and practice, and begins to mend them via an extremely simple alternative technique based on the workhorse of differential privacy, the Laplace mechanism \citep{dwork2006calibrating}.  Our approach is equivalent to a generalization of \citet{zhang2016differential}'s recently and independently proposed algorithm for beta-Bernoulli systems. We provide a theoretical analysis and empirical validation of the advantages of the proposed method.  We extend both our method and \cite{dimitrakakis2014robust,wang2015privacy}'s \emph{one posterior sample (OPS)} method to the case of approximate inference with privacy-preserving MCMC.  Finally, we demonstrate the practical applicability of this technique by showing how to use a privacy-preserving HMM model to analyze sensitive military records from the Iraq and Afghanistan wars leaked by the Wikileaks organization.  Our primary contributions are as follows:
\begin{itemize}
	\item We analyze the privacy cost of posterior sampling for exponential family posteriors via OPS.
	\item We explore a simple Laplace mechanism alternative to OPS for exponential families.
	\item Under weak conditions we establish the consistency of the Laplace mechanism approach and its data efficiency advantages over OPS.
	\item We extend the OPS and Laplace mechanism methods to approximate inference via MCMC.
	
	\item We demonstrate the practical implications with a case study on sensitive military records.
\end{itemize}

%% file: background.tex
\section{BACKGROUND}
We begin by discussing preliminaries on differential privacy and its application to Bayesian inference.  Our novel contributions will begin in Section \ref{subsec:expMechExpFam}.
\subsection{DIFFERENTIAL PRIVACY}
Differential privacy is a formal notion of the privacy of data-driven algorithms.  For an algorithm to be differentially private the probabilities of the outputs of the algorithms may not change much when one individual's data point is modified, thereby revealing little information about any one individual's data.  More precisely, a randomized algorithm $\mathcal{M}(\mathbf{X})$ is said to be $(\epsilon,\delta)$-differentially private if
\begin{equation}
\label{eqn:DP}
Pr(\mathcal{M}(\mathbf{X}) \in \mathcal{S}) \leq \exp(\epsilon) Pr(\mathcal{M}(\mathbf{X}') \in \mathcal{S}) + \delta
\end{equation}
for all measurable subsets $\mathcal{S}$ of the range of $\mathcal{M}$ and for all datasets $\mathbf{X}$, $\mathbf{X}'$ differing by a single entry \citep{dwork2013algorithmic}.  If $\delta = 0$, the algorithm is said to be $\epsilon$-differentially private.

\subsubsection{The Laplace Mechanism}
One straightforward method for obtaining $\epsilon$-differential privacy, known as the \emph{Laplace mechanism} \citep{dwork2006calibrating}, adds Laplace noise to the revealed information, where the amount of noise depends on $\epsilon$, and a quantifiable notion of the sensitivity to changes in the database.  Specifically, the $L1$ sensitivity $\triangle h$ for function $h$ is defined as
\begin{equation}
\triangle h = \max_{\mathbf{X}, \mathbf{X}^\prime}{\|h(\mathbf{X}) - h(\mathbf{X}^\prime)\|_1}
\end{equation}
for all datasets $\mathbf{X}$, $\mathbf{X}^\prime$ differing in at most one element.  The Laplace mechanism adds noise via
\begin{align}
\label{eqn:Laplace}
\mathcal{M}_L(\mathbf{X}, h, \epsilon) &= h(\mathbf{X}) + (Y_1, Y_2, \ldots, Y_d) \mbox{ ,}\\
Y_j &\sim \mbox{Laplace}(\triangle h/\epsilon), \forall j \in \{1, 2, \ldots, d\} \mbox{ ,} \nonumber
\end{align}
where $d$ is the dimensionality of the range of $h$.  The $\mathcal{M}_L(\mathbf{X}, h, \epsilon)$ mechanism is $\epsilon$-differentially private. 
\subsubsection{The Exponential Mechanism}
The exponential mechanism \citep{mcsherry2007mechanism} aims to output responses of high utility while maintaining privacy.  Given a utility function $u(\mathbf{X}, \mathbf{r})$ that maps database $\mathbf{X}$/output $\mathbf{r}$ pairs to a real-valued score, the exponential mechanism $\mathcal{M}_E(\mathbf{X}, u, \epsilon)$ produces random outputs via
\begin{equation}
Pr(\mathcal{M}_E(\mathbf{X}, u, \epsilon) = \mathbf{r}) \propto \exp \Big(\frac{\epsilon u(\mathbf{X}, \mathbf{r})}{2 \triangle u}\Big) \mbox{ ,}
\end{equation}
where the sensitivity of the utility function is
\begin{equation}
\triangle u \triangleq \max_{r, (\mathbf{X}^{(1)}, \mathbf{X}^{(2)})}\|u(\mathbf{X}^{(1)}, r) - u(\mathbf{X}^{(2)}, r) \|_1 \ \mbox{ ,} \label{eqn:expSens}
\end{equation}
in which $(\mathbf{X}^{(1)}, \mathbf{X}^{(2)})$ are pairs of databases that differ in only one element.
\subsubsection{Composition Theorems}
A key property of differential privacy is that it holds under composition, via an additive accumulation.
\begin{theorem}
If $\mathcal{M}_1$ is $(\epsilon_1, \delta_1)$-differentially private, and $\mathcal{M}_2$ is $(\epsilon_2, \delta_2)$-differentially private, then $\mathcal{M}_{1,2}(\mathbf{X}) = (\mathcal{M}_1(\mathbf{X}), \mathcal{M}_2(\mathbf{X}))$ is ($\epsilon_1 + \epsilon_2, \delta_1 + \delta_2)$-differentially private.
\end{theorem}
This allows us to view the total $\epsilon$ and $\delta$ of our procedure as a privacy ``budget'' that we spend across the operations of our analysis.  There also exists an ``advanced composition'' theorem which provides privacy guarantees in an adversarial adaptive scenario called $k$-fold composition, and also allows an analyst to trade an increased $\delta$ for a smaller $\epsilon$ in this scenario \citep{dwork2010boosting}.  Differential privacy is also immune to data-independent post-processing.

\subsection{PRIVACY AND BAYESIAN INFERENCE}

Suppose we would like a differentially private draw of parameters and latent variables of interest $\theta$ from the posterior $Pr(\theta|\mathbf{X})$, where $\mathbf{X} = \{\mathbf{x}_1, \ldots, \mathbf{x}_N\}$ is the private dataset.  We can accomplish this by interpreting posterior sampling as an instance of the exponential mechanism with utility function $u(\mathbf{X}, \theta) = \log Pr(\theta,\mathbf{X})$, i.e. the log joint probability of the chosen $\theta$ assignment and the dataset $\mathbf{X}$ \citep{wang2015privacy}.  We then draw $\theta$ via
\begin{equation}
\hspace{-0.4cm} f(\theta;\mathbf{X}, \epsilon) \propto \exp \Big(\frac{\epsilon \log Pr(\theta,\mathbf{X})}{2 \triangle \log Pr(\theta,\mathbf{X})}\Big) = {Pr(\theta,\mathbf{X})}^{\frac{\epsilon }{2 \triangle \log Pr(\theta,\mathbf{X})}} \label{eqn:expMechPosterior}
\end{equation}
\ifarxiv
\else
where the sensitivity is $\triangle \log Pr(\theta,\mathbf{X}) 
 \triangleq $
\begin{align}
\hspace{-0.25cm} \max_{\theta, (\mathbf{X}^{(1)}, \mathbf{X}^{(2)})}\|\log Pr(\theta,\mathbf{X}^{(1)}) - \log Pr(\theta,\mathbf{X}^{(2)}) \|_1 \ \label{eqn:expSensPosterior}
\end{align}
\fi
in which $\mathbf{X}^{(1)}$ and $\mathbf{X}^{(2)}$ differ in one element.
If the data points are conditionally independent given $\theta$, 
\begin{equation}
\log Pr(\theta, \mathbf{X}) = \log Pr(\theta) + \sum_{i=1}^N \log Pr(\mathbf{x}_i|\theta)\mbox{ ,}
\end{equation}
where $Pr(\theta)$ is the prior and $Pr(\mathbf{x}_i|\theta)$ is the likelihood term for data point $\mathbf{x}_i$.  Since the prior does not depend on the data, and each data point is associated with a single log-likelihood term $\log Pr(\mathbf{x}_i|\theta)$ in $\log Pr(\theta, \mathbf{X})$, from the above two equations we have 
\begin{align}
 \triangle \log Pr(\theta,\mathbf{X}) 
 = \max_{\mathbf{x},\mathbf{x}', \theta} |\log Pr(\mathbf{x}'|\theta) - \log Pr(\mathbf{x}|\theta)| \mbox{ .}  \label{eqn:sensPost}
\end{align}
This gives us the privacy cost of posterior sampling:
\begin{theorem}
If $\max_{\mathbf{x},\mathbf{x}' \in \chi, \theta \in \Theta} |\log Pr(\mathbf{x}'|\theta) - \log Pr(\mathbf{x}|\theta)| \leq C$, releasing one sample from the posterior distribution $Pr(\theta|\mathbf{X})$ with any prior is $2C$-differentially private. \label{cor:generalizedOPS}
\end{theorem}
\citet{wang2015privacy} derived this form of the result from first principles, while noting that the exponential mechanism can be used, as we do here.  Although they do not explicitly state the theorem, they implicitly use it to show two noteworthy special cases, referred to as the \emph{One Posterior Sample (OPS)} procedure.  We state the first of these cases:
\begin{theorem}
If $\max_{\mathbf{x} \in \chi, \theta \in \Theta} |\log Pr(\mathbf{x}|\theta)| \leq B$, releasing
one sample from the posterior distribution $Pr(\theta|\mathbf{X})$ with any prior is $4B$-differentially private.
\end{theorem}
This follows directly from Theorem \ref{cor:generalizedOPS}, since if $|\log Pr(\mathbf{x}|\theta)| \leq B$, $C = \triangle \log Pr(\theta,\mathbf{X})  = 2B$.

Under the exponential mechanism, $\epsilon$ provides an adjustable knob trading between privacy and fidelity.  When $\epsilon=0$, the procedure samples from a uniform distribution, giving away no information about $\mathbf{X}$.  When $\epsilon = 2 \triangle \log Pr(\theta,\mathbf{X})$, the procedure reduces to sampling $\theta$ from the posterior $Pr(\theta|\mathbf{X}) \propto Pr(\theta,\mathbf{X})$.  As $\epsilon$ approaches infinity the procedure becomes increasingly likely to sample the $\theta$ assignment with the highest posterior probability.  Assuming that our goal is to sample rather than to find a mode, we would cap $\epsilon$ at $2 \triangle \log Pr(\theta,\mathbf{X})$ in the above procedure in order to correctly sample from the true posterior.  More generally, if our privacy budget is $\epsilon'$, and $\epsilon' \geq 2q \triangle \log Pr(\theta,\mathbf{X})$, for integer $q$, we can draw $q$ posterior samples within our budget.

As observed by \citet{huang2012exponential}, the exponential mechanism can be understood via statistical mechanics.  We can write it as a Boltzmann distribution (a.k.a. a Gibbs measure)
\begin{align}
& f(\theta; \mathbf{x}, \epsilon) \propto \exp \Big (\frac{-E(\theta)}{T} \Big ) \mbox{ , } T = \frac{2 \triangle u(\mathbf{X},\theta)}{\epsilon} \mbox{ ,} \label{eqn:expMechTemperature}
\end{align}
where $E(\theta) = -u(\mathbf{X},\theta) = -\log Pr(\theta,\mathbf{X})$ is the energy of state $\theta$ in a physical system, and $T$ is the temperature of the system (in units such that Boltzmann's constant is one).  Reducing $\epsilon$ corresponds to increasing the temperature, which can be understood as altering the distribution such that a Markov chain moves through the state space more rapidly.

%% file: exponentialFamilies.tex
\section{PRIVACY FOR EXPONENTIAL FAMILIES: EXPONENTIAL VS LAPLACE}

\begin{table*}[t]
\begin{tabular}{|c|c|c|c|c|c|}
\hline
 Mechanism & Sensitivity & $ S(\mathbf{X})$ is & Release & ARE & Pay Gibbs cost \\
\hline
\hline
Laplace & $\sup_{\mathbf{X}, \mathbf{X}^\prime}\|\sum_{i=1}^N S(\mathbf{x}'^{(i)}) - \sum_{i=1}^N S(\mathbf{x}^{(i)})\|_1 $ & Noised & Statistics & $1$ & Once \\
\hline
Exponential & $\sup_{\mathbf{x},\mathbf{x}' \in \chi, \theta \in \Theta} | \theta^\intercal \Big ( S(\mathbf{x}') - S(\mathbf{x})\Big )$ & Rescaled & One & $1 + T$ & Per update \\ 
(OPS) & $\ \ \ \ \ \ \ \ \ \ \ \ \ \ \ \ \ \ + \log h(\mathbf{x}') - \log h(\mathbf{x})|$ &  & Sample & & (unless converged) \\
\hline
\end{tabular}
\caption{Comparison of the properties of the two methods for private Bayesian inference. \label{tab:compareMechanisms}}
\end{table*}

By analyzing the privacy cost of sampling from exponential family posteriors in the general case we can recover the privacy properties of many standard distributions.  These results can be applied to full posterior sampling, when feasible, or to Gibbs sampling updates, as we discuss in Section \ref{sec:privateMCMC}.  In this section we analyze the privacy cost of sampling from exponential family posterior distributions exactly (or at an appropriate temperature) via the exponential mechanism, following \citet{dimitrakakis2014robust} and \citet{wang2015privacy}, and via a method based on the Laplace mechanism, which is a generalization of \citet{zhang2016differential}.  The properties of the two methods are compared in Table \ref{tab:compareMechanisms}.
\subsection{THE EXPONENTIAL MECHANISM}
\label{subsec:expMechExpFam}
Consider exponential family models with likelihood
\begin{equation}
Pr(\mathbf{x}|\theta) = h(\mathbf{x}) g(\theta) \exp \Big (\theta^\intercal S(\mathbf{x}) \Big ) \mbox{ ,} \nonumber 
\end{equation}
where $S(\mathbf{x})$ is a vector of sufficient statistics for data point $\mathbf{x}$, and $\theta$ is a vector of natural parameters.
For $N$ i.i.d. data points, we have
\begin{equation}
Pr(\mathbf{X}|\theta) = \Big ( \prod_{i=1}^N h(\mathbf{x}^{(i)})\Big ) g(\theta)^N \exp \Big (\theta^\intercal \sum_{i=1}^N S(\mathbf{x}^{(i)}) \Big ) \mbox{ .} \nonumber 
\end{equation}
Further suppose that we have a conjugate prior which is also an exponential family distribution,
\begin{equation}
Pr(\theta| \chi, \alpha) =f(\chi, \alpha) g(\theta)^\alpha \exp \Big (\alpha \theta^\intercal \chi \Big ) \mbox{ ,} \nonumber 
\end{equation}
where $\alpha$ is a scalar, the number of prior ``pseudo-counts,'' and $\chi$ is a parameter vector.  The posterior is proportional to the prior times the likelihood,
\begin{equation}
Pr(\theta|\mathbf{X}, \chi, \alpha) \propto g(\theta)^{N + \alpha} \exp \Big (\theta^\intercal \big (\sum_{i=1}^N S(\mathbf{x}^{(i)}) + \alpha \chi \big) \Big ) \mbox{.} \label{eqn:posterior}
\end{equation}
\ifarxiv
\else
To compute the sensitivity of the posterior, we have
\begin{align}
|\log Pr&(\mathbf{x}'|\theta) - \log Pr(\mathbf{x}|\theta)|\\
 &= | \theta^\intercal \Big ( S(\mathbf{x}') - S(\mathbf{x})\Big ) + \log h(\mathbf{x}') - \log h(\mathbf{x})| \mbox{ .} \nonumber 
\end{align}
From Equation \ref{eqn:sensPost}, we obtain $ \triangle \log Pr(\theta, \mathbf{X}) 
 =$
\begin{align}
 \sup_{\mathbf{x},\mathbf{x}' \in \chi, \theta \in \Theta} | \theta^\intercal \Big ( S(\mathbf{x}') - S(\mathbf{x})\Big ) + \log h(\mathbf{x}') - \log h(\mathbf{x})| \mbox{ .} \label{eqn:expFamsens}
\end{align}
A posterior sample at temperature $T$,
\begin{align}
Pr_T(\theta|\mathbf{X}, \chi, \alpha) &\propto g(\theta)^{\frac{N + \alpha}{T}} \exp \Big (\theta^\intercal \frac{\sum_{i=1}^N S(\mathbf{x}^{(i)}) + \alpha \chi}{T}\Big ) \mbox{ , } \nonumber \\
 T &= \frac{2 \triangle \log p(\theta,X) }{\epsilon} \mbox{ ,} \label{eqn:posteriorTemperature}
\end{align}
has privacy cost $\epsilon$, by the exponential mechanism. As an example, consider a beta-Bernoulli model,
\begin{align} 
Pr(p& |\alpha, \beta) = \frac{1}{B(\alpha, \beta)} p^{\alpha-1}(1-p)^{\beta-1} \nonumber \\
= & \frac{1}{B(\alpha, \beta)} \exp (( \alpha-1 \big )\log p + ( \beta-1 \big )\log (1-p) ) \nonumber \\
\hspace{-0.35cm}
 Pr(x|p) &= p^x(1-p)^{1-x} = \exp ( x \log p+ (1-x) \log(1-p)  ) \nonumber
\end{align}
where $B(\alpha, \beta)$ is the beta function.  Given $N$ binary-valued data points $\mathbf{X} = x^{(1)}, \ldots, x^{(N)}$ from the Bernoulli distribution, the posterior is
\begin{align}
Pr&(p|\mathbf{X}, \alpha, \beta) \propto \nonumber \\
& \exp \Big ( \big (n_+ +  \alpha-1 \big ) \log p + \big (n_- +  \beta-1 \big ) \log (1-p)\Big ) \nonumber  \\
 &n_+ = \sum_{i=1}^N x^{(i)} , \ \ \ \ \ \ n_- = \sum_{i=1}^N (1-x^{(i)}) \nonumber \mbox{ .}
\end{align}
The sufficient statistics for each data point are $S(x) = [ x, 1 - x]^\intercal$.  The natural parameters for the likelihood are $\theta = [\log p , \log(1-p)]^\intercal$, and $h(x) = 0$.  The exponential mechanism sensitivity for a \emph{truncated} version of this model, where $a_0 \leq p \leq 1-a_0$, can be computed from Equation \ref{eqn:expFamsens}, $ \triangle \log Pr(\theta,\mathbf{X}) =$
\begin{align}
 \sup_{x,x' \in \{0,1\}, p \in [a_0,1-a_0]}&| x\log p + (1-x) \log( 1-p) \nonumber \\
 & - \big (x'\log p + (1-x') \log( 1-p) \big )| \nonumber \\
 &= -\log a_0 + \log(1-a_0) \label{eqn:expMechSensBeta} \mbox{ .} 
\end{align}
\fi
Note that if $a_0 = 0$, corresponding to a standard untruncated beta distribution, the sensitivity is unbounded.  This makes intuitive sense because some datasets are impossible if $p = 0$ or $p = 1$, which violates differential privacy.

%% file: exponentialFamiliesLaplace.tex
\subsection{THE LAPLACE MECHANISM}
\ifarxiv
\else
\begin{figure}[t]
\hspace{-0.55cm}
\includegraphics[scale=0.48, trim=1cm 7cm 1cm 7cm,clip]{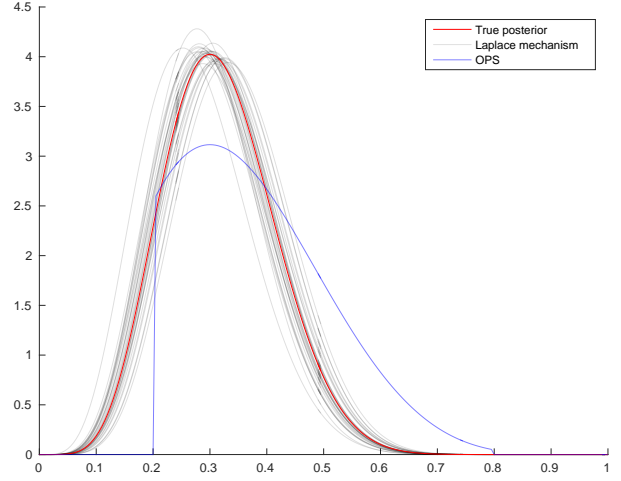}
\caption{Privacy-preserving approximate posteriors for a beta-Bernoulli model ($\epsilon = 1$, the true parameter $p = 0.3$, OPS truncation point $a_0 = 0.2$, and number of observations $N=20$).  For the Laplace mechanism, 30 privatizing draws are rendered. \label{fig:posteriors}}
\end{figure}
\fi

One limitation of the exponential mechanism / OPS approach to private Bayesian inference is that the temperature $T$ of the approximate posterior is fixed for any  $\epsilon$ that we are willing to pay, regardless of the number of data points $N$ (Equation \ref{eqn:expMechTemperature}).  While the posterior becomes more accurate as $N$ increases, and the OPS approximation becomes more accurate by proxy, the OPS approximation remains a factor of $T$ flatter than the posterior at $N$ data points.
%
This is not simply a limitation of the analysis. An adversary can choose data such that the dataset-specific privacy cost of posterior sampling approaches the worst case given by the exponential mechanism as $N$ increases, by causing the posterior to concentrate on the worst-case $\theta$ (see the supplement for an example). 

Here, we provide a simple Laplace mechanism alternative for exponential family posteriors, which becomes increasingly faithful to the true posterior with $N$ data points, as $N$ increases, for any fixed privacy cost $\epsilon$, under general assumptions.
The approach is based on the observation that for exponential family posteriors, as in Equation \ref{eqn:posterior}, the data interacts with the distribution only through the aggregate sufficient statistics, $S(\mathbf{X}) = \sum_{i=1}^N S(\mathbf{x}^{(i)})$.  If we release privatized versions of these statistics we can use them to perform any further operations that we'd like, including drawing samples, computing moments and quantiles, and so on.  This can straightforwardly be accomplished via the Laplace mechanism:
\begin{align}
 \hat{S}(\mathbf{X}) &= \mbox{proj}(S(\mathbf{X}) + (Y_1, Y_2, \ldots, Y_d)) \mbox{ ,} \label{eqn:laplaceSS}\\
 Y_j &\sim \mbox{Laplace}(\triangle S(\mathbf{X})/\epsilon), \forall j \in \{1, 2, \ldots, d\} \mbox{ ,} \nonumber
\end{align}
where $\mbox{proj}(\cdot)$ is a projection onto the space of sufficient statistics, if the Laplace noise takes it out of this region.  For example, if the statistics are counts, the projection ensures that they are non-negative.
The $L_1$ sensitivity of the aggregate statistics is
\begin{align}
\triangle S(\mathbf{X}) &= \sup_{\mathbf{X}, \mathbf{X}^\prime}{\|\sum_{i=1}^N S(\mathbf{x}'^{(i)}) - \sum_{i=1}^N S(\mathbf{x}^{(i)})\|_1} \label{L1sensExpFam}\\
&= \sup_{\mathbf{x}, \mathbf{x}^\prime}{\|S(\mathbf{x}') -  S(\mathbf{x})\|_1} \mbox{ ,} \nonumber 
\end{align}
where $\mathbf{X}$, $\mathbf{X}'$ differ in at most one element.  Note that perturbing the sufficient statistics is equivalent to perturbing the parameters, which was recently and independently proposed by \citet{zhang2016differential} for beta-Bernoulli models such as Bernoulli naive Bayes.

A comparison of Equations \ref{L1sensExpFam} and \ref{eqn:expFamsens} reveals that the L1 sensitivity and exponential mechanism sensitivities are closely related.  The L1 sensitivity is generally easier to control as it does not involve $\theta$ or $h(\mathbf{x})$ but otherwise involves similar terms to the exponential mechanism sensitivity.  For example, in the beta posterior case, where $S(\mathbf{x}) = [x, 1-x]$ is a binary indicator vector, the L1 sensitivity is 2.  This should be contrasted to the exponential mechanism sensitivity of Equation \ref{eqn:expMechSensBeta}, which depends heavily on the truncation point, and is unbounded for a standard untruncated beta distribution.  The L1 sensitivity is fixed regardless of the number of data points $N$, and so the amount of Laplace noise to add becomes smaller relative to the total $S(\mathbf{X})$ as $N$ increases.

\ifarxiv
\else
\begin{figure}[t]
\hspace{-0.2cm}\includegraphics[scale=0.65, trim=4cm 8.5cm 4cm 8.5cm,clip]{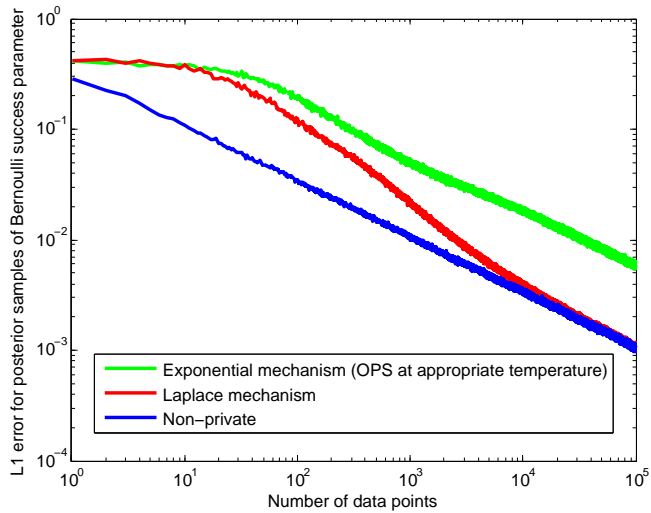}
\caption{ L1 error for private approximate samples from a beta posterior over a Bernoulli success parameter $p$, as a function of the number of $\mbox{Bernoulli}(p)$ observations, averaged over 1000 repeats.  The true parameter was $p=0.1$, the exponential mechanism posterior was truncated at $a_0 = 0.05$, and $\epsilon = 0.1$. \label{fig:betaL1Error}}
\end{figure}
\fi
Figure \ref{fig:posteriors} illustrates the differences in behavior between the two privacy-preserving Bayesian inference algorithms for a beta distribution posterior with Bernoulli observations.  The OPS estimator requires the distribution be truncated, here at $a_0 = 0.2$.  This controls the exponential mechanism sensitivity, which determines the temperature $T$ of the distribution, i.e. the extent to which the distribution is flattened, for a given $\epsilon$.  Here, $T = 2.7$.  In contrast, the Laplace mechanism achieves privacy by adding noise to the sufficient statistics, which in this case are the pseudo-counts of successes and failures for the posterior distribution.
In Figure \ref{fig:betaL1Error} we illustrate the fidelity benefits of posterior sampling based on the Laplace mechanism instead of the exponential mechanism as the amount of data increases.  In this case the exponential mechanism performs better than the Laplace mechanism only when the number of data points is very small (approximately $N=10$), and is quickly overtaken by the Laplace mechanism sampling procedure.  As $N$ increases the accuracy of sampling from the Laplace mechanism's approximate posterior converges to the performance of samples from the true posterior at the current number of observations $N$, while the exponential mechanism behaves similarly to the posterior with fewer than $N$ observations.  We show this formally in the next subsection.

%% file: laplaceTheory.tex
\subsection{THEORETICAL RESULTS}
\label{sec:theoreticalResults}
First, we show that the Laplace mechanism approximation of exponential family posteriors approaches the true posterior distribution \emph{evaluated at $N$ data points}.  Proofs are given in the supplementary.

\begin{Lem}
\label{lem:laplacekl}
For a minimal exponential family given a conjugate prior, where the posterior takes the form $Pr(\theta|\mathbf{X}, \chi, \alpha) \propto g(\theta)^{n + \alpha} \exp \Big (\theta^\intercal \big (\sum_{i=1}^n S(\mathbf{x}^{(i)}) + \alpha \chi \big) \Big ) $, where $p(\theta|\eta)$ denotes this posterior with a natural parameter vector $\eta$, if there exists a $\delta > 0$ such that these assumptions are met:

\begin{enumerate}
\item The data $\mathbf{X}$ comes i.i.d. from a minimal exponential family distribution with natural parameter $\theta_0 \in \Theta$
\item $\theta_0$ is in the interior of $\Theta$ 
\item The function $A(\theta)$ has all derivatives for $\theta$ in the interior of $\Theta$
\item $cov_{Pr(\mathbf{x}|\theta)}(S(\mathbf{x})))$ is finite for $\theta \in \mathcal{B}(\theta_0,\delta)$
\item $\exists w > 0$ s.t. $\det(cov_{Pr(\mathbf{x}|\theta)}(S(\mathbf{x})))) > w$ for $\theta \in \mathcal{B}(\theta_0,\delta)$
\item The prior $Pr(\theta|\chi,\alpha)$ is integrable and has support on a neighborhood of $\theta^*$
\end{enumerate}

then for any mechanism generating a perturbed posterior $\tilde{p}_N = p(\theta|\eta_N + \gamma)$ against a noiseless posterior $p_N = p(\theta|\eta_N)$ where $\gamma$ comes from a distribution that does not depend on the number of data observations $N$ and has finite covariance, this limit holds:

$\lim_{N \rightarrow \infty} E [ KL(\tilde{p}_N || p_N) ] = 0 \mbox { . }$

\end{Lem}

\begin{Cor}
The Laplace mechanism on an exponential family satisfies the noise distribution requirements of 
Lemma \ref{lem:laplacekl} when the sensitivity of the sufficient statistics is finite and either the exponential family is minimal, or if the exponential family parameters $\theta$ are identifiable. \label{cor:laplacemechKL}
\end{Cor}

These assumptions correspond to the data coming from a distribution where the Laplace regularity assumptions hold and the posterior satisfies the asymptotic normality given by the Bernstein-von Mises theorem. For example, in the beta-Bernoulli setting, these assumptions hold as long as the success parameter $p$ is in the open interval $(0,1)$. For $p=0$ or $1$, the relevant parameter is not in the interior of $\Theta$, and the result does not apply. In the setting of learning a normal distribution's mean $\mu$ where the variance $\sigma^2 > 0$ is known, the assumptions of Lemma \ref{lem:laplacekl} always hold, as the natural parameter space is an open set. However, Corollary \ref{cor:laplacemechKL} does not apply in this setting because the sensitivity is infinite (unless bounds are placed on the data).  Our efficiency result, in Theorem \ref{thm:laplaceare}, follows from Lemma \ref{lem:laplacekl} and the Bernstein-von Mises theorem.

\begin{theorem}
\label{thm:laplaceare}
Under the assumptions of Lemma \ref{lem:laplacekl}, the Laplace mechanism has an asymptotic posterior of $\mathcal{N}(\theta_0,2\mathbb{I}^{-1}/N)$ from which drawing a single sample has an asymptotic relative efficiency of 2 in estimating $\theta_0$, where $\mathbb{I}$ is the Fisher information at $\theta_0$.
\end{theorem}

Above, the asymptotic posterior refers to the normal distribution, whose variance depends on $N$, that the posterior distribution approaches as $N$ increases. This ARE result should be contrasted to that of the exponential mechanism \citep{wang2015privacy}.

\begin{theorem}
\label{thm:exponetialare}
The exponential mechanism applied to the exponential family with temperature parameter $T \geq 1$ has an asymptotic posterior of $\mathcal{N}(\theta^*,(1+T)\mathbb{I}^{-1}/N)$ and a single sample has an asymptotic relative efficiency of $(1+T)$ in estimating $\theta^*$,  where $\mathbb{I}$ is the Fisher information at $\theta^*$.
\end{theorem}

Here, the ARE represents the ratio between the variance of the estimator and the optimal variance $\mathbb{I}^{-1}/N$ achieved by the posterior mean in the limit.
Sampling from the posterior itself has an ARE of 2, due to the stochasticity of sampling, which the Laplace mechanism approach matches.
These theoretical results provide an explanation for the difference in the behavior of these two methods as $N$ increases seen in Figure \ref{fig:betaL1Error}. The Laplace mechanism will eventually approach the true posterior and the impact of privacy on accuracy will diminish when the data size increases. However, for the exponential mechanism with $T > 1$, the ratio of variances between the sampled posterior and the true posterior given $N$ data points approaches $(1+T)/2$, making the sampled posterior more spread out than the true posterior even as $N$ grows large.

So far we have compared the ARE values for \emph{sampling}, as an apples-to-apples comparison.  In reality, the Laplace mechanism has a further advantage as it releases a full posterior with privatized parameters, while the exponential mechanism can only release a finite number of samples with a finite $\epsilon$, which we discuss in Remark \ref{rem:laplacefullare}. 

\begin{Rem}
\label{rem:laplacefullare}
Under the the assumptions of Lemma \ref{lem:laplacekl}, by using the full privatized posterior instead of just a sample from it, the Laplace mechanism can release the privatized posterior's mean, which has an asymptotic relative efficiency of 1 in estimating $\theta^*$.
\end{Rem}

%% file: privateMCMC.tex
\section{PRIVATE GIBBS SAMPLING}
\label{sec:privateMCMC}
We now shift our discussion to the case of approximate Bayesian inference.  While the analysis of \citet{dimitrakakis2014robust} and \citet{wang2015privacy} shows that posterior sampling is differentially private under certain conditions, exact sampling is not in general tractable.  It does not directly follow that approximate sampling algorithms such as MCMC are also differentially private, or private at the same privacy level.  \citet{wang2015privacy} give two results towards understanding the privacy properties of approximate sampling algorithms.  First, they show that if the approximate sampler is ``close'' to the true distribution in a certain sense, then the privacy cost will be close to that of a true posterior sample:
\begin{Prop}
If procedure $\mathcal{A}$ which produces samples from distribution $P_\mathbf{X}$ is $\epsilon$-differentially private, then any approximate sampling procedures $\mathcal{A}^\prime$ that produces a sample from $P^\prime_\mathbf{X}$ such that $\| P_\mathbf{X} - P^\prime_\mathbf{X} \|_1 \leq \delta$ for any $\mathbf{X}$ is $(\epsilon, (1 + \exp(\epsilon)\delta)$-differentially private.
\end{Prop}
Unfortunately, it is not in general feasible to verify the convergence of an MCMC algorithm, and so this criterion is not generally verifiable in practice.  In their second result, Wang et al. study the privacy properties of stochastic gradient MCMC algorithms, including stochastic gradient Langevin dynamics  (SGLD) \citep{welling2011bayesian} and its extensions.  SGLD is a stochastic gradient method with noise injected in the gradient updates which converges in distribution to the target posterior.

In this section we study the privacy cost of MCMC, allowing us to quantify the privacy of many real-world MCMC-based Bayesian analyses.  We focus on the case of Gibbs sampling, under exponential mechanism and Laplace mechanism approaches.  By reinterpreting Gibbs sampling as an instance of the exponential mechanism, we obtain the ``privacy for free'' cost of Gibbs sampling. Metropolis-Hastings and annealed importance sampling also have privacy guarantees, which we show in the supplementary materials. 

\subsection{EXPONENTIAL MECHANISM}
\label{sec:GibbsExpMech}
We consider the privacy cost of a Gibbs sampler, where data $\mathbf{X}$ are behind the privacy wall, current sampled values of parameters and latent variables $\theta = [\theta_1, \ldots, \theta_D]$ are publicly known, and a Gibbs update is a randomized algorithm which queries our private data in order to randomly select a new value $\theta'_l$ for the current variable $\theta_l$. 
The transition kernel for a Gibbs update of $\theta_l$ is
\begin{equation}
T^{(Gibbs,l)}(\theta, \theta') = Pr(\theta'_l\big| \theta_{\neg l}, \mathbf{X}) \mbox{ ,} \label{eqn:Gibbs}
\end{equation}
where $\theta_{\neg l}$ refers to all entries of $\theta$ except $l$, which are held fixed, i.e. $\theta'_{\neg l} = \theta_{\neg l}$.  This update can be understood via the exponential mechanism:
\begin{equation}
T^{(Gibbs,l, \epsilon)}(\theta, \theta') \propto  Pr(\theta'_l, \theta_{\neg l}, \mathbf{X})^{\frac{\epsilon}{2 \triangle \log Pr(\theta'_l,\theta_{\neg l},\mathbf{X})}} \label{eqn:expMechGibbs} \mbox{ ,}
\end{equation}
with utility function $u(\mathbf{X},\theta'_l;\theta_{\neg l}) = \log Pr(\theta'_l,\theta_{\neg l},\mathbf{X})$, over the space of possible assignments to $\theta_l$, holding $\theta_{\neg l}$ fixed. A Gibbs update is therefore $\epsilon$-differentially private, with $\epsilon = 2 \triangle \log Pr(\theta'_l,\theta_{\neg l},\mathbf{X})$.
This update corresponds to Equation \ref{eqn:expMechPosterior} except that the set of responses for the exponential mechanism is restricted to those where $\theta'_{\neg l} = \theta_{\neg l}$.
Note that
\begin{equation}
\triangle \log Pr(\theta'_l,\theta_{\neg l},\mathbf{X}) \leq \triangle \log Pr(\theta,\mathbf{X})
\end{equation}
as the worst case is computed over a strictly smaller set of outcomes.
In many cases each parameter and latent variable $\theta_l$ is associated with only the $\l$th data point $\mathbf{x}_l$, in which case the privacy cost of a Gibbs scan can be improved over simple additive composition.
In this case a random sequence scan Gibbs pass, which updates all $N$ $\theta_l$'s exactly once, is $2\triangle \log Pr(\theta,\mathbf{X})$-differentially private by parallel composition \citep{song2013stochastic}.  Alternatively, a random scan Gibbs sampler, which updates a random $Q$ out of $N$ $\theta_l$'s,  is $4\triangle \log Pr(\theta,\mathbf{X})\frac{Q}{N} $-differentially private from the \emph{privacy amplification} benefit of subsampling data \citep{li2012sampling}.

\subsection{LAPLACE MECHANISM}
Suppose that the conditional posterior distribution for a Gibbs update is in the exponential family.  Having privatized the sufficient statistics arising from the data for the likelihoods involved in each update, via Equation \ref{eqn:laplaceSS}, and publicly released them with privacy cost $\epsilon$, we may now perform the update by drawing a sample from the approximate conditional posterior, i.e. Equation \ref{eqn:posterior} but with $S(\mathbf{X}) = \sum_{i=1}^N(\mathbf{x}^{(i)})$ replaced by $\hat{S}(\mathbf{X})$.  Since the privatized statistics can be made public, we can also subsequently draw from an approximate posterior based on $\hat{S}(\mathbf{X})$ with any other prior (selected based on public information only), without paying any further privacy cost.  This is especially valuable in a Gibbs sampling context, where the ``prior'' for a Gibbs update often consists of factors from other variables and parameters to be sampled, which are updated during the course of the algorithm.

In particular, consider a Bayesian model where a Gibbs sampler interacts with data only via conditional posteriors and their corresponding likelihoods that are exponential family distributions.  We can privatize the sufficient statistics of the likelihood just once at the beginning of the MCMC algorithm via the Laplace mechanism with privacy cost $\epsilon$, and then approximately sample from the posterior by running the entire MCMC algorithm based on these privatized statistics without paying any further privacy cost.  This is typically much cheaper in the privacy budget than exponential mechanism MCMC which pays a privacy cost for every Gibbs update, as we shall see in our case study in Section \ref{sec:caseStudy}.  The MCMC algorithm does not need to converge to obtain privacy guarantees, unlike the OPS method.  This approach applies to a very broad class of models, including Bayesian parameter learning for fully-observed MRF and Bayesian network models.  Of course, for this technique to be useful in practice, the aggregate sufficient statistics for each Gibbs update must be large relative to the Laplace noise.  For latent variable models, this typically corresponds to a setting with many data points per latent variable, such as the HMM model with multiple emissions per timestep which we study in the next section.

%% file: caseStudy.tex
\section{CASE STUDY: WIKILEAKS IRAQ \& AFGHANISTAN WAR LOGS}

\ifarxiv
\else
\begin{figure}[t]
\hspace{-0.37cm}
\includegraphics[scale=0.23]{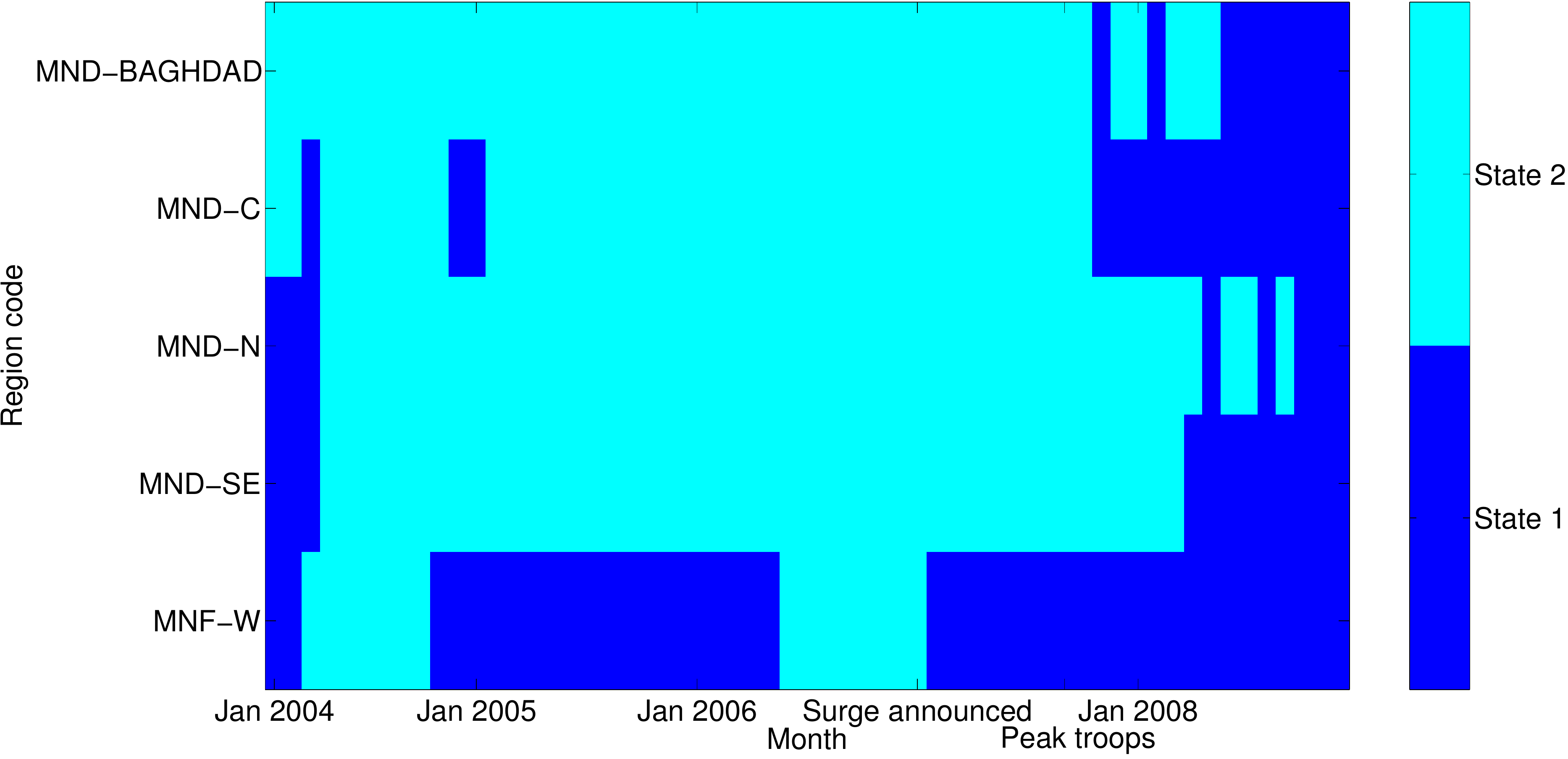}
\caption{State assignments of privacy-preserving HMM on Iraq (Laplace mechanism, $\epsilon = 5$). \label{fig:clusterAssignmentsIraq}}
\end{figure}
\fi

\ifarxiv
\else
\begin{figure}[t]
\hspace{-0.35cm} \includegraphics[scale=0.27]{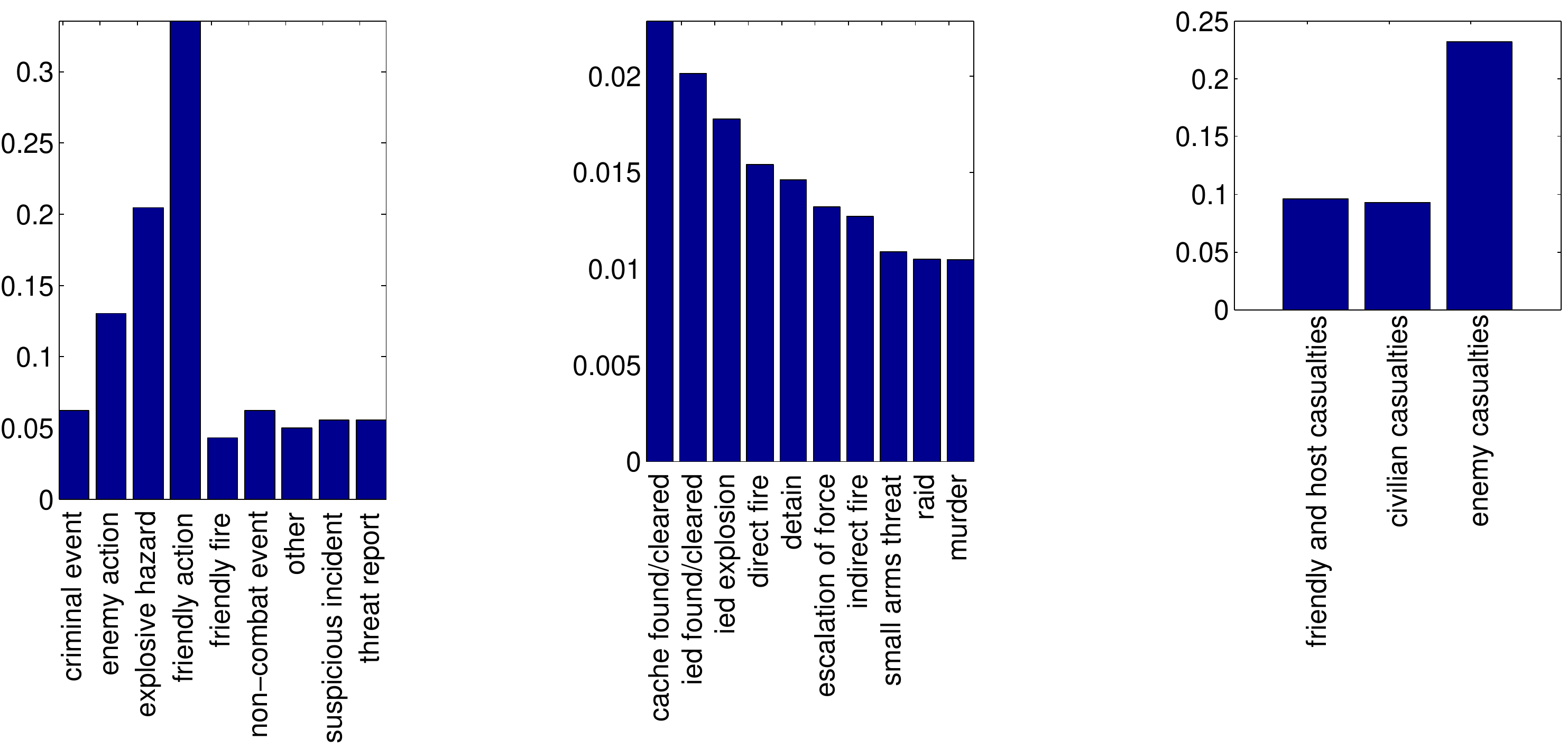}
\caption{State 1 for Iraq (\emph{type}, \emph{category}, \emph{casualties}). \label{fig:cluster1Iraq}}
\end{figure}
\fi

\ifarxiv
\else
\begin{figure}[t]
\hspace{-0.2cm}
\includegraphics[scale=0.27]{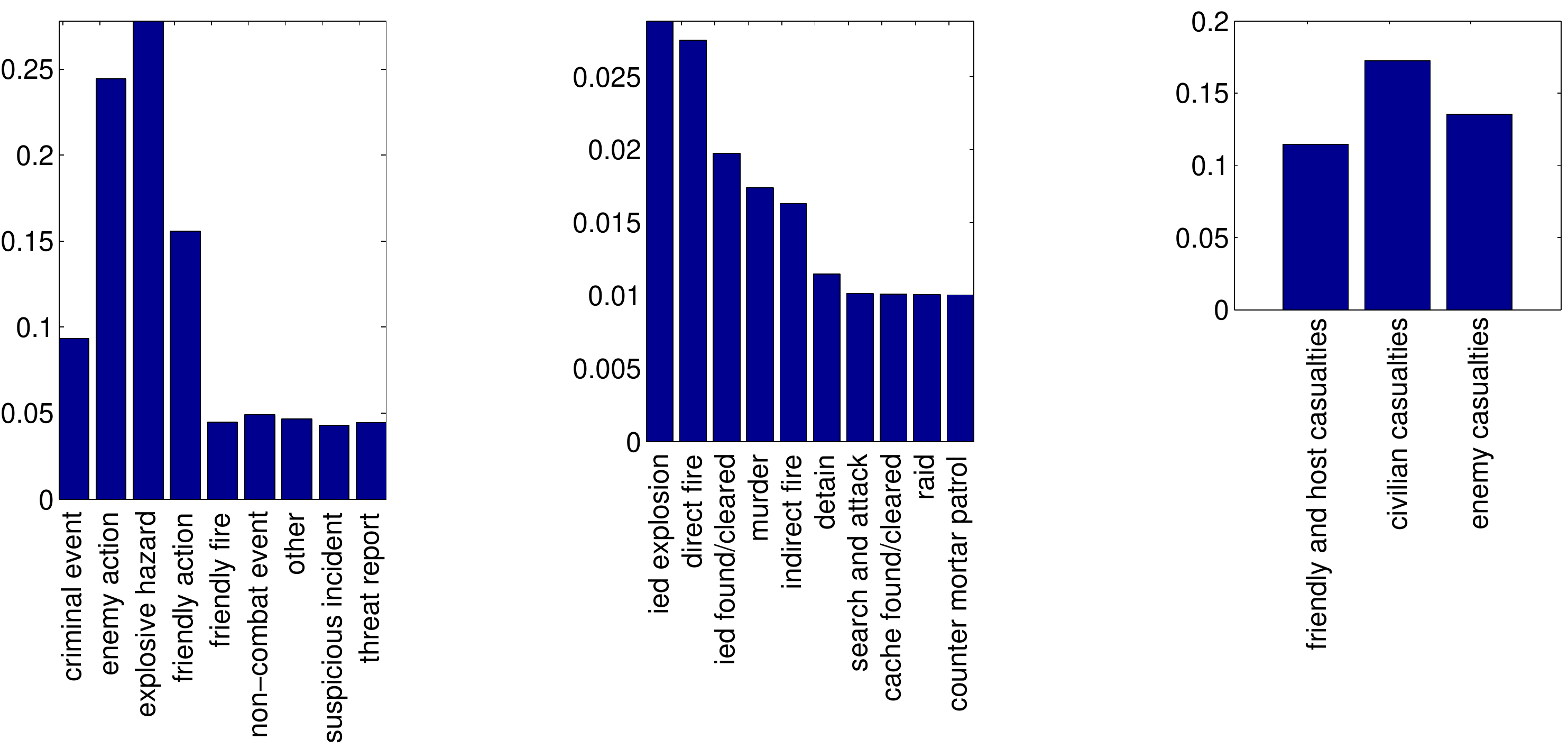}
\caption{State 2 for Iraq (\emph{type}, \emph{category}, \emph{casualties}). \label{fig:cluster2Iraq}}
\end{figure}
\fi

\label{sec:caseStudy}
A primary goal of this work is to establish the practical feasibility of privacy-preserving Bayesian data analysis using complex models on real-world datasets.  In this section we investigate the performance of the methods studied in this paper for the analysis of sensitive military data.  In July and October 2010, the Wikileaks organization disclosed collections of internal U.S. military field reports from the wars in Afghanistan and Iraq, respectively.  Both disclosures contained data from between January 2004 to December 2009, with {\raise.17ex\hbox{$\scriptstyle\sim$}}75,000 entries from the war in Afghanistan, and {\raise.17ex\hbox{$\scriptstyle\sim$}}390,000 entries from Iraq.  Hillary Clinton, at that time the U.S. Secretary of State, criticized the disclosure, stating that it ``puts the lives of United States and its partners' service members and civilians at risk.''\footnote{Fallon, Amy (2010). ``Iraq war logs: disclosure condemned by Hillary Clinton and Nato.'' The Guardian. Retrieved on 2/22/2016.}  These risks, and the motivations for the leak, could potentially have been mitigated by releasing a differentially private analysis of the data
, which protects the contents of each individual log entry while revealing high-level trends. 
Note that since the data are publicly available, although our \emph{models} were differentially private, other aspects of this manuscript such as the evaluation may reveal certain information, as in other works such as \citet{wang2015differentially, wang2015privacy}.

\begin{figure*}[t]
\begin{minipage}{0.33 \linewidth}
\centering \includegraphics[scale=0.4, trim=3.5cm 8.5cm 3.5cm 8.5cm,clip]{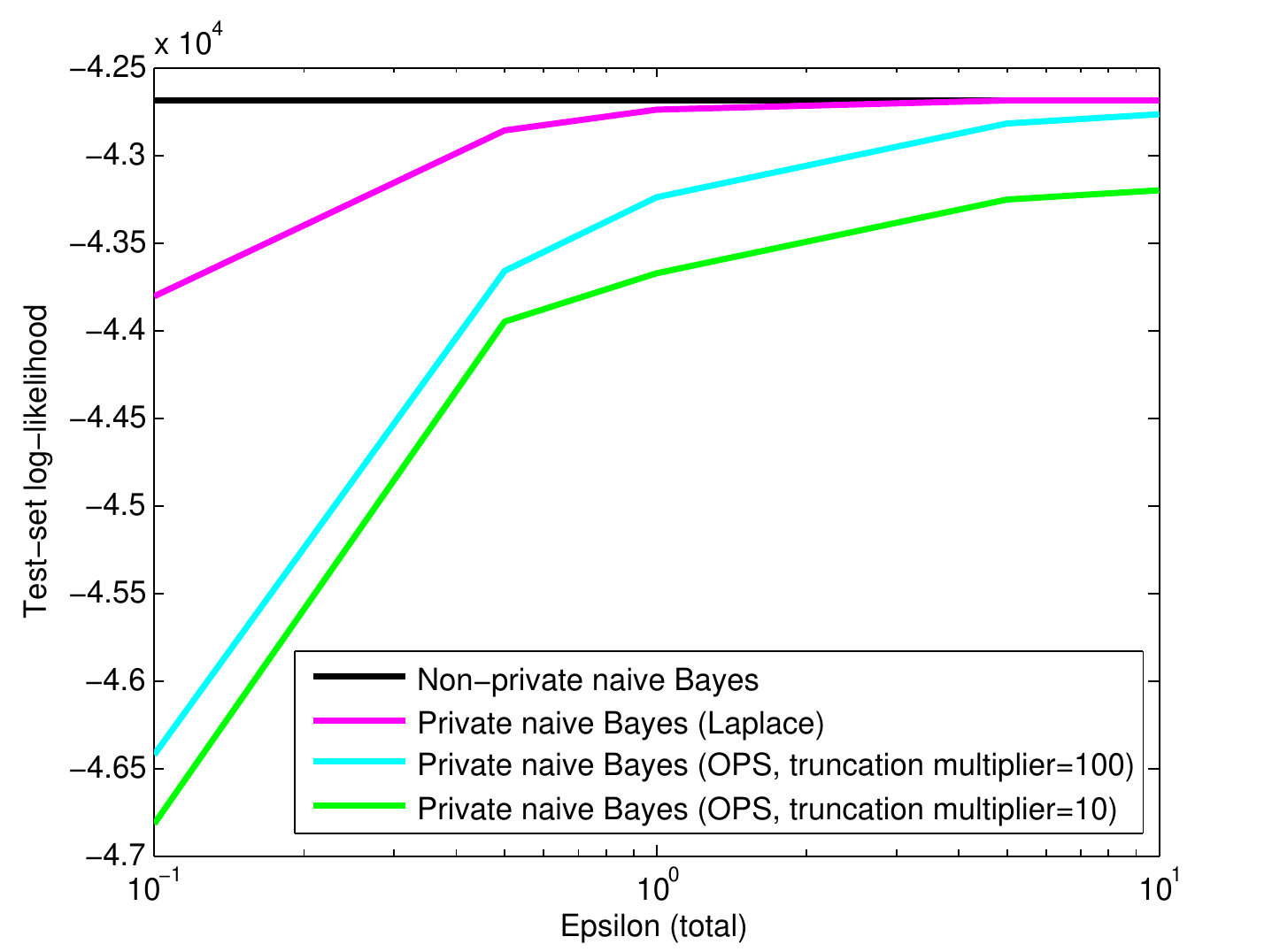}
\end{minipage}
\begin{minipage}{0.33 \linewidth}
\centering
\includegraphics[scale=0.4, trim=3.5cm 8.5cm 3.5cm 8.5cm,clip]{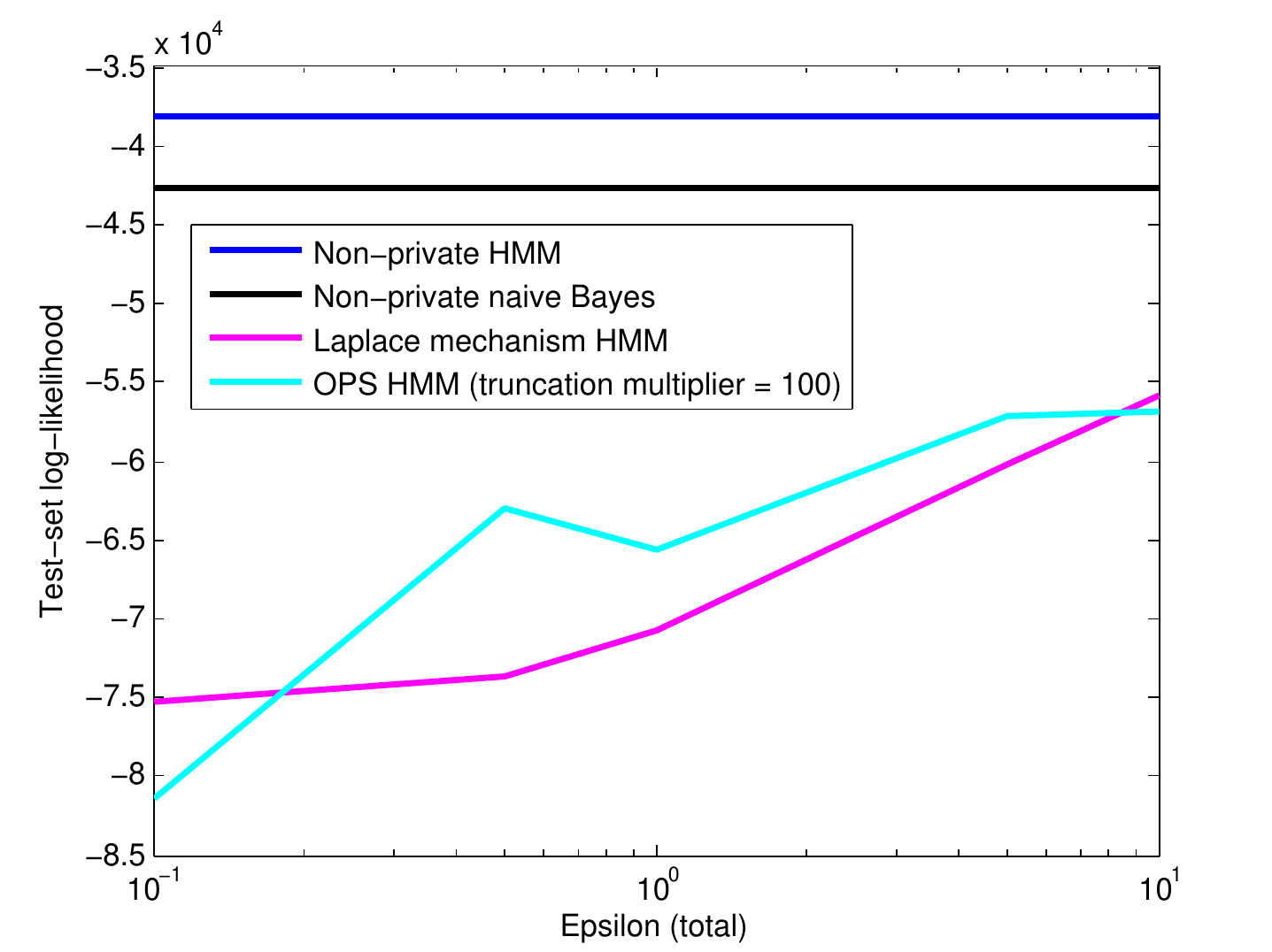}
\end{minipage}
\begin{minipage}{0.33 \linewidth}
\centering
\includegraphics[scale=0.4, trim=3.5cm 8.5cm 3.5cm 8.5cm,clip]{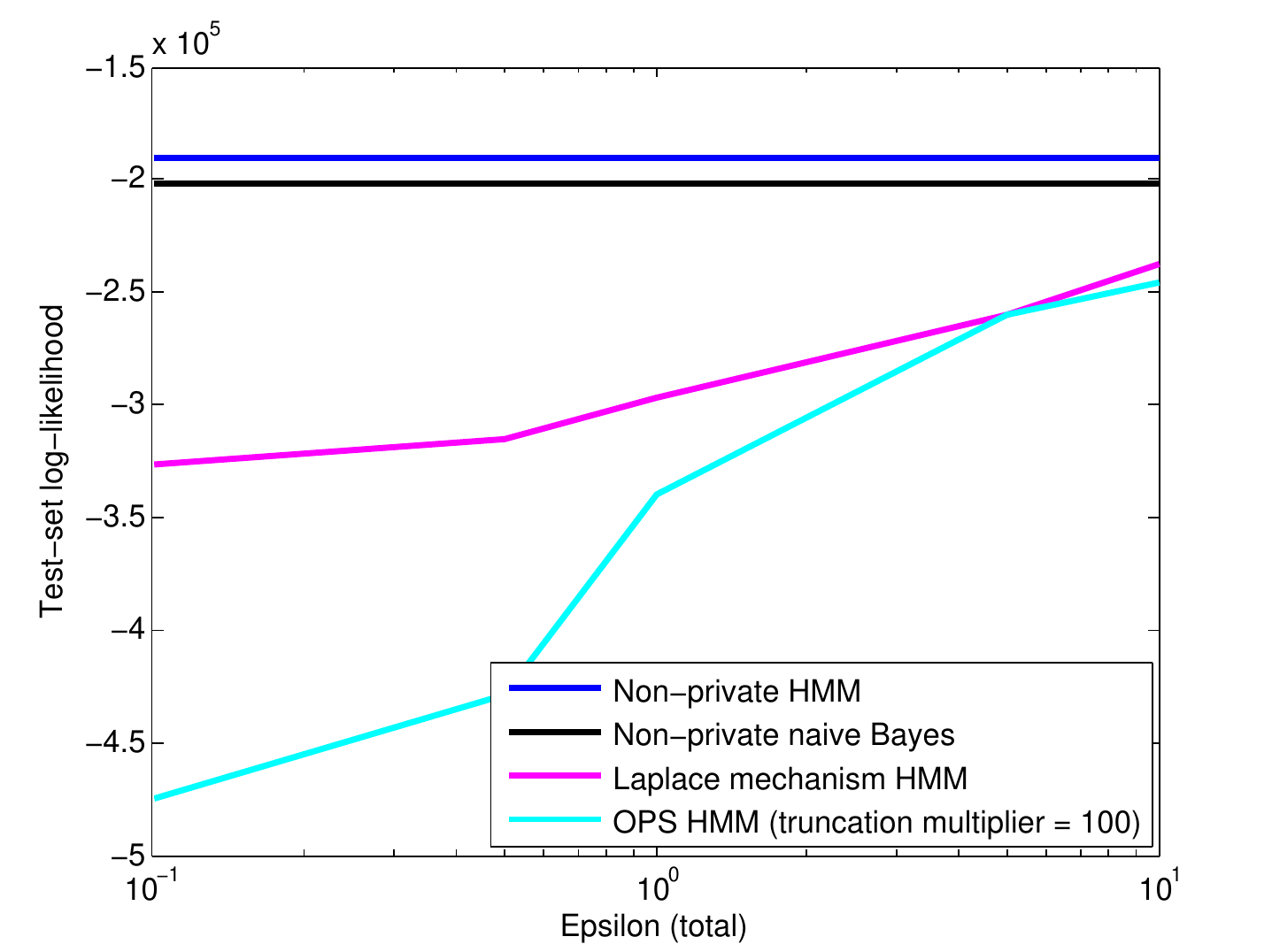}
\end{minipage}
\caption{Log-likelihood results. \textbf{Left:} Naive Bayes  (Afghanistan). \textbf{Middle}: Afghanistan. \textbf{Right}: Iraq. For OPS, Dirichlets were truncated at $a_0 = \frac{1}{M K_d}$, $M=10$ or $100$, where $K_d =$ feature $d$'s dimensionality. \label{fig:HMM_LL}}
\end{figure*}

The disclosed war logs each correspond to an individual event, and contain textual reports, as well as fields such as coarse-grained \emph{types} (\emph{friendly action}, \emph{explosive hazard}, \ldots), fine-grained \emph{categories} (\emph{mine found/cleared}, \emph{show of force}, \ldots), and casualty counts (\emph{wounded}/\emph{killed}/\emph{detained}) for the different factions (\emph{Friendly}, \emph{HostNation} (i.e. Iraqi and Afghani forces), \emph{Civilian}, and \emph{Enemy}, where the names are relative to the U.S. military's perspective).
We use the techniques discussed in this paper to privately infer a hidden Markov model on the log entries.  The HMM was fit to the non-textual fields listed above, with one timestep per month, and one HMM chain per region code.  A naive Bayes conditional independence assumption was used in the emission probabilities for simplicity and parameter-count parsimony.  Each field was modeled via a discrete distribution per latent state, with casualty counts binarized ($0$ versus $>0$), and with \emph{wounded}/\emph{killed}/\emph{detained} and \emph{Friendly}/\emph{HostNation} features combined, respectively, via disjunction of the binary values.  This decreased the number of features to privatize, while slightly increasing the size of the counts per field to protect and simplifying the model for visualization purposes.  After preprocessing to remove empty timesteps and near-empty region codes (see the supplementary), the median number of log entries per region/timestep pair was 972 for Iraq, and 58 for Afghanistan.  The number of log entries per timestep was highly skewed for Afghanistan, due to an increase in density over time.

The models were trained via Gibbs sampling, with the transition probabilities collapsed out, following \citet{goldwater2007fully}.  We did not collapse out the naive Bayes parameters in order to keep the conditional likelihood in the exponential family.  The details of the model and inference algorithm are given in the supplementary material.  We trained the models for 200 Gibbs iterations, with the first 100 used for burn-in.  Both privatization methods have the same overall computational complexity as the non-private sampler.  The Laplace mechanism's computational overhead is paid once up-front, and did not greatly affect the runtime, while OPS roughly doubled the runtime.  For visualization purposes we recovered parameter estimates via the posterior mean based on the latent variable assignments of the final iteration, and we reported the most frequent latent variable assignments over the non-burn-in iterations.  We trained a 2-state model on the Iraq data, and a 3-state model for the Afghanistan data, using the Laplace approach with total $\epsilon = 5$ ($\epsilon = 1$ for each of 5 features).

Interestingly, when given 10 states, the privacy-preserving model only assigned substantial numbers of data points to these 2-3 states, while a non-private HMM happily fit a 10-state model to the data.
The Laplace noise therefore appears to play the role of a regularizer, consistent with the noise being interpreted as a ``random prior,'' and along the lines of noise-based regularization techniques such as 
\citep{srivastava2014dropout, maaten2013learning}, although of course it may correspond to more regularization than we would typically like.  This phenomenon potentially merits further study, beyond the scope of this paper.

We visualized the output of the Laplace HMM for Iraq in Figures \ref{fig:clusterAssignmentsIraq}--\ref{fig:cluster2Iraq}.  State 1 shows the U.S. military performing well, with the most frequent outcomes for each feature being \emph{friendly action}, \emph{cache found/cleared}, and \emph{enemy casualties}, while the U.S. military performed poorly in State 2 (\emph{explosive hazard}, \emph{IED explosion}, \emph{civilian casualties}).  State 2 was prevalent in most regions until the situation improved to State 1 after the troop surge strategy of 2007.  This transition typically occurred after troops peaked in Sept.--Nov. 2007.  The results for Afghanistan, in the supplementary, provide a critical lens on the US military's performance, with enemy casualty rates (including detainments) lower than friendly/host casualties for all latent states, and lower than civilian casualties in 2 of 3 states.

\begin{figure}[t]
\begin{minipage}{\linewidth}
\hspace{-0.22cm}
\includegraphics[scale=0.23]{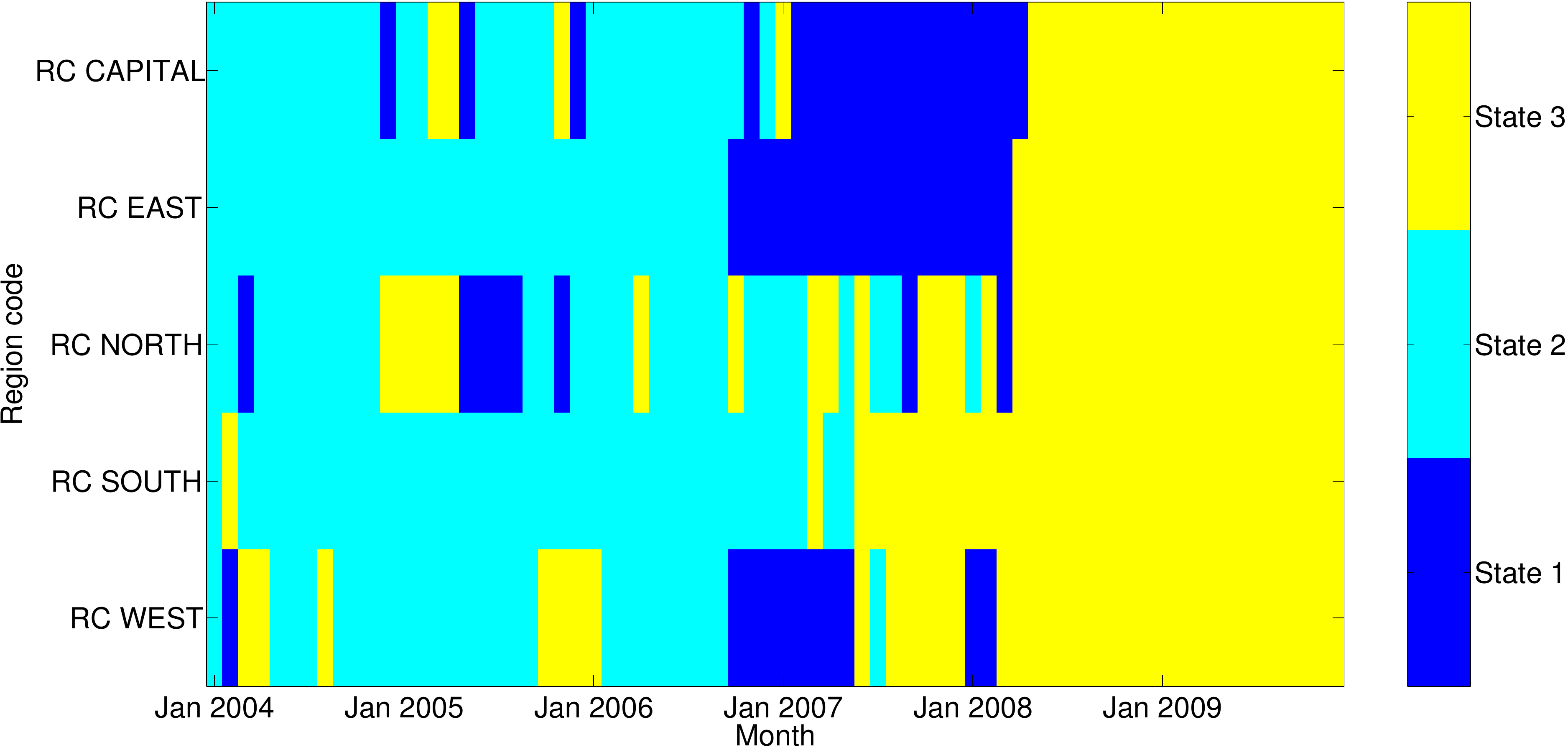}
\end{minipage}

\begin{minipage}{\linewidth}
\hspace{-0.22cm}
\includegraphics[scale=0.23]{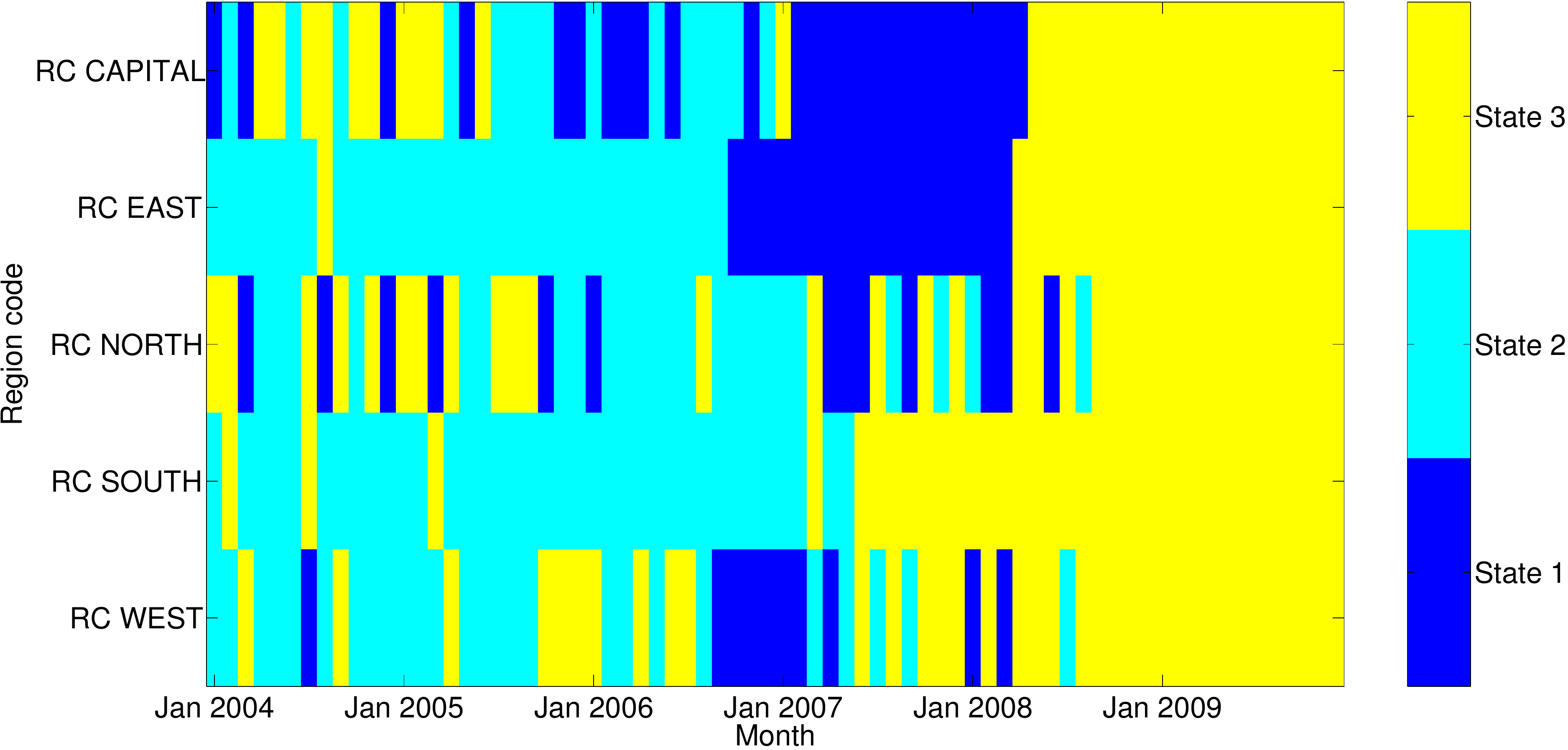}
\end{minipage}
\caption{State assignments for OPS privacy-preserving HMM on Afghanistan.   ($\epsilon = 5$, truncation point $a_0 = \frac{1}{100 K_d}$). \textbf{Top}: Estimate from last 100 samples. \textbf{Bottom}: Estimate from last one sample.
 \label{fig:clusterAssignmentsAfghanistanOPSMainPaper}}
\end{figure}

We also evaluated the methods at prediction.  A uniform random 10\% of the timestep/region pairs were held out for 10 train/test splits, and we reported average test likelihoods over the splits.  We estimated test log-likelihood for each split by averaging the test likelihood over the burned-in samples (Laplace mechanism), or using the final sample (OPS).  All methods were given 10 latent states, and $\epsilon$ was varied between 0.1 and 10.  We also considered a naive Bayes model, equivalent to a 1-state HMM.  The Laplace mechanism was superior to OPS for the naive Bayes model, for which the statistics are corpus-wide counts, corresponding to a high-data regime in which our asymptotic analysis was applicable.  OPS was competitive with the Laplace mechanism for the HMM on Afghanistan, where the amount of data was relatively low.  For the Iraq dataset, where there was more data per timestep, the Laplace mechanism outperformed OPS, particularly in the high-privacy regime.  For OPS, privacy at $\epsilon$ is only guaranteed if MCMC has converged.  Otherwise, from Section \ref{sec:GibbsExpMech}, the worst case is an impractical $\epsilon^{(Gibbs)} \leq 400\epsilon$ (200 iterations of latent variable and parameter updates with worst-case cost $\epsilon$).  OPS only releases one sample, which harmed the coherency of the visualization for Afghanistan, as latent states of the final sample were noisy relative to an estimate based on all 100 post burn-in samples 
(Figure \ref{fig:clusterAssignmentsAfghanistanOPSMainPaper}).
Privatizing the Gibbs chain at a privacy cost of $\epsilon^{(Gibbs)}$ would avoid this.

%% file: conclusion.tex
\section{CONCLUSION}
This paper studied the practical limitations of using posterior sampling to obtain privacy ``for free.''  We explored an alternative based on the Laplace mechanism, and analyzed it both theoretically and empirically.  We illustrated the benefits of the Laplace mechanism for privacy-preserving Bayesian inference to analyze sensitive war records.
The study of privacy-preserving Bayesian inference is only just beginning.  We envision extensions of these techniques to other approximate inference algorithms, as well as their practical application to  sensitive real-world data sets.  Finally, we have argued that asymptotic efficiency is important in a privacy context, leading to an open question: how large is the class of private methods that are asymptotically efficient?

%% file: supplementaryContent.tex
\section{ADVERSARIAL DATA EXPERIMENT}
In this appendix we describe an additional simulation experiment which supplements the analysis performed in the main manuscript.
\ifarxiv
\else
\begin{figure}[t]
\hspace{-0.82cm}
\includegraphics[scale=0.63]{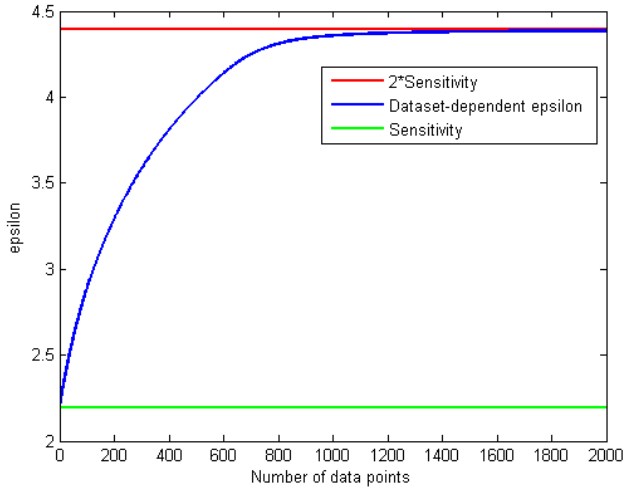}
\caption{An adversary greedily selects data points to add to a dataset to increase the dataset-specific privacy cost $\epsilon$ of posterior sampling via the exponential mechanism (OPS). \label{fig:adversarialData}}
\end{figure}
\fi
\citet{wang2015privacy}'s analysis finds that the privacy cost of posterior sampling does not directly improve with the number of data points $N$, unless the analyst deliberately modifies the posterior by changing the temperature before sampling.  In Figure \ref{fig:adversarialData} we report an experiment showing that this result is not just a limitation of the analysis: there do exist cases where the dataset-specific privacy cost of posterior sampling can approach the exponential mechanism worst case of $\epsilon = 2\triangle \log Pr(\theta,\mathbf{X})$ as the number of observations $N$ increases.

In the experiment, we consider a beta distribution posterior, symmetrically truncated at $a_0 = 0.1$, with Bernoulli observations.  We simulate an adversary who greedily selects data points to add to a dataset to increase the dataset-specific privacy cost $\epsilon$ of posterior sampling. The dataset-specific ``local'' privacy parameter $\epsilon$ is computed via a grid search over the Bernoulli success parameter $p$ and Bernoulli outcomes $x$, $x'$, for the case where the adversary adds a success, or a failure, and the adversary selects the success/failure outcome with the highest local $\epsilon$.  The adversary is able to make the dataset-specific $\epsilon$ approach the worst case by manipulating the partition function of the posterior.  The exponential mechanism's worst case for posterior sampling, $\epsilon = 2\triangle \log Pr(\theta,\mathbf{X})$, corresponds to a sum of two cost terms.  We must pay a cost of $\triangle \log Pr(\theta,\mathbf{X})$ from to the difference of log-likelihood terms, as we can always draw the worst-case $\theta$ (e.g., when $p$ is on the truncation boundary), plus another $\triangle \log Pr(\theta,\mathbf{X})$ in the worst case due to the difference of log partition-functions terms, which the adversary can alter up to the worst case, as they do in Figure \ref{fig:adversarialData}.  This is described formally in the supplementary of \citep{wang2015privacy}.


\section{PROOFS OF THEORETICAL RESULTS}
Here we provide proofs for the results presented in Section \ref{sec:theoreticalResults}.

\input{supplementaryExpFamilyProofs.tex}

\section{PRIVACY PROPERTIES OF OTHER MCMC ALGORITHMS}

In the main manuscript we showed the privacy cost of Gibbs sampling by interpreting it as an instance of the exponential mechanism.  Here, we show the privacy cost of two other widely used MCMC algorithms: Metropolis-Hastings and annealed importance sampling.

\subsection{METROPOLIS-HASTINGS UPDATES}
Since Gibbs updates are a special case of Metropolis-Hastings updates, one might conjecture that general Metropolis-Hastings updates may be differentially private as well.  However, the accept/reject decision contains a subtle non-determinacy which violates pure-$\epsilon$ differential privacy.  
\ifarxiv
\else
Consider a Metropolis-Hastings update with a symmetric proposal $\theta' \sim f(\theta,\theta') $ (a.k.a. a Metropolis update),
\fi
\begin{equation}
\hspace{-1.5mm}
Pr(\mbox{accept}; \mathbf{X}, \theta, \theta', T) = \min \Big (1, \big (\frac{Pr(\theta'|\mathbf{X})}{Pr(\theta|\mathbf{X})} \big )^{\frac{1}{T}} \Big )
\end{equation}
where $T$ is the temperature of the Markov chain.
For these updates, ``uphill'' moves are never rejected.  Since a move may be uphill in one database and downhill in a neighbor, we cannot bound the ratio of reject decisions, which violates differential privacy.  It turns out that Metropolis updates do have a weaker privacy guarantee, by resorting to $(\epsilon,\delta)$-differential privacy:
\begin{theorem}
\label{thm:metropolis}
Let $\mathbf{X}$ be private data and $\theta$ be a public current value of the variables we wish to infer.  A Metropolis update invariant to the posterior $Pr(\theta|\mathbf{X})$ at temperature $T = \frac{2 \triangle \log Pr(\theta,\mathbf{X})}{\epsilon}$, with symmetric proposal $\theta' \sim f(\theta,\theta')$ and with $Pr(\mbox{reject}; \mathbf{X}, \theta, T) = \int f(\theta,\theta')(1-Pr(\mbox{accept}; \mathbf{X}, \theta, \theta', T)) d \theta' \leq \delta$, is $(\epsilon,\delta)$-differentially private.
\end{theorem}
A proof of Theorem \ref{thm:metropolis} is provided below in Appendix \ref{app:metropolis}.  Essentially, we can bound the ratio of probabilities for accept decisions under neighboring databases, but not for reject decisions.  If rejections are rare, these privacy-violating outcomes are rare, which is sufficient for $(\epsilon,\delta)$-privacy.  On the other hand, $\delta$ must be very small for a meaningful level of privacy, e.g. less than the inverse of any polyomial in the number of data points $N$ \citep{dwork2013algorithmic}, so this may not typically correspond to a practical privacy-preserving sampling algorithm.  
\subsection{ANNEALED IMPORTANCE SAMPLING}
The privacy results for Gibbs sampling and Metropolis-Hastings updates reveal a close connection between privacy and the temperature of the Markov chain.
Low-temperature chains are high-fidelity but privacy-expensive, while high-temperature chains are low-fidelity but privacy-cheap, and also mix more rapidly.  This suggests that annealing methods, such as annealed importance sampling (AIS) \citep{neal2001annealed}, may be effective in this context, by allowing savings in the privacy budget in the early iterations of MCMC while also traversing the state space more rapidly.  AIS is a Monte Carlo method which anneals from a high-temperature distribution to the target distribution (in our case the posterior) via MCMC updates at a sequence of temperatures, producing importance weights for each sample to correct for the annealing.
AIS takes as input an annealing path, a sequence of unnormalized distributions $f_n(\theta), \ldots, f_0(\theta)$ at different temperatures.  We can obtain a privacy-preserving AIS annealing path by varying $\epsilon$:  
\begin{equation}
f_j(\theta) = {Pr(\theta,\mathbf{X})}^{\beta_j} \mbox{ , } \beta_j = \frac{\epsilon_j}{2 \triangle \log Pr(\theta,\mathbf{X})} \mbox{ ,} 
\end{equation}
where each intermediate distribution $f_j$ is an instance of Equation \ref{eqn:expMechPosterior}, and $\epsilon_j$ is the privacy cost for an exact sample from $f_j$.  We can sample at each temperature using the private Gibbs transition operator from Equation \ref{eqn:expMechGibbs}.  The privacy cost of an AIS sample is computed via the composition theorem,
\begin{equation}
\epsilon^{(AIS)} = \sum_j\sum_l \epsilon_j = \sum_j D\epsilon_j \mbox{ ,}
\end{equation}
where $l$ ranges over the $D$ variables to be updated.
If each Gibbs update only depends on a single data point $\mathbf{x}_l$, we can improve this via parallel composition \citep{song2013stochastic} to
\begin{equation}
\epsilon^{(AIS)} = \sum_j\epsilon_j \mbox{ .}
\end{equation}
On completion of the algorithm we must compute importance weights $\omega_i$ for the samples $\theta^{(i)}$:
\ifarxiv
\else
\begin{align}
&\log \omega_i = \sum_{j=0}^{n-1}\Big ( \log f_j(\theta^{(i,j)}) - \log f_{j+1}(\theta^{(i,j)}) \Big )\\
 &=  \sum_{j=0}^{n-1}\Big (\beta_j \log {Pr(\theta^{(i,j)},\mathbf{X})} - \beta_{j+1}\log {Pr(\theta^{(i,j)},\mathbf{X})} \Big )\nonumber \\
&= \frac{1}{2 \triangle \log Pr(\theta,\mathbf{X})} \sum_{j=0}^{n-1}(\epsilon_j - \epsilon_{j+1})\log {Pr(\theta^{(i,j)},\mathbf{X})} \mbox{.} \nonumber 
\end{align}
\fi
We only need to release private copies of the importance weights at the end of the procedure, as they are not used during the algorithm.  If we are not interested in computing normalization constants, we can release a normalized version of the weights, dividing by $\sum_i \omega_i$.  This is a discrete distribution which sums to one, and so it lives on the simplex.  This has $L1$ sensitivity at most 2, and can be protected by the Laplace mechanism.  Another possible alternative is to perform resampling of the $\theta^{(i)}$'s according to this distribution, approximated and protected via the exponential mechanism.

\section{PROOF OF THEOREM \ref{thm:metropolis}}
\label{app:metropolis}
Here, we prove the differential privacy result for Metropolis-Hastings given in Theorem \ref{thm:metropolis}, above.
\begin{proof2}
Let $\mathbf{X}^{(1)}$, $\mathbf{X}^{(2)}$ be neighboring databases.  By the definition of differential privacy, we need to bound the ratios of the probability of each outcome, $\{(\mbox{accept}, \mbox{reject}), \theta^{(new)}\}$ for these two databases.  We consider accept and reject outcomes separately.
\subsection{ACCEPT OUTCOME}
The probability of an accepted move to location $\theta^{(new)} = \mathbf{z'}$ is
\ifarxiv
\else
\begin{align*}
Pr(\mbox{accept}&, \theta^{(new)}= \theta'; \mathbf{X}, \theta) \\
&= f(\theta,\theta')Pr(\mbox{accept}; \mathbf{X}, \theta, \theta', T) \mbox{ .}
\end{align*}
\fi
We must bound the probability ratio of this outcome under the two neighboring datasets.
Consider first a slightly simpler question, the ratio of probabilities for an accept decision, having already selected the proposal $\theta'$,
\begin{equation*}
\frac{Pr(\mbox{accept}; \mathbf{X}^{(1)}, \theta,\theta')}{Pr(\mbox{accept}; \mathbf{X}^{(2)}, \theta, \theta')} \mbox{ .}
\end{equation*}
We will perform the computation in log space. We have the log of the acceptance probabilities as
\ifarxiv
\else
\begin{align*}
 &\log Pr(\mbox{accept}; \mathbf{X}, \theta, \theta', T)=\\
 &\min \Big (0, \frac{\epsilon }{2 \triangle \log Pr(\theta,\mathbf{X})} \big (\log Pr(\theta'|\mathbf{X}) - \log Pr(\theta|\mathbf{X}) \big ) \Big )\\
 &=\frac{\epsilon }{2 \triangle \log Pr(\theta,\mathbf{X})}\min \Big (0, \log Pr(\theta'|\mathbf{X}) - \log Pr(\theta|\mathbf{X}) \Big ) \mbox{.}
\end{align*}
\fi
The difference in log probabilities for the $\mbox{accept}$ outcome is
\ifarxiv
\else
\begin{flalign*}
\log & Pr(\mbox{accept}; \mathbf{X}^{(1)}, \theta, \theta') - \log Pr(\mbox{accept}; \mathbf{X}^{(2)}, \theta, \theta')\\
= & \frac{\epsilon }{2 \triangle \log Pr(\theta, \mathbf{X})} \times\\ &\Big (\min \big (0, \log Pr(\theta'|\mathbf{X}^{(1)}) - \log Pr(\theta|\mathbf{X}^{(1)}) \big ) \\
&-\min \big (0, \log Pr(\theta'|\mathbf{X}^{(2)}) - \log Pr(\theta|\mathbf{X}^{(2)}) \big ) \Big ) \mbox{ .} \\
\end{flalign*}
\fi
Let
\begin{align*}
a &= \log Pr(\theta'|\mathbf{X}^{(1)}) - \log Pr(\theta|\mathbf{X}^{(1)})\\
b &= \log Pr(\theta'|\mathbf{X}^{(2)}) - \log Pr(\theta|\mathbf{X}^{(2)}) \mbox{ .}
\end{align*}
There are four cases to consider:
\ifarxiv
\else
\begin{itemize}
\item $a \leq 0, b \leq 0$:
$$ min(0,a) - min(0,b) = a - b$$
\item $a > 0, b \leq 0$:
$$ min(0,a) - min(0,b) = - b \leq a - b$$
\item $a \leq 0, b > 0$:
$$ min(0,a) - min(0,b) = a \leq 0$$
\item $a > 0, b > 0$:
$$ min(0,a) - min(0,b) = 0 \mbox{ .}$$
\end{itemize}
\fi
So either $min(0,a) - min(0,b) \leq 0$, in which case the difference in log probabilities is $\leq 0 \leq \epsilon$, or $min(0,a) - min(0,b) \leq a - b$.  In the former, we are done, so consider the latter case:
\ifarxiv
\else
\begin{flalign*}
\log Pr(\mbox{accept}; \mathbf{X}^{(1)}&, \theta, \theta') - \log Pr(\mbox{accept}; \mathbf{X}^{(2)}, \theta, \theta')\\
\leq \frac{\epsilon }{2 \triangle \log Pr(\theta, \mathbf{X})} \Big ( &\big ( \log Pr(\theta'|\mathbf{X}^{(1)}) - \log Pr(\theta|\mathbf{X}^{(1)}) \big )\\
- &\big ( \log Pr(\theta'|\mathbf{X}^{(2)}) - \log Pr(\theta|\mathbf{X}^{(2)}) \big ) \Big )\\
=\frac{\epsilon }{2 \triangle \log Pr(\theta, \mathbf{X})} \Big ( & \log Pr(\theta'|\mathbf{X}^{(1)}) - \log Pr(\theta'|\mathbf{X}^{(2)})\\
 + &\log Pr(\theta|\mathbf{X}^{(2)}) - \log Pr(\theta|\mathbf{X}^{(1)}) \Big )\\
=\frac{\epsilon }{2 \triangle \log Pr(\theta, \mathbf{X})} \Big ( & \log Pr(\theta'|\mathbf{X}^{(1)}) - \log Pr(\theta'|\mathbf{X}^{(2)})\\
 + & \log Pr(\mathbf{X}^{(1)}) - \log Pr(\mathbf{X}^{(2)})\\
 + &\log Pr(\theta|\mathbf{X}^{(2)}) - \log Pr(\theta|\mathbf{X}^{(1)})\\
 + &\log Pr(\mathbf{X}^{(2)}) - \log Pr(\mathbf{X}^{(1)}) \Big )\\
=\frac{\epsilon }{2 \triangle \log Pr(\theta, \mathbf{X})} \Big ( & \log Pr(\theta',\mathbf{X}^{(1)}) - \log Pr(\theta',\mathbf{X}^{(2)})\\
 + &\log Pr(\theta,\mathbf{X}^{(2)})
  - \log Pr(\theta,\mathbf{X}^{(1)}) \Big )\\
\leq \frac{\epsilon }{2 \triangle \log Pr(\theta, \mathbf{X})} \Big (&\triangle \log Pr(\theta,\mathbf{X}) + \triangle \log Pr(\theta,\mathbf{X}) \Big )\\
 = \epsilon \mbox{ .}&&
\end{flalign*}
\fi
The inequality in the last line follows from Equation \ref{eqn:expSensPosterior} in the main paper.

Having bounded the log ratio of probabilities by $\epsilon$ for the simpler case where the proposal $\theta'$ is given, we can now bound the ratios for the full output, of the form $(\mbox{accept}, \theta^{(new)})$, as required for $\epsilon$-differential privacy, by simply cancelling the log transition probabilities:
\ifarxiv
\else
\begin{align*}
 &\log Pr(\mbox{accept}, \theta^{(new)} = \theta'; \mathbf{X}^{(1)}, \theta)\\
  &- \log Pr(\mbox{accept}, \theta^{(new)} = \theta'; \mathbf{X}^{(2)}, \theta) \nonumber \\
 = &\log f(\theta,\theta')+ \log Pr(\mbox{accept}; \mathbf{X}^{(1)}, \theta, \theta')\\
  &-(\log f(\theta,\theta')+ \log Pr(\mbox{accept}; \mathbf{X}^{(2)}, \theta, \theta'))  \nonumber \\ 
 =& \log Pr(\mbox{accept}; \mathbf{X}^{(1)}, \theta, \theta') - \log Pr(\mbox{accept}; \mathbf{X}^{(2)}, \theta, \theta') \nonumber \\
 \leq & \mbox{ } \epsilon \mbox{ .}
\end{align*}
\fi
This is as desired for pure-$\epsilon$ privacy, and so the weaker $(\epsilon,\delta)$-criterion holds for this outcome as well.
\subsection{REJECT OUTCOME}
If we could also similarly bound the difference in log probabilities between neighboring databases for the outcome $(\mbox{reject}, \theta^{(new)} = \theta)$ by $\epsilon$, then the Metropolis update would be $\epsilon$-differentially private.  Consider first the reject probabilities after the proposal $\theta'$ is selected:
\ifarxiv
\else
\begin{align*}
 Pr & (\mbox{reject}; \mathbf{X}, \theta, \theta') \\
 &= 1-\min \Big (1, \big (\frac{Pr(\theta'|\mathbf{X})}{Pr(\theta|\mathbf{X})} \big )^{\frac{\epsilon }{2 \triangle \log Pr(\theta, \mathbf{X})}} \Big ) \\
 &= \max \Big (0, 1-\big (\frac{Pr(\theta'|\mathbf{X})}{Pr(\theta|\mathbf{X})} \big )^{\frac{\epsilon }{2 \triangle \log Pr(\theta, \mathbf{X})}} \Big ) \mbox{ .}
\end{align*}
\fi
When $Pr(\theta'|\mathbf{X}) > Pr(\theta|\mathbf{X})$, the probability of a reject decision is 0.  It is possible to construct scenarios where the probability of a reject decision is 0 for all proposals $\theta'$, e.g. when $\theta$ is at a global minimum, so we cannot in general lower bound the overall probability of a reject,
\ifarxiv
\else
\begin{align*}
Pr(\mbox{reject}&, \theta^{(new)}= \theta; \mathbf{X}, \theta)\\
 &= \int f(\theta,\theta')(1-Pr(\mbox{accept}; \mathbf{X}, \theta, \theta', T)) d\theta' \mbox{ .}
\end{align*}
\fi
If $Pr(\mbox{reject}, \theta^{(new)} = \theta; \mathbf{X}, \theta) = 0$ occurs in database $\mathbf{X}^{(1)}$ and not in $\mathbf{X}^{(2)}$, the ratio of probabilities for this outcome will be infinite due to a division by 0, violating $\epsilon$-differential privacy.  Under our assumptions, we have an additional guarantee that  $Pr(\mbox{reject}, \theta^{(new)} = \theta; \mathbf{X}, \theta) \leq \delta$, i.e. the probability of a rejection outcome, and therefore the probability of an outcome that violates $\epsilon$-differential privacy, is less than $\delta$.
To demonstrate $(\epsilon,\delta)$ privacy and complete the proof, we observe that this condition implies that the $(\epsilon,\delta)$-criterion holds for the reject outcome:
\ifarxiv
\else
\begin{align*}
Pr(&\mbox{reject}, \theta^{(new)} = \theta; \mathbf{X}^{(1)}, \theta)\\
 &\leq \delta \leq \exp(\epsilon)Pr(\mbox{reject}, \theta^{(new)} = \theta; \mathbf{X}^{(2)}, \theta) + \delta \mbox{ .}
\end{align*}
\fi
\end{proof2}

\begin{table*}[t]
	\begin{tabular}{ll}
	$x_{i,d}^{(r,t)}$ & Discrete-valued feature $d$ of log entry $i$, from region $r$, timestep $t$.\\
	$z_{r,t}$ & Latent state at region $r$, timestep $t$.\\
	$A_{k,k}$ & Transition probability from state $k$ to $k'$.\\
	$\theta_j^{(k,d)}$ & Discrete emission probability for cluster $k$'s $d$'th feature being outcome $j$.\\
	$\alpha, \beta$ & Dirichlet concentration parameters.\\
	$N_{r,t}$ & Number of log entries (observations) in region $r$ at timestep $t$. \\
	$D$ & Number of features in the observations.\\
	$K$ & Number of latent clusters.
	\end{tabular}
	\caption{Notation for the Wikileaks naive Bayes HMM model. \label{tab:HMMnotation}}
\end{table*}

\section{DETAILS OF WIKILEAKS WAR LOGS HMM}

\begin{figure*}[t]
\begin{minipage}{0.49\linewidth}
\centering
 \includegraphics[scale=0.5, trim=3.5cm 8.5cm 3.5cm 8.5cm,clip]{figures/HMM/naiveBayesLL}
\end{minipage}
\begin{minipage}{0.49\linewidth}
\centering
 \includegraphics[scale=0.5, trim=3.5cm 8.5cm 3.5cm 8.5cm,clip]{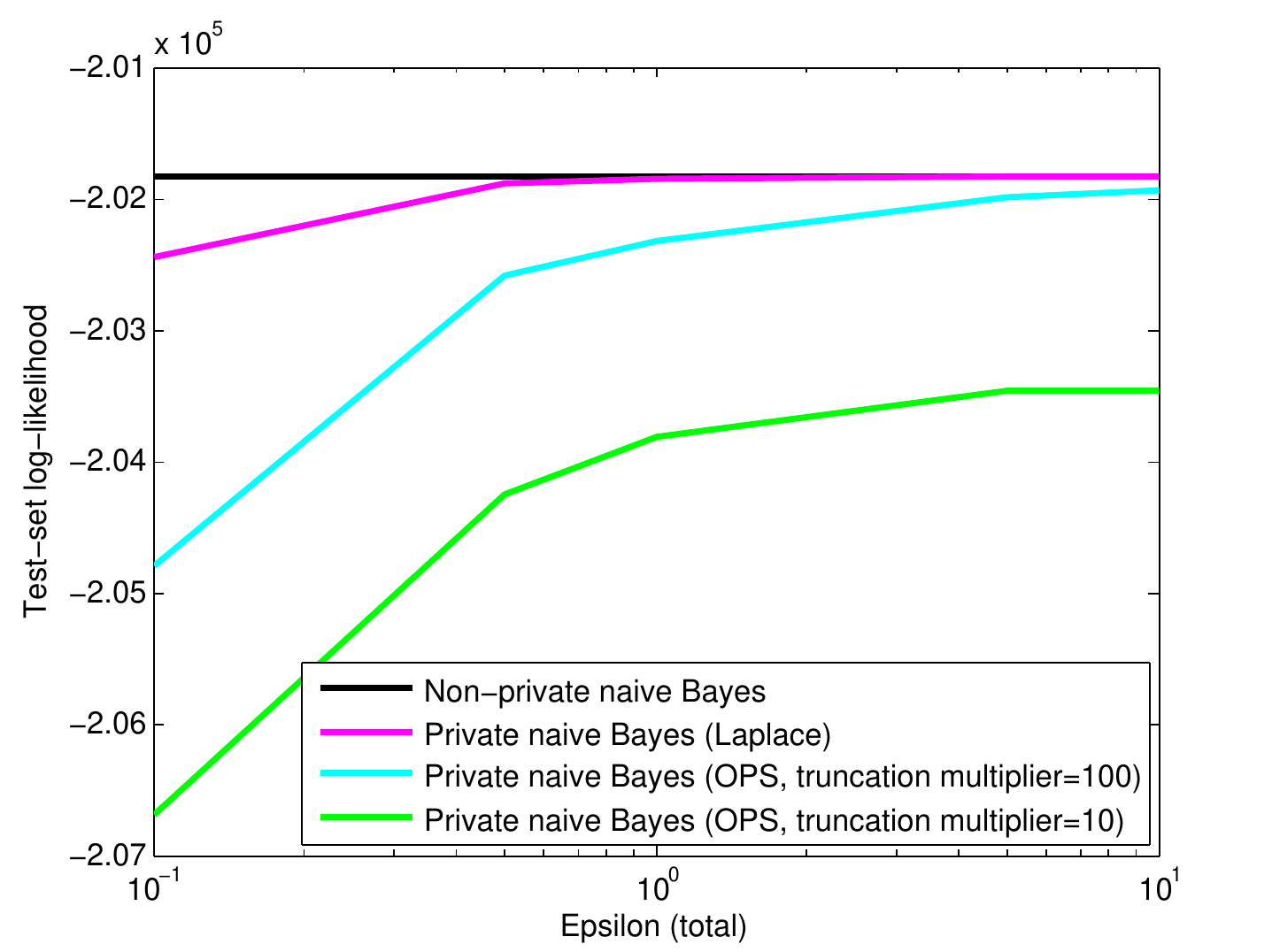}
\end{minipage}

\begin{minipage}{0.49\linewidth}
\centering
 \includegraphics[scale=0.5, trim=3.5cm 8.5cm 3.5cm 8.5cm,clip]{figures/HMM/HMM_LL_Afghanistan}
\end{minipage}
\begin{minipage}{0.49\linewidth}
\centering
 \includegraphics[scale=0.5, trim=3.5cm 8.5cm 3.5cm 8.5cm,clip]{figures/HMM/HMM_LL_Iraq}
\end{minipage}

\caption{Log-likelihood of held-out data for a naive Bayes model, equivalent to the HMM with one timestep (\textbf{Top}) and the full HMM (\textbf{Bottom}).  \textbf{Left:} Afghanistan. \textbf{Right:} Iraq. Truncation point for the truncated Dirichlet distributions for OPS was set to $a_0 = \frac{1}{M K_d}$, with truncation multiplier $M=10$ and $M=100$.  \label{fig:naiveBayesLL}}
\end{figure*}

In this appendix we describe the technical details of the HMM model with naive Bayes observations, which we apply to the Wikileaks War Logs data.  The assumed generative process of the model is:
\begin{itemize}
	\item For $k = 1, \ldots, K$ //For each latent cluster
	\begin{itemize}
		\item $\mathbf{A}_{k,:} \sim Dirichlet(\alpha)$ //$K$-dimensional
		\item For $d = 1, \ldots, D$ //For each feature
		\begin{itemize}
			\item $\theta^{(k,d)} \sim \mbox{Dirichlet}(\beta)$ //$K_d$-dimensional
		\end{itemize}
	\end{itemize}
	\item $\mathbf{A}_{0,:} \sim Dirichlet(\alpha)$ //Dummy state 
	\item For $r = 1, \ldots, R$ //For each region
	\begin{itemize}
		\item $z_{r,0} = 0$ //Dummy initial state 
		\item For $t = 1, \ldots, T$ //For each timestep
		\begin{itemize}
			\item $z_{r,t} \sim \mbox{Discrete}(\mathbf{A}_{z_{r,t-1},:})$
			\item For $i = 1, \ldots, N_{r,t}$ //For each log entry
			\begin{itemize}
				\item For $d = 1, \ldots, D$ //For each feature
				\begin{itemize}
					\item $x^{(r,t)}_{i,d} \sim \mbox{Discrete}(\theta^{(z_{r,t},d)})$.
				\end{itemize}
			\end{itemize}
		\end{itemize}
	\end{itemize}
\end{itemize}

Here, $\alpha$ and $\beta$ correspond to the concentration parameters for appropriately dimensioned Dirichlet distributions.  See Table \ref{tab:HMMnotation} for a summary of the notation.  The generative model corresponds to the joint probability
\ifarxiv
\else
\begin{align}
 & Pr(\mathbf{A}, \theta, \mathbf{Z}, \mathbf{X} |\alpha, \beta) = \\
 & \prod_{k=0}^K Pr(\mathbf{A}_{k,:}|\alpha) \prod_{k=1}^K \prod_{d=1}^D Pr(\theta^{(k,d)}|\beta) \nonumber \\
  & \times \prod_{r=1}^R \prod_{t=1}^T Pr(z_{r,t}|z_{r,t-1}, \mathbf{A}) \nonumber \\
  &\times\prod_{r=1}^R \prod_{t=1}^T \prod_{i=1}^{N_{r,t}} Pr(x^{(r,t)}_{i,d} | z_{r,t}, \theta) \mbox{ .} \nonumber
\end{align}
\fi
Inspired by \citet{goldwater2007fully}, we marginalize out the transition matrix $\mathbf{A}$.  Let $\mathbf{X}^{(r,t)}$ be an $N_{r,t} \times D$ matrix containing the log entry observations at region $r$, timestep $t$.  We obtain the following partially collapsed Gibbs update for $z_{r,t}$:
\ifarxiv
\else
\begin{align}
Pr&(z_{r,t}|z_{r,t-1}, z_{r,t+1}, \mathbf{X}^{(r,t)}, \theta, \alpha) \label{eqn:HMMGibbs} \\
 \propto &  Pr(z_{r,t}|z_{r,t-1}) Pr(z_{r,t+1}|z_{r,t}) Pr(\mathbf{X}^{(r,t)}|z_{r,t}, \theta) \nonumber \\
=& \frac{n_{z_{r,t},z_{r,t-1}} + \alpha}{n_{z_{r,t-1}} + K\alpha} \frac{n_{z_{r,t+1},z_{r,t}} + \mathbb{I}[z_{r,t-1} = z_{r,t} = z_{r,t+1}] + \alpha}{n_{z_{r,t}} + \mathbb{I}[z_{r,t-1} = z_{r,t}]+ K\alpha} \nonumber \\
& \times Pr(\mathbf{X}^{(r,t)}|z_{r,t}, \theta) \mbox{ ,} \nonumber
\end{align}
\fi
where $n_{z,z'}$ are transition counts, excluding the current z to be updated, and the transition probabilities are implicitly conditioned on all other $\mathbf{z}$'s, which they depend on via the transition counts.  The indicator functions arise from bookkeeping as the counts are modified by changing the current state.  Due to conjugacy we have a simple update for $\theta^{(k,d)}$,
\begin{align}
Pr(\theta^{(k,d)}| \mathbf{X}, \mathbf{Z}, \beta) \sim \mbox{Dirichlet}(n_{d,k,:} + \beta) \mbox{ ,} \label{eqn:updateTheta}
\end{align}
where $n_{d,k,:} = \sum_{r,t}n_{r,t,d,:}$ is a $K_d$-dimensional count vector of counts for feature $d$ in cluster $k$.
\subsection{PRESERVING PRIVACY}
To privatize the likelihood via the Laplace mechanism, we first write Equation \ref{eqn:naiveBayeslikelihood} in exponential family form. 
The conditional likelihood for $\mathbf{X}^{(r,t)}$ given $z_{r,t}$ can be written as
\begin{align}
Pr(\mathbf{X}^{(r,t)} | z_{r,t}, \theta) &= \prod_{i=1}^{N_{r,t}} Pr(x^{(r,t)}_{i,d} | z_{r,t}, \theta) \label{eqn:naiveBayeslikelihood} \\
&= \prod_{i=1}^{N_{r,t}} \prod_{d=1}^D  \theta^{(z_{r,t},d)}_{x^{(r,t)}_{i,d}} \nonumber \\
&= \prod_{d=1}^D\prod_{j=1}^{K_d}  \theta_j^{(z_{r,t},d)^{n_{r,t,d,j}}} \mbox{ ,}  \nonumber
\end{align}
where $n_{r,t,d,j} = \sum_{i=1}^{N_{r,t}} \mathbb{I}[x^{(r,t)}_{i,d} = j]$, and $\mathbb{I}[\cdot]$ is the indicator function.  From here we obtain the exponential family form
\begin{equation}
Pr(\mathbf{X}^{(r,t)} | z_{r,t}, \theta) = \exp \Big ( \sum_{d=1}^D\sum_{j=1}^{K_d} n_{r,t,d,j} \log \theta_j^{(z_{r,t},d)} \Big ) \mbox{ .}
\end{equation}
The sufficient statistics are the counts $n_{r,t,d,j}$, which we can privatize via the Laplace mechanism, resulting in private counts $\hat{n}_{r,t,d,j}$.  As a sum of indicator vectors, each count vector $n_{r,t,d,:}$ has L1 sensitivity = 2.  We can perform the Gibbs updates for $\mathbf{Z}$ in a privacy-preserving manner by substituting the private counts for the counts in Equation \ref{eqn:HMMGibbs}.
To preserve privacy when updating $\theta$, Equation \ref{eqn:updateTheta} can be estimated based on the privacy-preserving counts $\hat{n}_{r,t,d,:}$.  Importantly, we only need to compute private counts $\hat{n}_{r,t,d,j}$ once, at the beginning of the algorithm, and these privatized counts can be reused for all of the Gibbs updates.

\subsection{EXPERIMENTAL DETAILS}
We performed some simple preprocessing steps before the experiment.  Casualty count fields for each log entry were binarized ($0$ versus $>0$).  The wounded \emph{wounded}/\emph{killed}/\emph{detained} fields were merged disjunctively into one casualty indicator field.  The \emph{Friendly} (i.e. U.S. military) and \emph{HostNation} (Iraq or Afghanistan) casualty indicators were combined into one field via disjunction.  For the Iraq dataset, there were some missing data issues that had to be addressed.  No data was available for the 5th month, which was removed.  Most regions had no data for the final year of the Iraq data, so this was also removed.  Finally, we removed the MND-S and MND-NE region codes from our analysis, as these regions had very little data.

\begin{figure*}[t]
\begin{minipage}{\linewidth}
\hspace{-0.5cm}
\centering
\includegraphics[scale=0.4]{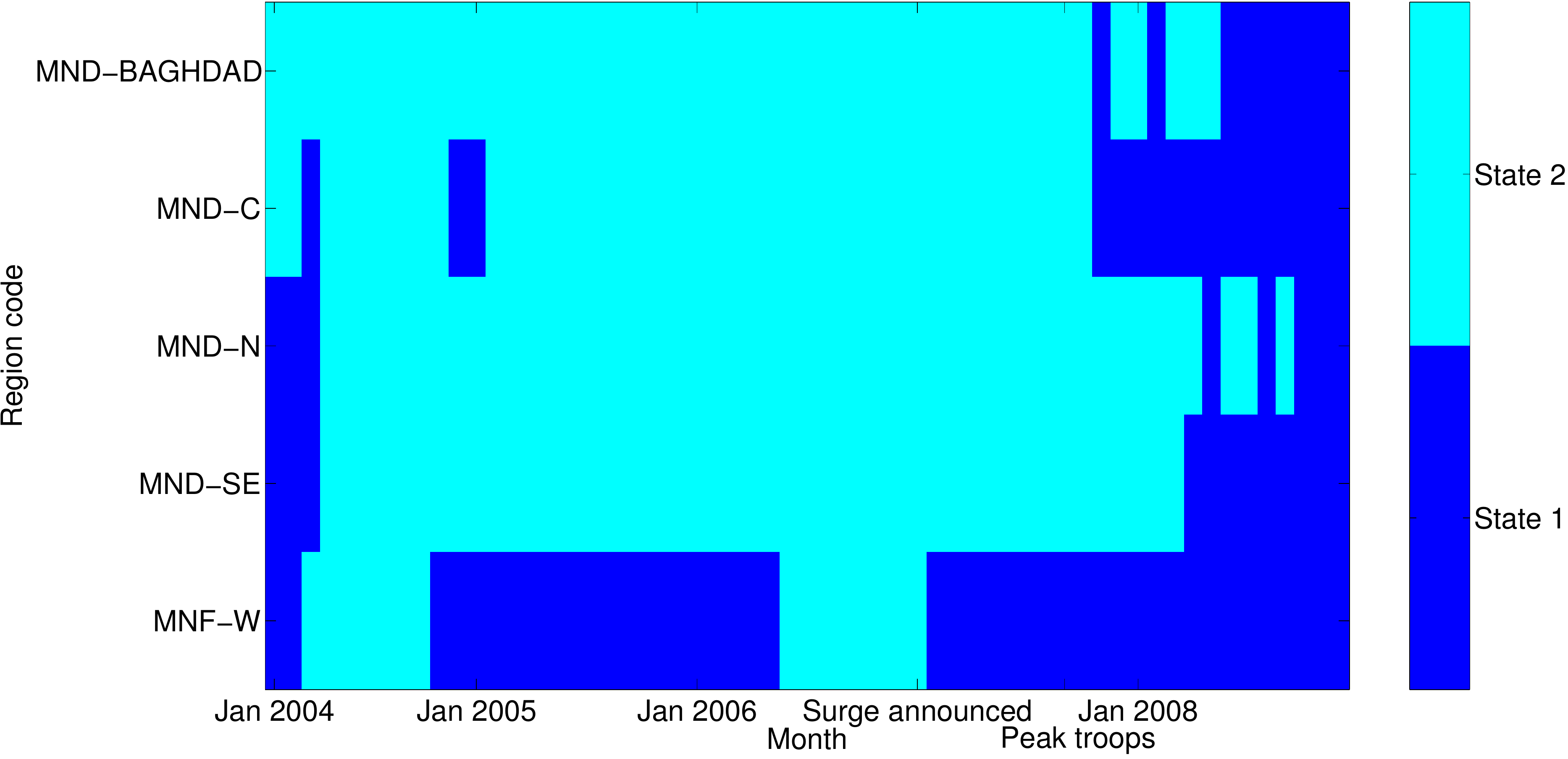}
\end{minipage}

\begin{minipage}{\linewidth}
\centering
\includegraphics[scale=0.4]{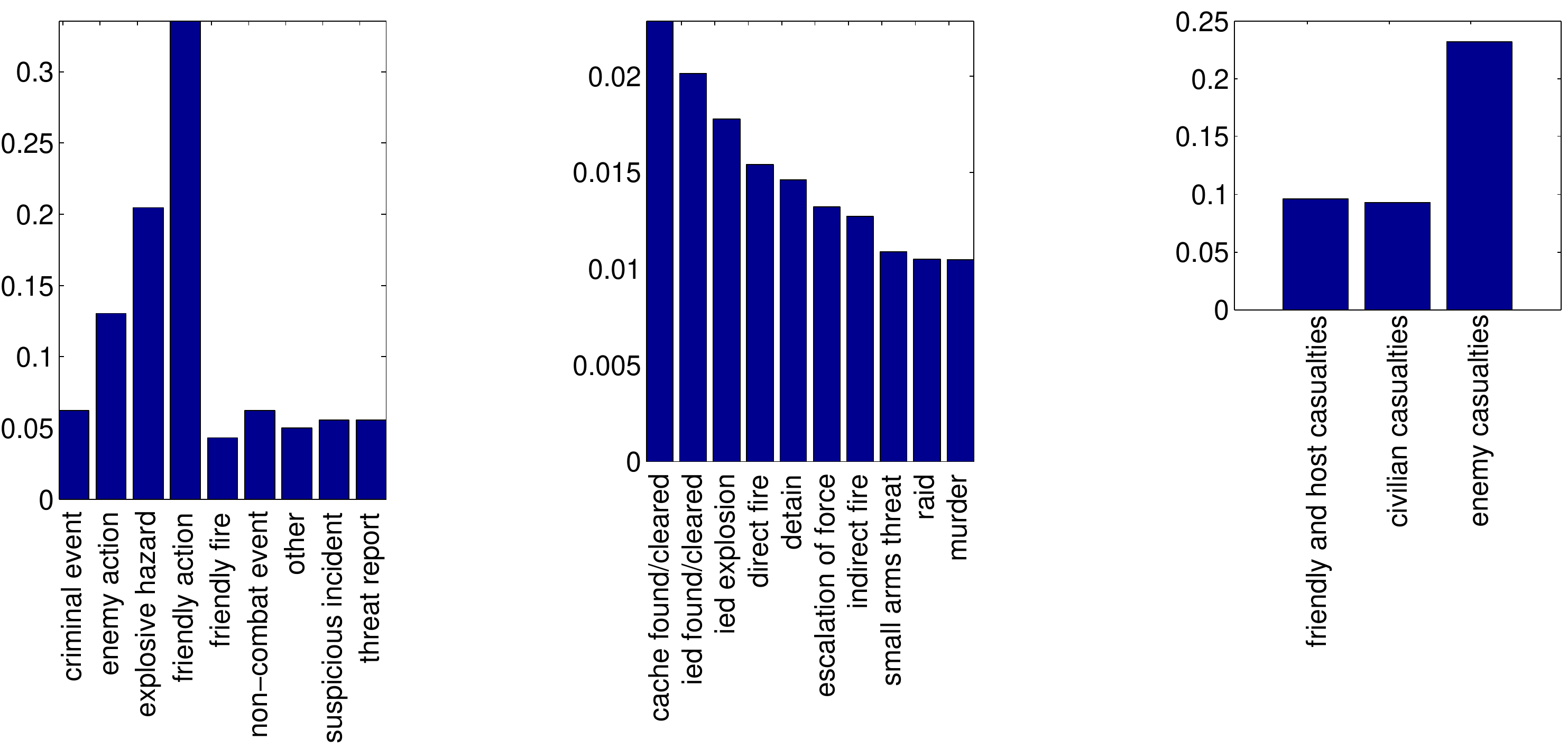}
\end{minipage}

\begin{minipage}{\linewidth}
\centering
\includegraphics[scale=0.4]{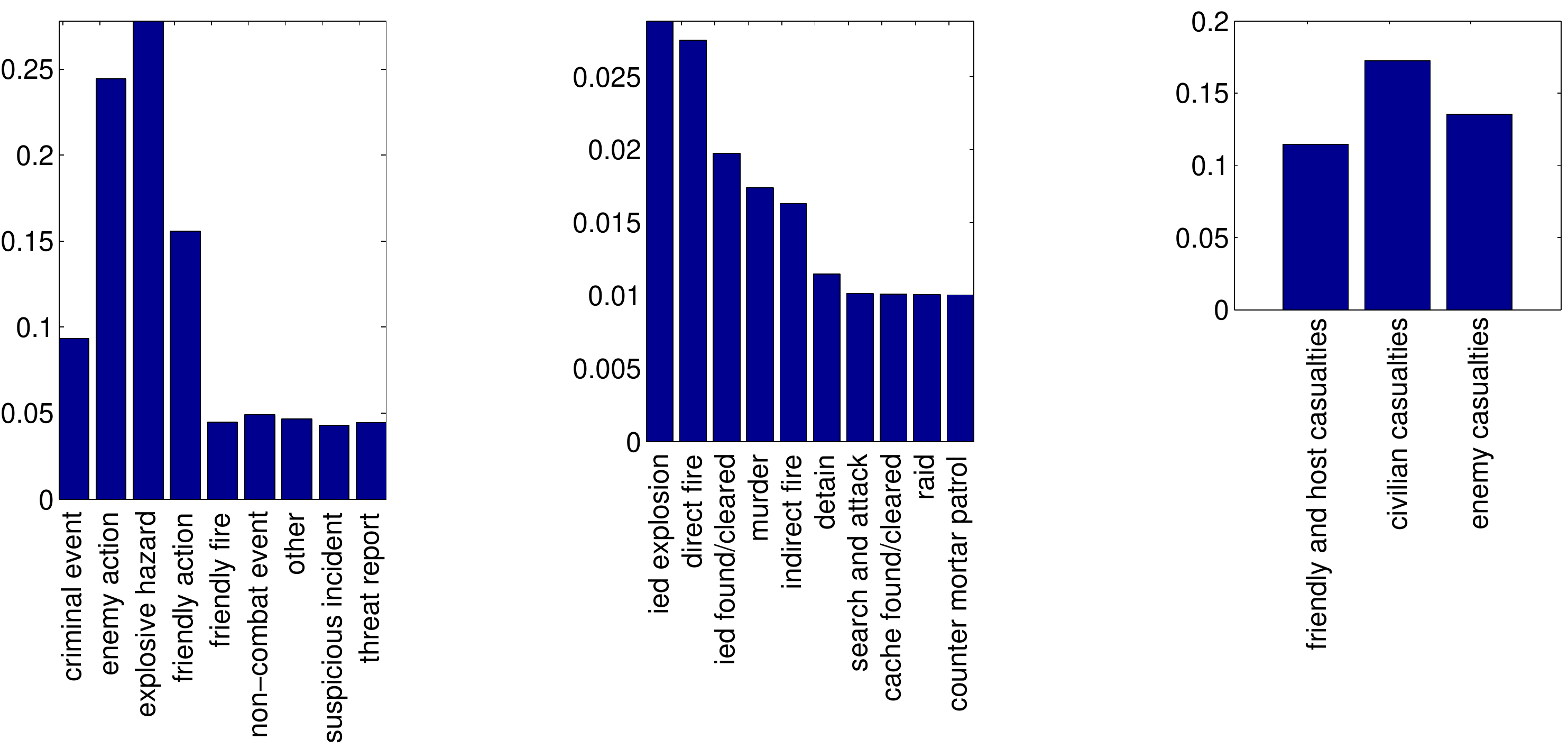}
\end{minipage}

\ifarxiv
\else
\caption{State assignments of privacy-preserving HMM on Iraq (Laplace mechanism, $\epsilon = 5$) (\textbf{Top}).\\ \textbf{Middle:} State 1. \textbf{Bottom:} State 2. \label{fig:LaplaceIraqClusters}}
\fi

\end{figure*}

\begin{figure*}[t]
\begin{minipage}{\linewidth}
\hspace{-1.3cm}
\includegraphics[scale=0.25]{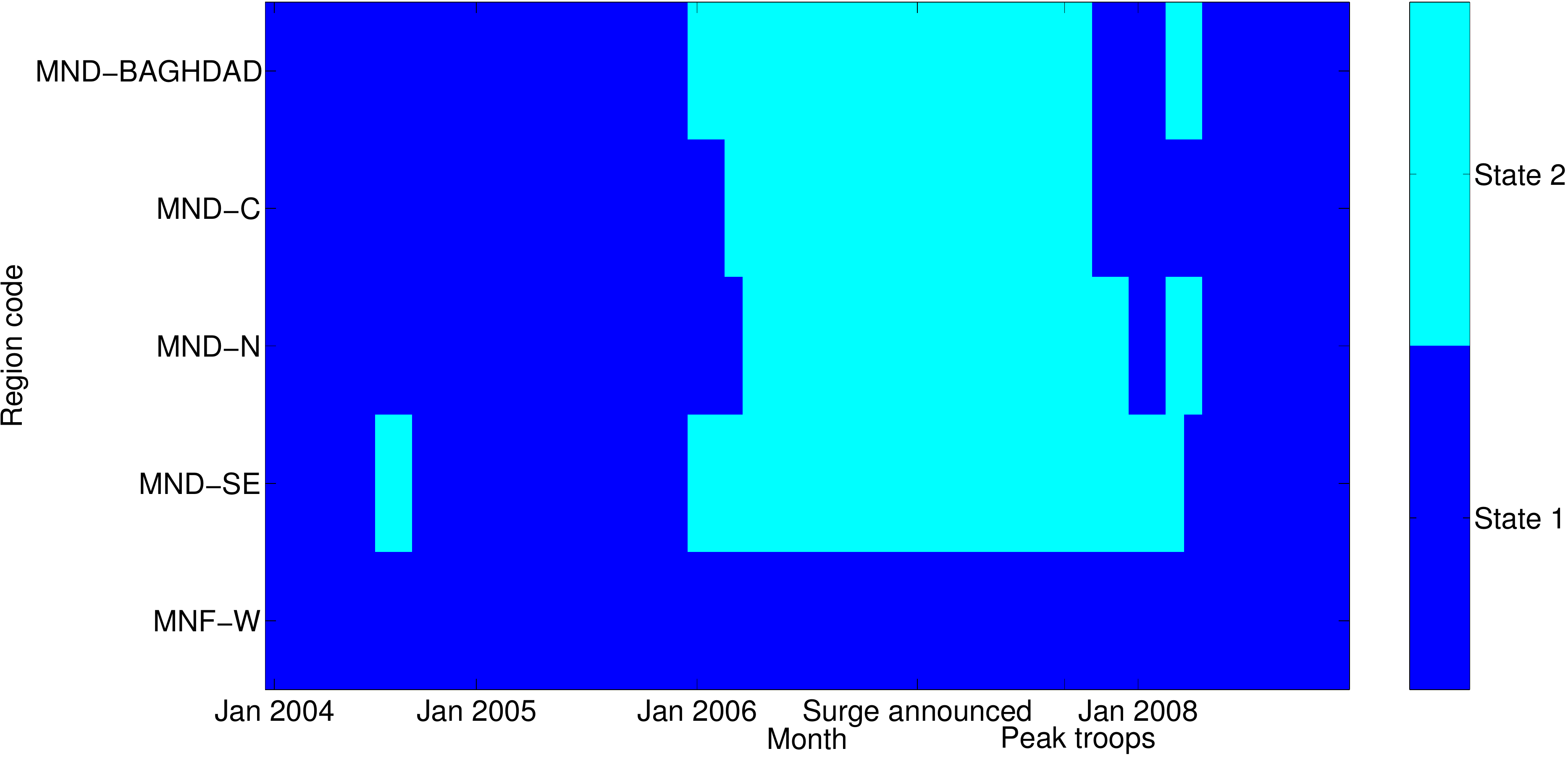}
\includegraphics[scale=0.25]{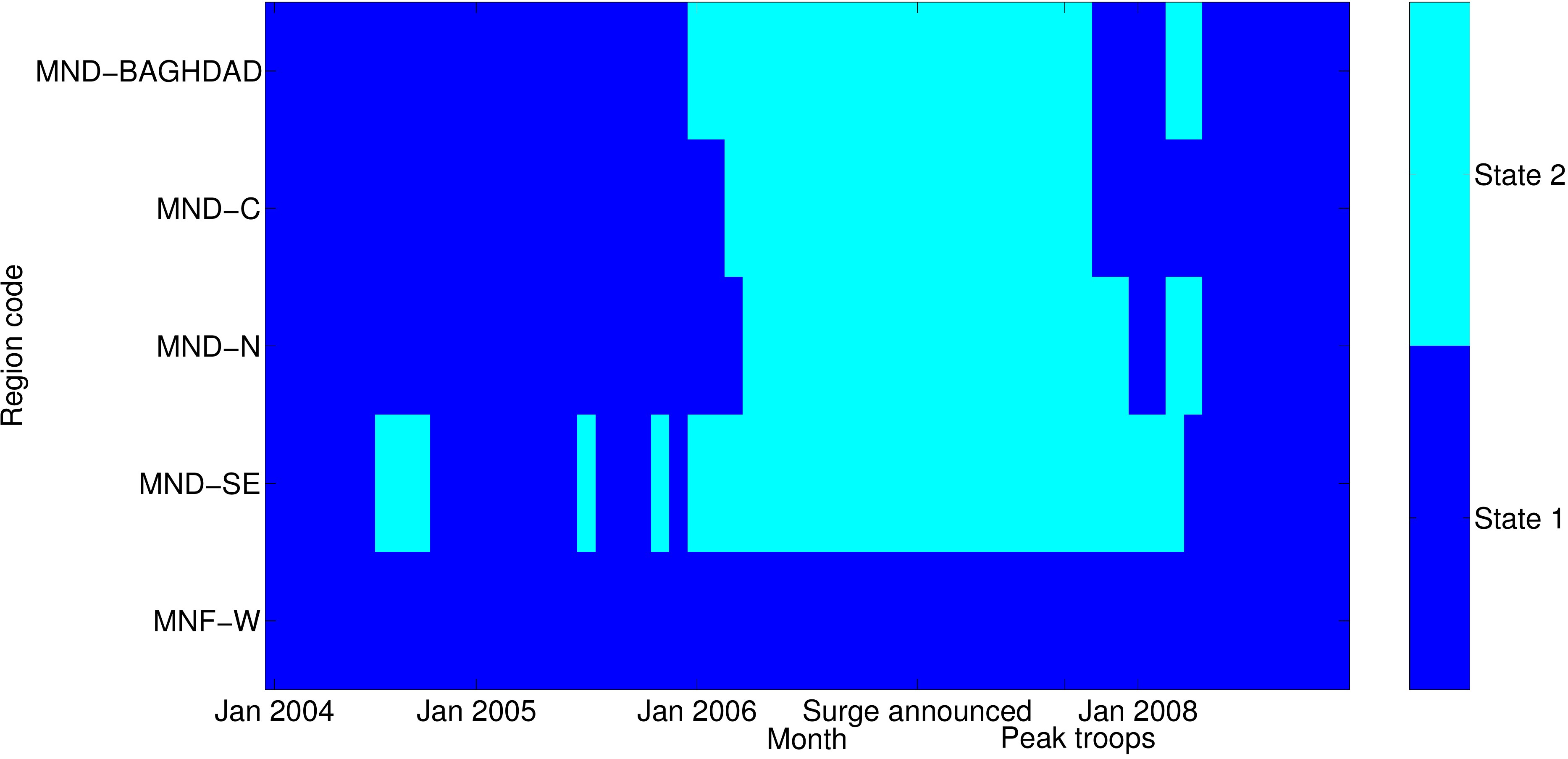}
\end{minipage}
\begin{minipage}{\linewidth}
\centering
\includegraphics[scale=0.4]{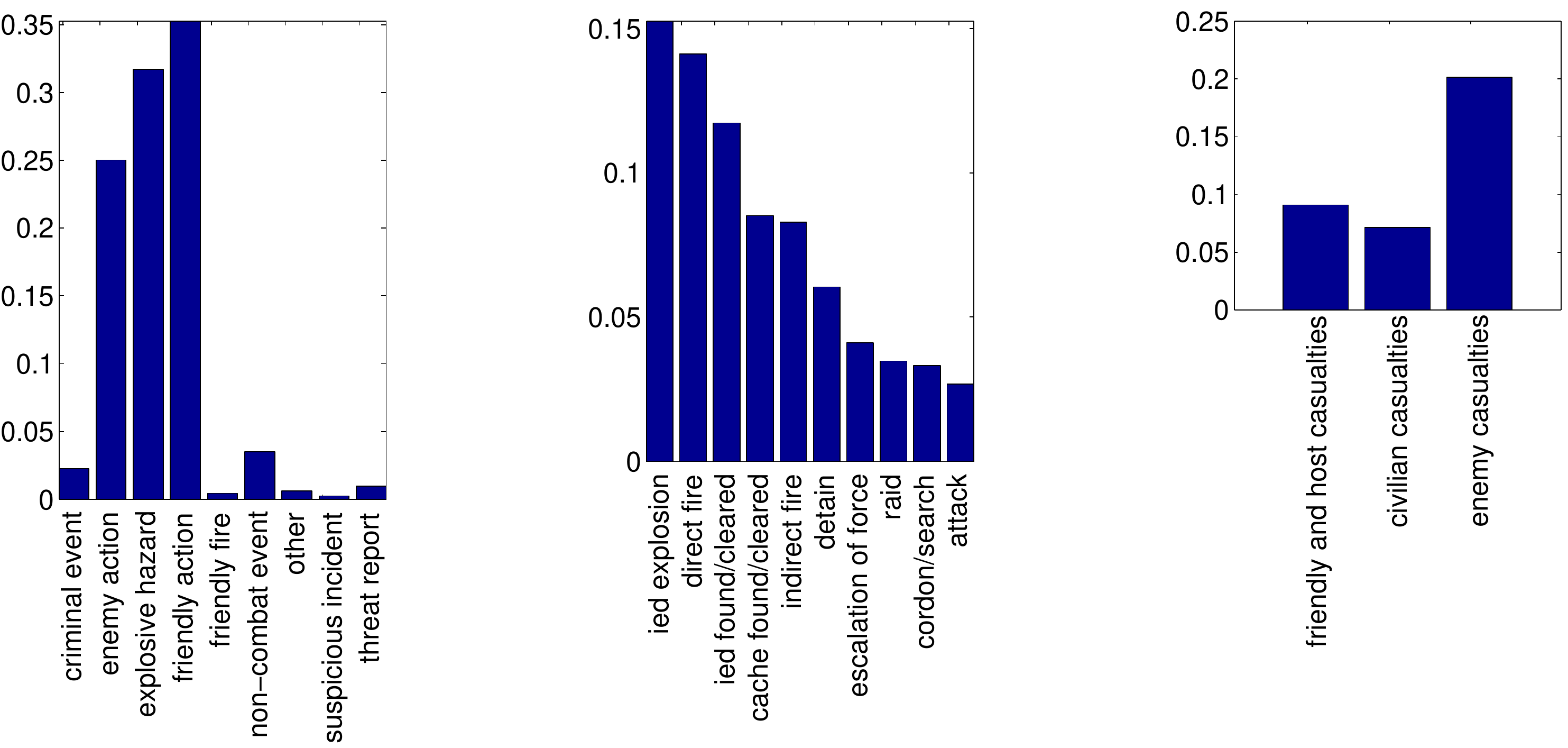}
\end{minipage}

\begin{minipage}{\linewidth}
\centering
\includegraphics[scale=0.4]{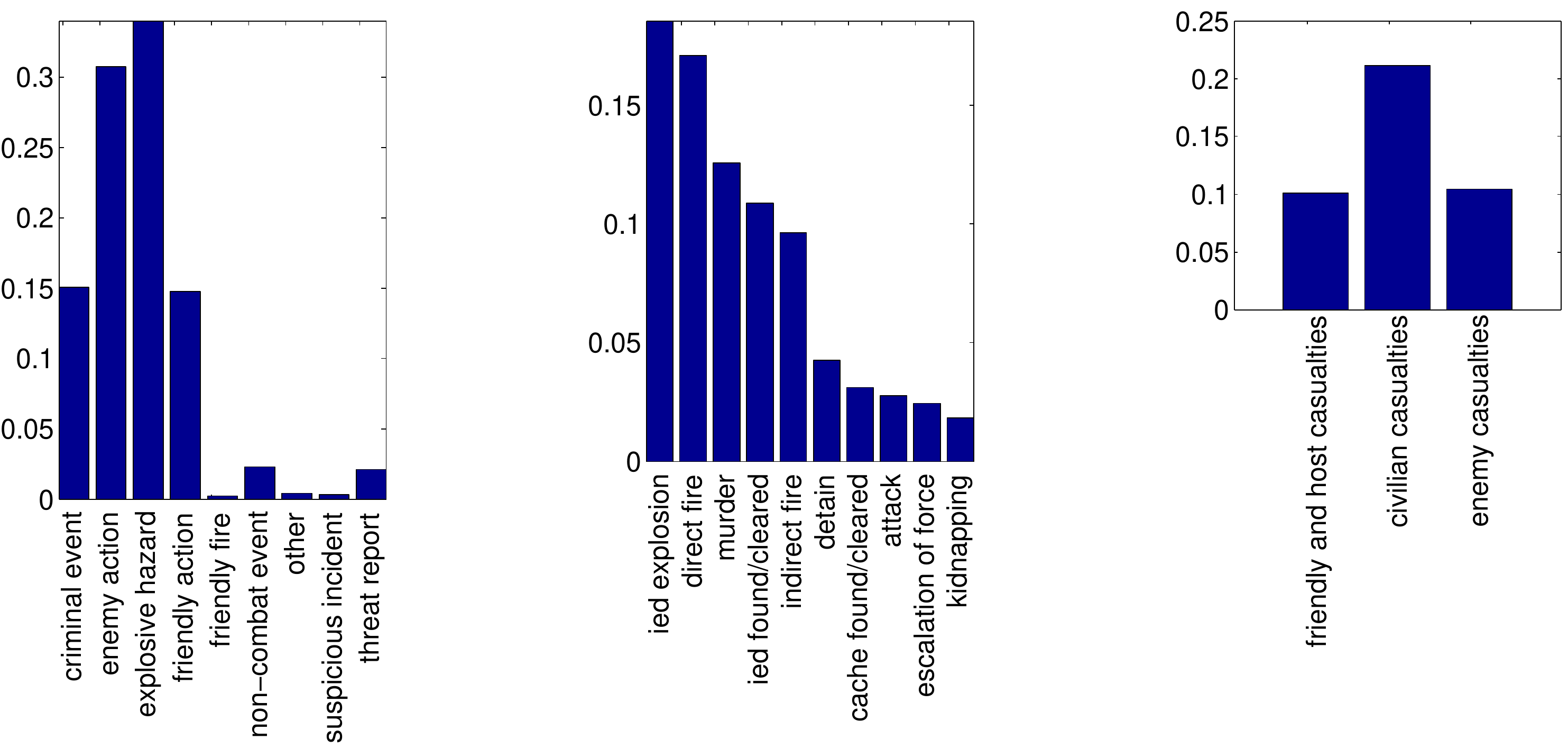}
\end{minipage}
\caption{State assignments and parameters for OPS privacy-preserving HMM on Iraq.   (OPS, $\epsilon = 5$, truncation point $a_0 = \frac{1}{100 K_d}$). \textbf{Top Left}: Estimate from last 100 samples. \textbf{Top Right}: Estimate from last one sample. \textbf{Middle}: State 1. \textbf{Bottom}: State 2.
 \label{fig:clusterAssignmentsIraqOPS}}
\end{figure*}

\begin{figure*}[t]
\begin{minipage}{\linewidth}
\centering
\includegraphics[scale=0.3]{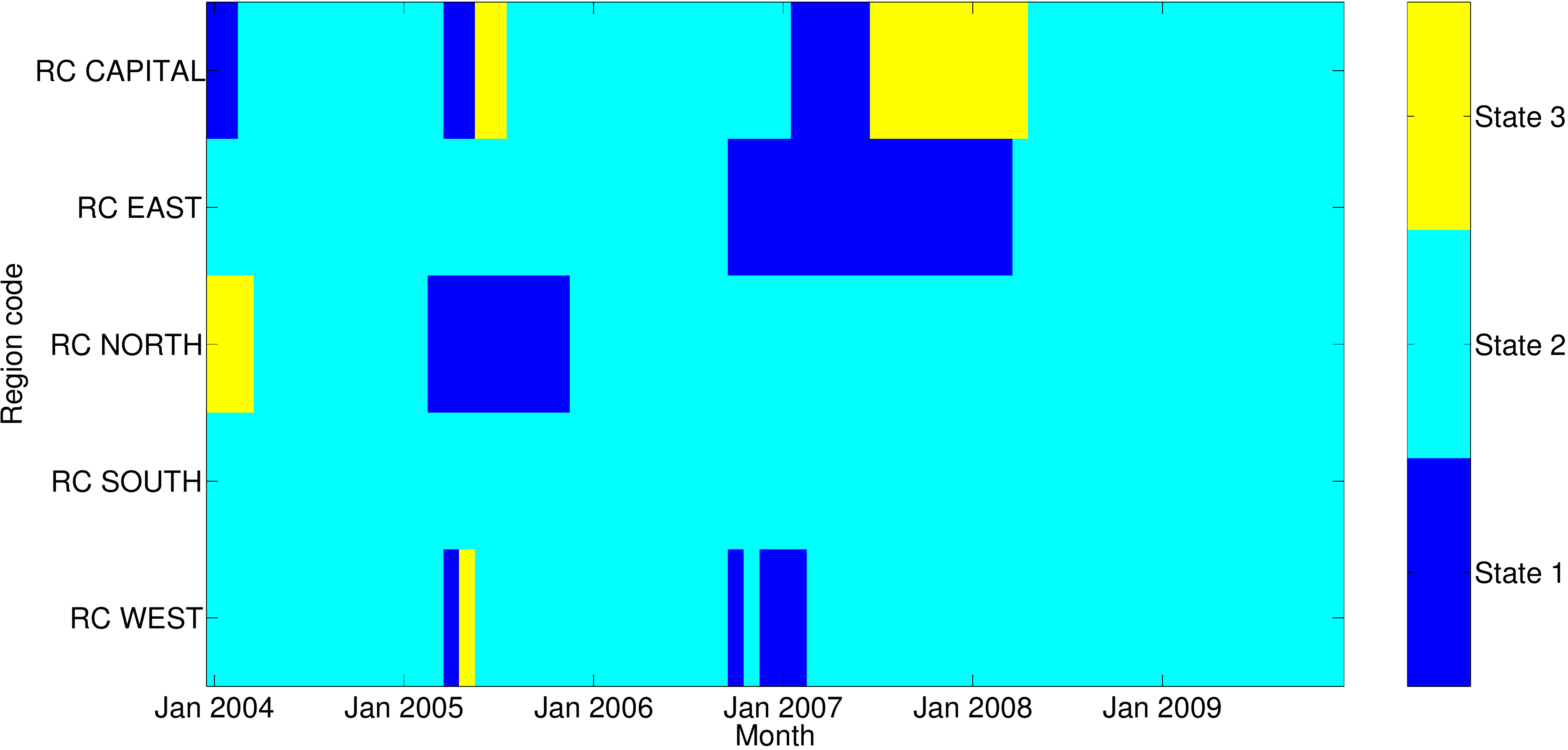}
\end{minipage}

\begin{minipage}{\linewidth}
\centering
\includegraphics[scale=0.35]{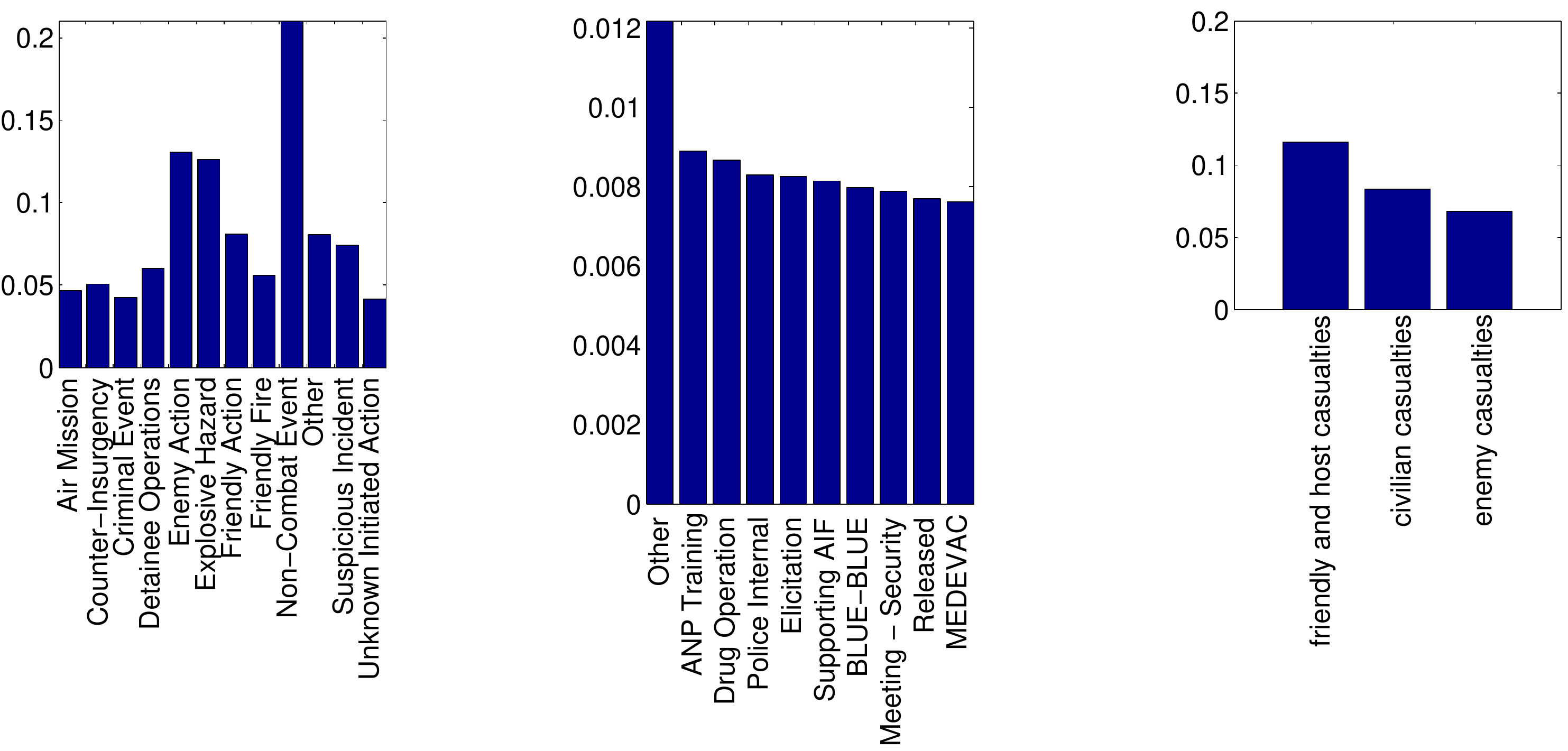}
\end{minipage}

\begin{minipage}{\linewidth}
\centering
\includegraphics[scale=0.35]{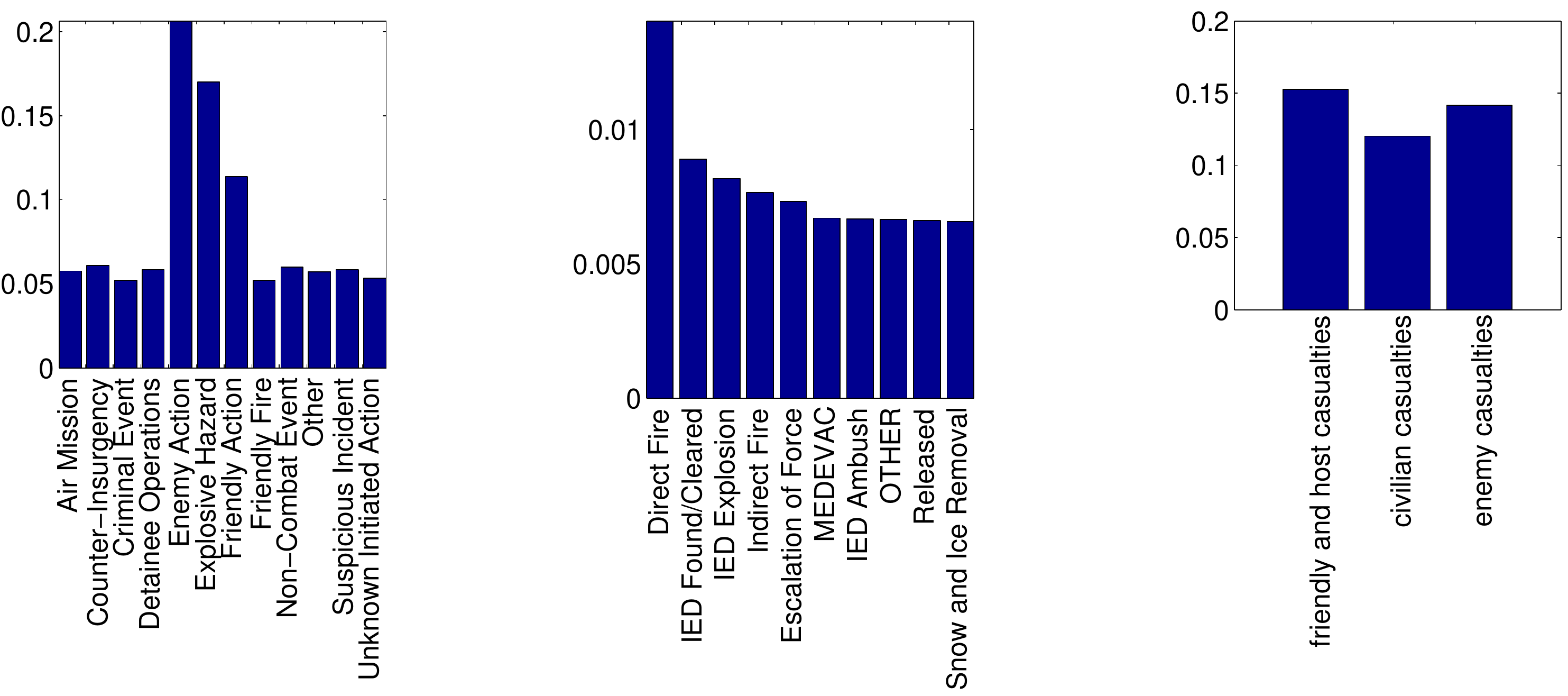}
\end{minipage}

\begin{minipage}{\linewidth}
\centering 
\includegraphics[scale=0.35]{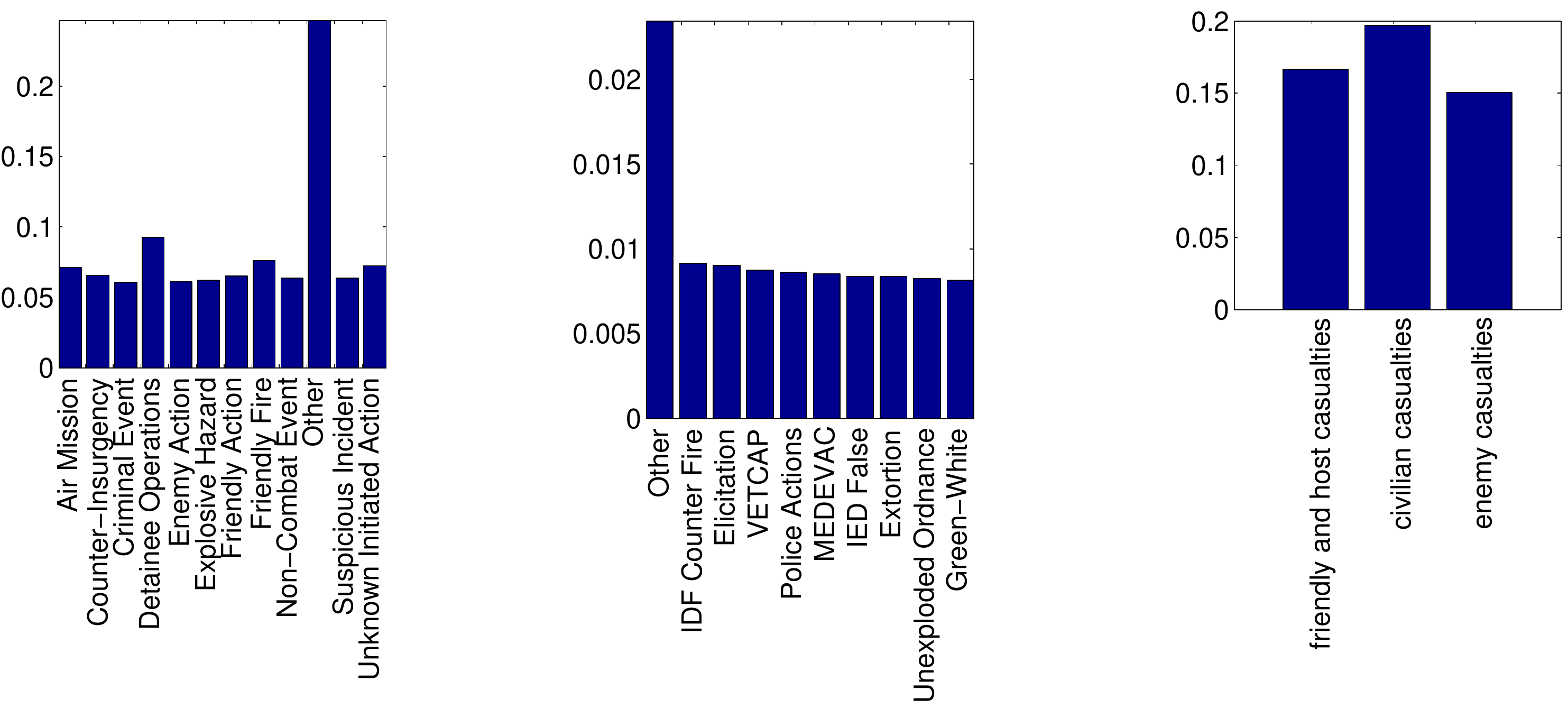}
\end{minipage}
\caption{State assignments of privacy-preserving HMM on Afghanistan (Laplace mechanism, $\epsilon = 5$) (\textbf{Top}). Parameters for States 1, 2, and 3, ordered from top to bottom. \label{fig:LaplaceAfghanistanClusters}}
\end{figure*}

\begin{figure*}[t]
\begin{minipage}{\linewidth}
\hspace{-1.3cm}
\includegraphics[scale=0.25]{figures/HMM/OPS/Afghanistan_cluster_assignments_largerFont}
\includegraphics[scale=0.25]{figures/HMM/OPS/Afghanistan_cluster_assignments_oneSample_largerFont}
\end{minipage}
\begin{minipage}{\linewidth}
\centering
\includegraphics[scale=0.35]{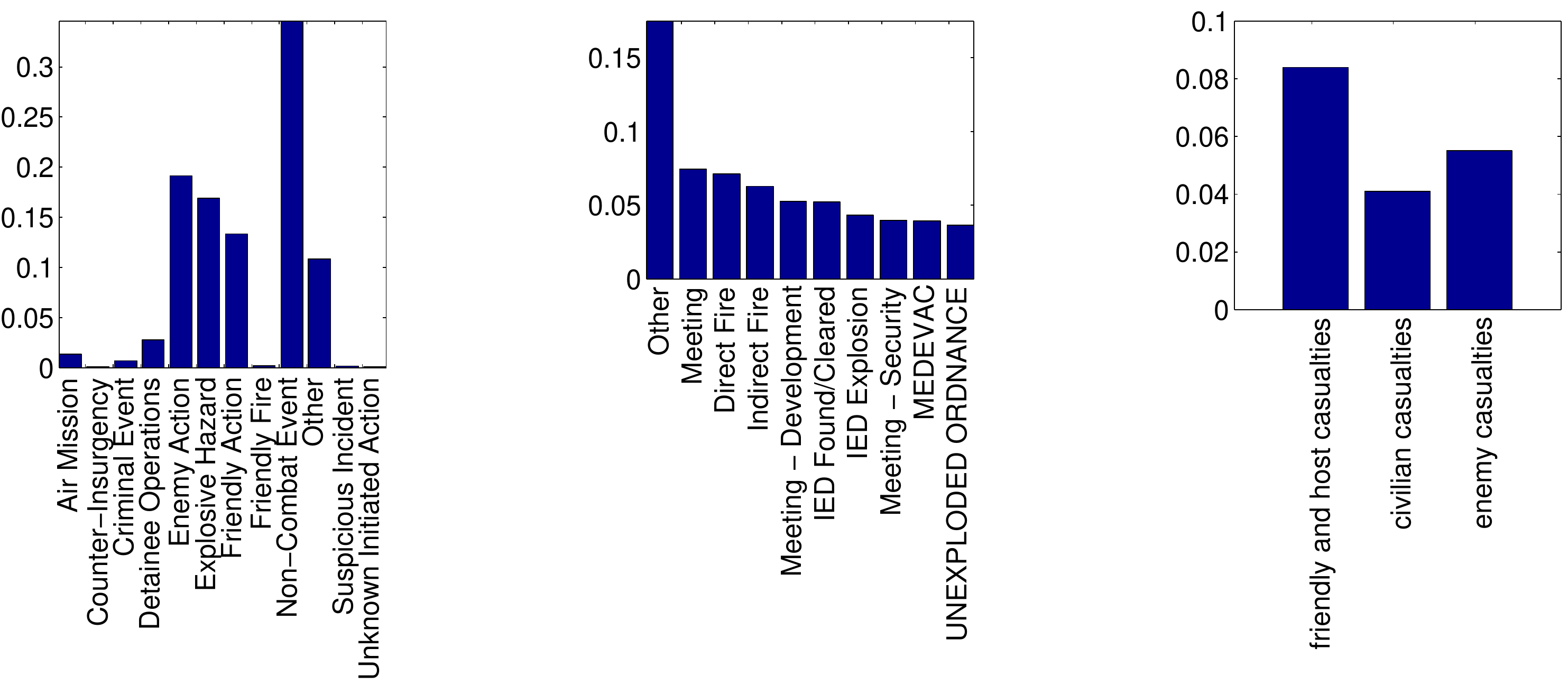}
\end{minipage}

\begin{minipage}{\linewidth}
\centering
\includegraphics[scale=0.35]{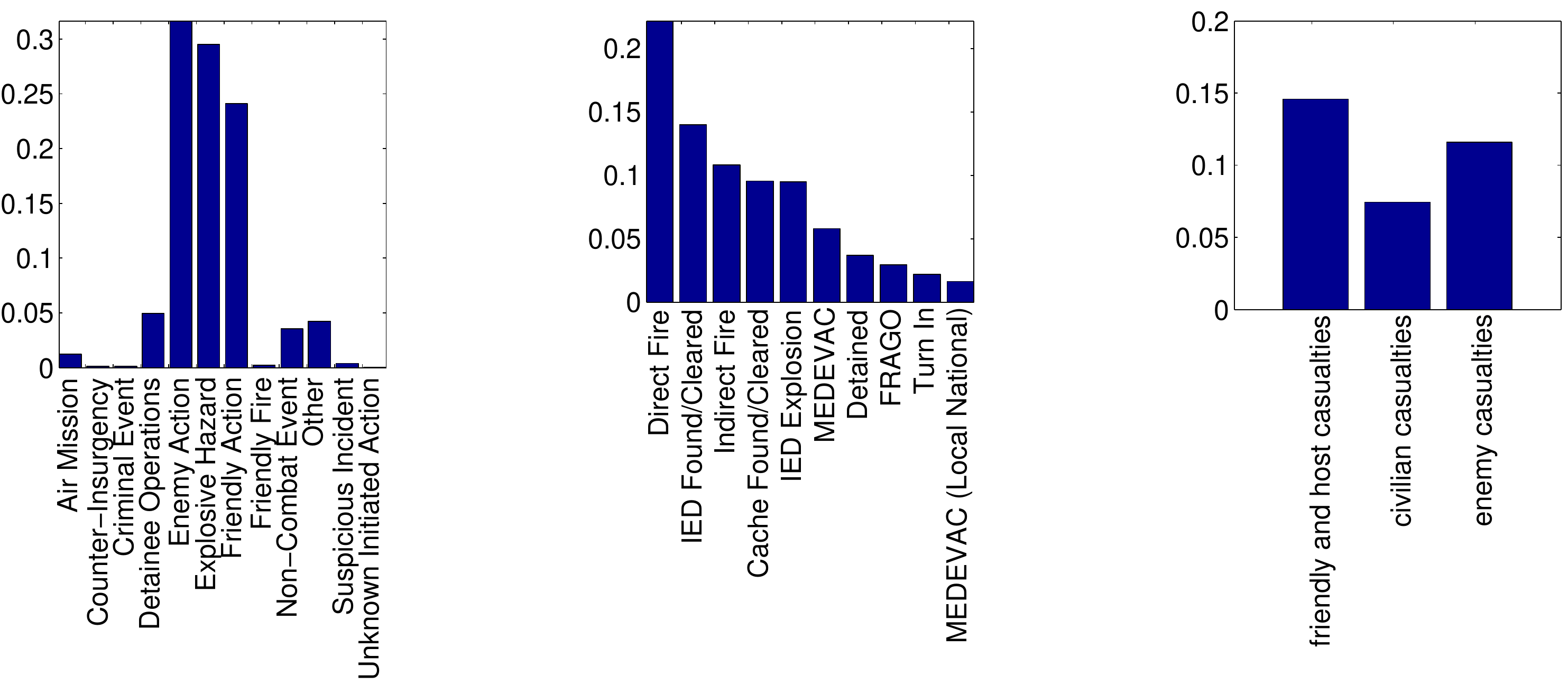}
\end{minipage}
\begin{minipage}{\linewidth}
\centering
\includegraphics[scale=0.35]{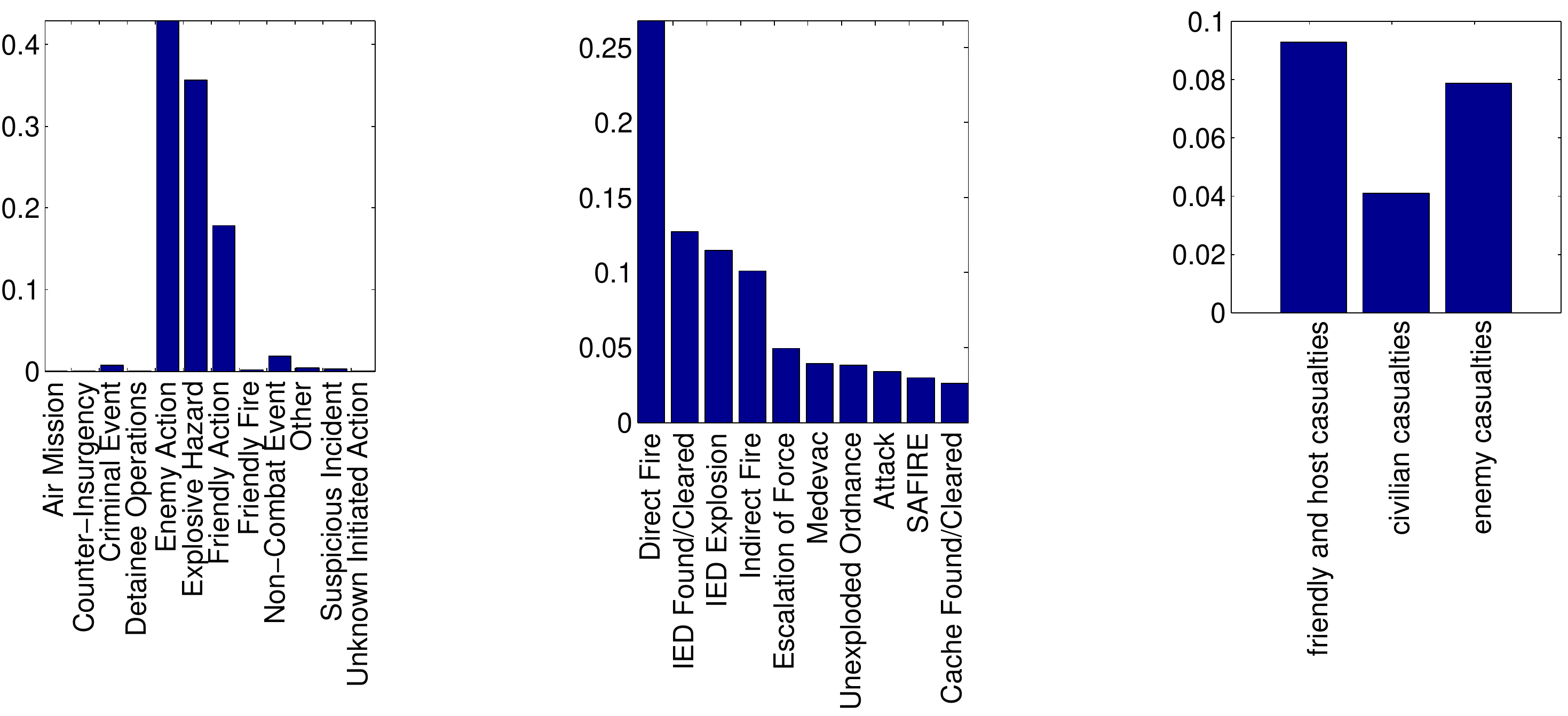}
\end{minipage}
\caption{State assignments and parameters for OPS privacy-preserving HMM on Afghanistan.   (OPS, $\epsilon = 5$, truncation point $a_0 = \frac{1}{100 K_d}$). \textbf{Top Left}: Estimate from last 100 samples. \textbf{Top Right}: Estimate from last one sample. Parameters for States 1, 2, and 3, ordered from top to bottom.
 \label{fig:clusterAssignmentsAfghanistanOPS}}
\end{figure*}

To simulate from truncated Dirichlet distributions for the Gibbs updates of the OPS method, we used the approach of \cite{fang2000statistical}, which involves sequentially drawing each component based on a truncated Beta distribution.  Full visualization results are shown in Figures \ref{fig:LaplaceIraqClusters} to \ref{fig:clusterAssignmentsAfghanistanOPS}.  Log-likelihood results on held-out data are given in Figure \ref{fig:naiveBayesLL}.  In this experiment, we randomly held-out 10\% of the region/timestep pairs for testing for each of 5 train/test splits, and reported the average log-likelihood over the repeats.

%% file: supplementaryExpFamilyProofs.tex
\subsection{PROOF OF LAPLACE MECHANISM ASYMPTOTIC KL-DIVERGENCE} \label{subsect:proofKLARE}

Our results hold specifically over the class of exponential families. A family of distributions parameterized by $\theta$ which has the form 

\begin{equation}
Pr(\mathbf{x}|\theta) = h(\mathbf{x}) \exp \Big (\theta^\intercal S(\mathbf{x}) - A(\theta) \Big )
\end{equation} 
is said to be an exponential family. Breaking down this structure into its parts, $\theta$ is a vector known as the natural parameters for the distribution and lies in some space $\Theta$. $S(\mathbf{x})$ represents a vector of sufficient statistics that fully capture the information needed to determine how likely $\mathbf{x}$ is under this distribution. $A(\theta)$ represents the log-normalizer, a term used to make this a probability distribution sum to one over all possibilities of $\mathbf{x}$. $h(\mathbf{x})$ is a base measure for this family, independent of which distribution in the family is used.

As we are interested in learning $\theta$, we are considering algorithms that generate a posterior distribution for $\theta$. The exponential families always have a conjugate prior family which is itself an exponential family. When speaking of these prior and posterior distributions, $\theta$ becomes the random variable and we introduce a new vector of natural parameters $\eta$ in a space $M$ to parameterize these distributions. To ease notation, we will express this conjugate prior exponential family as $Pr(\theta|\eta) = f(\theta) \exp \Big (\eta^\intercal T(\theta) - B(\eta) \Big )$, which is simply a relabelling of the exponential family structure.  The posterior from this conjugate prior is often written in an equivalent form

$Pr(\theta|\mathbf{X}, \chi, \alpha) \propto g(\theta)^{N + \alpha} \exp \Big (\theta^\intercal \big (\sum_{i=1}^N S(\mathbf{x}^{(i)}) + \alpha \chi \big) \Big ) \mbox{ ,}$

where the vector $\chi$ and the scalar $\alpha$ together specify the vector $\eta$ of natural parameters for this distribution. From the interaction of $\chi, \alpha,$ and $\mathbf{X}$ on the posterior, one can see that this prior acts like $\alpha$ observations with average sufficient statistics $\chi$ have already been observed. This parameterization with $\chi$ and $\alpha$ has many nice intuitive properties, but our proofs center around the natural parameter vector $\eta$ for this prior.

These two forms for the posterior can be reconciled by letting $\eta = (\alpha\chi + \sum_{i=1}^N S(\mathbf{x}^{(i)}), N + \alpha)$ and $T(\theta) = (\theta, -A(\theta))$. This definition for the natural parameters $\eta$ and sufficient statistics $T(\theta)$ fully specify the exponential family the posterior resides in, with $B(\eta)$ defined as the appropriate log-normalizer for this distribution (and $f(\theta) = 1$ is merely a constant). We note that the space of  $T(\Theta)$ is not the full space $\mathbb{R}^{d+1}$, as the last component of $T(\theta)$ is a function of the previous components. Plugging in these expressions for $\eta$ and $ T(\theta)$ we get the following form for the conjugate prior:

\begin{align}
Pr(\theta|\mathbf{X}, \chi, \alpha) &= \exp \Big ( \theta^\intercal(\alpha\chi + \sum_{i=1}^N S(\mathbf{x}^{(i)})) \nonumber\\
& - (N+\alpha)A(\theta) \nonumber\\
&- B(\eta) \Big ) \mbox{.}
\label{eq:conjprior}
\end{align}

We begin by defining minimal exponential families, a special class of exponential families with nice properties. To be minimal, the sufficient statistics must be linearly independent. We will later relax the requirement that we consider only minimal exponential families.

\begin{Defn}
An exponential family of distributions generating a random variable $\mathbf{x} \in \mathcal{X}$ with $S(\mathbf{x}) \in \mathbb{R}^d$ is said to be minimal if $\not \exists \phi \in \mathbb{R}^d, \phi \not = 0$ s.t. $\exists c \in \mathbb{R} $ s.t. $ \forall \mathbf{x} \in \mathcal{X} $ $\phi^\intercal S(\mathbf{x}) = c \mbox{.}$ \label{def:minimalexpfam}
\end{Defn}

Next we present a few simple algebraic results of minimal exponential families.

\begin{Lem}
\label{lem:expfamkl} 
For two distributions $p,q$ from the same minimal exponential family, 

\begin{equation}
KL(p||q) = A(\theta_q) - A(\theta_p) - (\theta_q - \theta_p) ^\intercal \nabla A(\theta_p) \label{eq:expfamkl}
\end{equation}

where $\theta_p,\theta_q$ are the natural parameters of $p$ and $q$, and $A(\theta)$ is the log-normalizer for the exponential family.
\end{Lem}

\begin{Lem}
\label{lem:expfammoments} 
A minimal exponential family distribution satisfies these equalities:
$$
\nabla A(\theta) = E_{Pr(\mathbf{x}|\theta)}[S(\mathbf{x})]
$$

$$
\nabla^2 A(\theta) = cov_{Pr(\mathbf{x}|\theta)}(S(\mathbf{x})) \mbox{ . }
$$
\end{Lem}

\begin{Lem}
\label{lem:expfamconvex} 
For a minimal exponential family distribution, its log-normalizer $A(\theta)$ is a strictly convex function over the natural parameters. This implies a bijection between $\theta$ and $E_{Pr(\mathbf{x}|\theta)}[S(\mathbf{x})]$.
\end{Lem}

These are standard results coming from some algebraic manipulations as seen in \citep{brown1986expfam}, and we omit the proof of these lemmas. Lemma \ref{lem:expfamconvex} immediately leads to a useful corollary about minimal families and their conjugate prior families.

\begin{Cor}
For a minimal exponential family distribution, the conjugate prior family given in equation \eqref{eq:conjprior} is also minimal.
\end{Cor}

\begin{proof2}

$T(\theta) = (\theta,-A(\theta))$ forms the sufficient statistics for the conjugate prior. Since $A(\theta)$ is strictly convex, there can be no linear relationship between the components of $\theta$ and $A(\theta)$. Definition \ref{def:minimalexpfam} applies.
\end{proof2}

Our next result looks at sufficient conditions for getting a KL divergence of 0 in the limit when adding a finite perturbance vector $\gamma$ to the natural parameters. The limit is taken over $N$, which will later be tied to the amount of data used in forming the posterior. As we now discuss posterior distributions also forming exponential families, our natural parameters will now be denoted by $\eta$ and the random variables are now $\theta$.

\begin{Lem}
\label{lem:expfamkltaylor}
Let $p(\theta | \eta)$ denote the distribution from an exponential family of natural parameter $\eta$, and let $\gamma$ be a constant vector of the same dimensionality as $\eta$, and let $\eta_N$ be a sequence of natural parameters. If for every $\zeta$ on the line segment connecting $\eta$ and $\eta+\gamma$ we have  the spectral norm $||\nabla^2B(\zeta)|| < D_N$ for some constant $D_N$, then

$$
KL(p(\theta|\eta_N + \gamma) || p(\theta|\eta_N)) \leq D_N||\gamma|| \mbox{ . }
$$
\end{Lem}

\begin{proof2}
This follows from noticing that equation \eqref{eq:expfamkl} in Lemma \ref{lem:expfamkl} becomes the first-order Taylor approximation of $B(\eta_N)$ centered at $B(\eta_N + \gamma)$. From Taylor's theorem, there exists $\alpha$ between $\eta_N$ and $\eta_N + \gamma$ such that $\frac{1}{2}\gamma^\intercal \nabla^2 B(\alpha) 
\gamma$ is equal to the error of this approximation. 

\begin{equation}
B(\eta_N) = B(\eta_N + \gamma) + (-\gamma) ^\intercal \nabla B(\eta_N + \gamma) + \frac{1}{2}\gamma^\intercal \nabla^2 B(\alpha) \gamma \label{eq:lognormtaylor}
\end{equation}

From rearranging equation \eqref{eq:expfamkl},

\begin{align}
B(\eta_N + \gamma) = B(\eta_N) - KL(p(\theta|\eta_N + \gamma) || p(\theta|\eta_N)) \nonumber\\
+ (\gamma)^\intercal \nabla B(\eta_n + \gamma) 
\end{align}

Using this substitution in \eqref{eq:lognormtaylor} gives

\begin{equation}
B(\eta_N) = B(\eta_N) - KL(p(\theta|\eta_N + \gamma) || p(\theta|\eta_N)) + \frac{1}{2}\gamma^\intercal \nabla^2 B(\alpha) \gamma \mbox { . }
\end{equation}

Solving for $KL(p(\theta|\eta_N + \gamma) || p(\theta|\eta_N))$ then gives the desired result:

$$
KL(p(\theta|\eta_N + \gamma) || p(\theta|\eta_N))  =   \frac{1}{2}\gamma^\intercal \nabla^2 B(\alpha) \gamma  \leq D_N||\gamma|| \mbox { . }
$$

\end{proof2}

This provides the heart of our results: If $||\nabla^2B(\zeta)||$ is small for all $\zeta$ connecting $\eta$ and $\eta+\gamma$, then we can conclude that $KL(p(\theta|\eta_N + \gamma) || p(\theta|\eta_N))$ is small with respect to $||\gamma||$. We wish to show that for $\eta_N$ arising from observing $N$ data points we have $D_N$ approaching 0 as $N$ grows.  To achieve this, we will analyze a relationship between the norm of the natural parameter $\eta$ and the covariance of the distribution it parameterizes. This relationship shows that posteriors with plenty of observed data have low covariance over $T(\theta)$, which permits us to use Lemma \ref{lem:expfamkltaylor} to bound the KL divergence of our perturbed posteriors. Before we reach this relationship, first we prove that our posteriors have a well-defined mode, as our later relationship will require this mode to be well-behaved.

\begin{Lem}
\label{lem:modeexists}

Let $Pr(\mathbf{x}|\theta) = h(\mathbf{x}) \exp \Big (\theta^\intercal S(\mathbf{x}) - A(\theta) \Big )$ be a likelihood function for $\theta$ and let there be a conjugate prior $Pr(\theta|\eta) = f(\theta) \exp \Big (\eta^\intercal T(\theta) - B(\eta) \Big )$, where both distributions are minimal exponential families. Let $M$ be the space of natural parameters $\eta$, and $\Theta$ be the space of $\theta$. Furthermore, assume $\eta$ is the parameterization arising from the natural conjugate prior, such that $\eta = (\alpha\chi,\alpha)$. If the following conditions hold:

\begin{enumerate}
\item $\eta$ is in the interior of $M$
\item $\alpha > 0$
\item $A(\theta)$ is  a real, continuous, and differentiable
\item $B(\eta)$ exists, the distribution $Pr(\theta|\eta)$ is normalizable.
\end{enumerate}

then 

$$
\argmax_{\theta \in \Theta} {\eta^\intercal T(\theta)} = \theta^*_\eta
$$ 

is a well-defined function of $\eta$, and $\theta^*_\eta$ is in the interior of $\Theta$.

\end{Lem}

\begin{proof2}

Using our structure for the conjugate prior from \eqref{eq:conjprior}, we can expand the expression $\eta^\intercal T(\theta)$.

$$
\eta^\intercal T(\theta) = \alpha\chi^\intercal \theta - \alpha A(\theta)
$$

We note that the first term is linear in $\theta$, and that by minimality and Lemma \ref{lem:expfamconvex}, $A(\theta)$ is strictly convex. This implies $\eta^\intercal T(\theta)$ is strictly concave over $\theta$. Thus any interior local maximum must also be the unique global maximum.

The gradient of with $\eta^\intercal T(\theta)$ respect to $\theta$ is simple to compute.

$$
\nabla(\eta^\intercal T(\theta)) = \alpha\chi^\intercal  - \alpha \nabla A(\theta) 
$$

This expression can be set to zero, and solving for $\theta^*_\eta$ shows it must satisfy 

\begin{equation}
\nabla A(\theta^*_\eta) = \chi \mbox{ .}\label{eq:themode}
\end{equation}

We remark by Lemma \ref{lem:expfammoments} that $\nabla A(\theta^*_\eta)$ is equal to  $E_{Pr(\mathbf{x}|\theta^*_\eta)}[S(\mathbf{x})]$, and so this is the $\theta$ that generates a distribution with mean $\chi$.

By the strict concavity, this is sufficient to prove $\theta^*_\eta$ is a unique local maximizer and thus the global maximum. 

To see that $\theta^*_\eta$ must be in the interior of $\Theta$, we use the fact that $A(\theta)$ is continuously differentiable. This means $\nabla A(\theta)$ is a continuous function of $\theta$. Since $\eta$ is in the interior of $M$, we can construct an open neighborhood around $\chi$. The preimage of an open set under a continuous function is also an open set, so this implies an open neighborhood exists around $\theta^*_\eta$.

\end{proof2}

Now that we know $\theta^*_\eta$ is well defined for $\eta$ in the interior of $M$, we can express our relationship on high magnitude posterior parameters and the covariance of the distribution over $T(\theta)$ they generate.

\begin{Lem}
\label{lem:covargoesdown}

Let $Pr(\mathbf{x}|\theta) = h(\mathbf{x}) \exp \Big (\theta^\intercal S(\mathbf{x}) - A(\theta) \Big )$ be a likelihood function for $\theta$ and let there be a conjugate prior $Pr(\theta|\eta) = f(\theta) \exp \Big (\eta^\intercal T(\theta) - B(\eta) \Big )$, where both distributions are minimal exponential families. Let $M$ be the space of natural parameters $\eta$, and $\Theta$ be the space of $\theta$. Furthermore, assume $\eta$ is the parameterization arising from the natural conjugate prior, such that $\eta = (\alpha\chi,\alpha)$.

If $\exists \eta_0, \delta_1 > 0, \delta_2 > 0$ such that the conditions of Lemma \ref{lem:modeexists} hold for $\eta \in \mathcal{B}(\eta_0,\delta_1)$, and we have these additional assumptions,

\begin{enumerate}
\item the cone $\{k\eta' | k > 1, \eta' \in \overline{\mathcal{B}(\eta_0,\delta_1)}\}$ lies entirely in $M$
\item $A(\theta)$ is differentiable of all orders
\item $\exists P$ s.t. $\forall \theta \in \cup_{\eta'\in \overline{\mathcal{B}(\eta_0,\delta_1)}} \mathcal{B}(\theta^*_{\eta'},\delta_2)$  all partial derivatives up to order 7 of $A(\theta)$ have magnitude bounded by $P$ \label{cond:covargoesdown_boundedderiv}
\item $\exists w>0$ such that $\forall \theta \in \cup_{\eta'\in \overline{\mathcal{B}(\eta_0,\delta_1)}} \mathcal{B}(\theta^*_{\eta'},\delta_2)$ we have $\det(\nabla^2 A(\theta)) > w$ \label{cond:covargoesdown_determinant}

\end{enumerate}

then there exists $C,K$ such that for $k > K$ the following bound holds $\forall \eta \in \mathcal{B}(\eta_0,\delta_1)$:

$||cov(T(\theta) | k\eta)|| < \frac{C}{k} \mbox {.}$

\end{Lem} 

\begin{proof2}

This result follows from the Laplace approximation method for $B(\eta) = \int_\Theta e^{\eta^\intercal T(\theta)} d\theta$. The inner details of this approximation are show in Lemma \ref{lem:laplaceapprox}. Here we show that our setting satisfies all the regularity assumptions for this approximation. First we define functions $s(\theta,\eta)$ and $F_k(\eta)$.

\begin{align}
s(\theta,\eta) &= \eta^\intercal T(\theta) = \alpha\chi^\intercal \theta - \alpha A(\theta) \\
F_k(\eta) &= B(k\eta) \nonumber\\
&= \int_\Theta e^{k\eta^\intercal T(\theta)} d\theta \nonumber\\
&= \int_\Theta e^{ks(\theta,\eta)} d\theta
\end{align}

With these definitions, we may now begin to check the assumptions of Lemma \ref{lem:laplaceapprox} hold. We copy these assumptions below, with a substitution of $\theta$ for $\phi$ and $\eta$ for $Y$. The full details of Lemma \ref{lem:laplaceapprox} can be found at the end of section \ref{subsect:proofKLARE}.

\begin{enumerate}
\item $\phi^*_Y = \argmax_{\phi \in M} s(\phi,Y) = g(Y)$, a function of $Y$.
\item $\phi^*_{Y'}$ is in the interior of $M$ for all $Y' \in \mathcal{B}(Y_0,\delta_1)$.
\item $g(Y)$ is continuously differentiable over the neighborhood $\mathcal{B}(Y_0,\delta_1)$. \label{cond:approx_lipschitz}
\item $s(\phi,Y')$ has derivatives of all orders for $Y' \in \mathcal{B}(Y_0,\delta_1)$,$\phi \in \mathcal{B}(\phi^*_{Y'},\delta_2)$ and all partial derivatives up to order 7 are bounded by some constant $P$ on this neighborhood. \label{cond:approx_deriv}
\item $\exists w>0$ such that $\forall Y' \in \mathcal{B}(Y_0,\delta_1),\forall \phi \in \mathcal{B}(\phi^*_{Y'},\delta_2)$ we have $det(\nabla^2_\phi s(\phi,Y)) > w$. \label{cond:approx_eigen}
\item $F_1(Y')$ exists for $Y' \in \mathcal{B}(Y_0,\delta_1)$, the integral is finite.
\end{enumerate}

We now show these conditions hold one-by-one. Let $\eta$ denote an arbitrary element of $B(\eta_0,\delta)$.

\begin{enumerate}
\item $\theta^*_\eta$ is a well-defined function (Lemma \ref{lem:modeexists}).
\item $\theta^*_{\eta}$ is in the interior of $\Theta$ (Lemma \ref{lem:modeexists}).
\item $g(\eta)$ follows the inverse of $\nabla A(\theta) : \mathbb{R}^d \rightarrow \mathbb{R}^d$. This vector mapping has a Jacobian $\nabla^2 A(\theta)$ which assumption \ref{cond:covargoesdown_determinant} guarantees has non-zero determinant on this neighborhood. This satisfies the Inverse Function Theorem to show $g(\eta)$ is continuously differentiable.
\item $s(\theta,\eta)$ has derivatives of all orders, and are suitably bounded as $s$ is composed of a linear term and the differentiable function $A(\theta)$, where we have bounded the derivatives of $A(\theta)$.
\item Assumption \ref{cond:covargoesdown_determinant} from this lemma translates directly.
\item $F_1(\eta) = B(\eta)$ which exists by virtue of $\eta$ being in the space of valid natural parameters.
\end{enumerate}

This completes all the requirements of Lemma \ref{lem:laplaceapprox}, which guarantees the existence of $C$ and $K$ such that for any $k>K$ and any $\eta \in \mathcal{B}(\eta_0,\delta_1)$, if we let $\psi$ denote $k\eta$, we have:

$||\nabla^2_\psi B(\psi)|| = ||\nabla^2_\psi \log F_k(\psi/k)|| < \frac{C}{k} \mbox{.}$

We conclude by noting that $\nabla^2_\psi B(\psi)$ is the covariance of the posterior with parameterization $\psi  = k\eta$.

\end{proof2}

Now that all our machinery is in place, it remains to be seen under what conditions the posterior satisfies the conditions of the previous Lemmas, along with extending to the case where $\gamma$ is a random variable, and not just a fixed finite vector.

\begin{Lem}
\label{lem:laplacekl}
For a minimal exponential family given a conjugate prior, where the posterior takes the form $Pr(\theta|\mathbf{X}, \chi, \alpha) \propto g(\theta)^{n + \alpha} \exp \Big (\theta^\intercal \big (\sum_{i=1}^n S(\mathbf{x}^{(i)}) + \alpha \chi \big) \Big ) $, where $p(\theta|\eta)$ denotes this posterior with a natural parameter vector $\eta$, if there exists a $\delta > 0$ such that these assumptions are met:

\begin{enumerate}
\item the data $\mathbf{X}$ comes i.i.d. from a minimal exponential family distribution with natural parameter $\theta_0 \in \Theta$
\item $\theta_0$ is in the interior of $\Theta$ 
\item the function $A(\theta)$ has all derivatives for $\theta$ in the interior of $\Theta$
\item $cov_{Pr(\mathbf{x}|\theta)}(S(\mathbf{x})))$ is finite for $\theta \in \mathcal{B}(\theta_0,\delta)$ 
\item $\exists w > 0$ s.t. $\det(cov_{Pr(\mathbf{x}|\theta)}(S(\mathbf{x})))) > w$ for $\theta \in \mathcal{B}(\theta_0,\delta)$ \label{cond:suffstatcovdet}
\item the prior $Pr(\theta|\chi,\alpha)$ is integrable and has support on a neighborhood of $\theta^*$
\end{enumerate}

then for any mechanism generating a perturbed posterior $\tilde{p}_N = p(\theta|\eta_N + \gamma)$ against a noiseless posterior $p_N = p(\theta|\eta_N)$ where $\gamma$ comes from a distribution that does not depend on the number of data observations $N$ and has finite covariance, this limit holds:

$\lim_{N \rightarrow \infty} E [ KL(\tilde{p}_N || p_N) ] = 0 \mbox { . }$

\end{Lem}

\begin{proof2}

We begin by fixing the randomness of the noise $\gamma$ that the mechanism will add to the natural parameters of the posterior.

We wish to show that the KL divergence goes to zero in the limit, which we will achieve by showing that for large enough data sizes, both the perturbed and unperturbed posteriors lie w.h.p. in a region where we can use Lemmas \ref{lem:expfamkltaylor} and \ref{lem:covargoesdown} apply.

To compute the posterior, after drawing a collection $\mathbf{X}$ of $N$ data observations, we compute the sum of the sufficient statistics and add them to the prior's parameters.

$$
\eta_N = \Big (\alpha\chi + \sum S(\mathbf{x}^{(i)}), \alpha + N \Big)
$$

$\eta_N$ is a random variable depending on the data observations $\mathbf{X}$. To analyze how it behaves, a couple related random variables will be defined, all implicitly conditioned on the constant $\theta_0$. Let $\mathbf{Y}$ denote a random variable matching the distribution of a single observation, and let $\mathbf{U}_N = \frac{1}{N}\sum S(\mathbf{x}^{(i)})$ which  has covariance $\frac{1}{N} cov(S(\mathbf{Y}))$. The expected value for $\mathbf{U}_N$ is of course $E[S(\mathbf{Y})]$.

By a vector version of the Chebyshev inequality for a random vector $\mathbf{U}$, \citep{chen2007new}

\begin{align}
Pr \Big ( (\mathbf{U} - E[\mathbf{U}])^\intercal (cov(\mathbf{U}))^{-1} (\mathbf{U} - E[\mathbf{U}]) \geq \nu \Big )\mbox{,} \nonumber\\
\leq \frac{d}{\nu} \mbox { , }
\end{align}

where $d$ is the dimensionality of $\mathbf{U}$. Using the spectral norm $||(cov(\mathbf{U}_N))^{-1}||$ and the $l_2$ norm $||\mathbf{U}_N - E[\mathbf{U}_N]||$ with some some rearrangement, we can show the following inequalities. We note that the covariance of $\mathbf{U}_N$ must be invertible, since the covariance of $\mathbf{Y}$ is invertible by assumption \eqref{cond:suffstatcovdet}.

\begin{equation}
Pr \Big ( ||\mathbf{U}_N - E[\mathbf{U}_N]|| \cdot ||(cov(\mathbf{U}_N))^{-1}||  \geq \nu \Big ) \leq \frac{d}{\nu}
\end{equation}

\begin{equation}
Pr \Big ( ||\mathbf{U}_N - E[\mathbf{U}_N]||   \geq \nu ||cov(\mathbf{U}_N)||\Big ) \leq \frac{d}{\nu}
\end{equation}

\begin{equation}
Pr \Big ( ||\mathbf{U}_N - E[S(\mathbf{Y})]||   \geq \frac{\nu}{N}||cov(\mathbf{Y})|| \Big ) \leq \frac{d}{\nu} \label{eq:whp}
\end{equation}

Thus for any $\epsilon> 0$, $\tau > 0$, there exists $N_{\epsilon,\tau}$  such that when the number of data observations $N$ exceeds $N_{\epsilon,\tau}$

\begin{equation}
Pr \big ( ||\mathbf{U}_{N} - E[\mathbf{Y}]||   \geq \epsilon \big ) \leq \tau \mbox{.} \label{eq:Uconc}
\end{equation}

We now define two modified vectors of natural parameters $\eta_a = \frac{\eta_N}{N} = (\mathbf{U}_N,1) + \frac{1}{N}(\alpha\chi,\alpha) $ and $\eta_b= \frac{\eta_N+\gamma}{N} = (\mathbf{U}_n,1) + \frac{1}{N}(\alpha\chi,\alpha) + \frac{1}{N}\gamma$. From these definitions, one can see

$$
E[\eta_a] = ( E[\mathbf{Y}], 1) + \frac{1}{N} (\alpha\chi,\alpha)
$$

$$
E[\eta_b] = E[\eta_a] + \frac{1}{N} \gamma
$$

\begin{equation}
 || \eta_a - (E[\mathbf{Y}],1) || \leq || (\mathbf{U}_N,1)  - (E[\mathbf{Y}],1) || + \frac{1}{N}||\alpha\chi|| \label{eq:etaatri}
\end{equation}

\begin{align}
 || \eta_b - (E[\mathbf{Y}],1) || \leq || (\mathbf{U}_N,1)  - (E[\mathbf{Y}],1) || \nonumber\\
 + \frac{1}{N}\big (||\alpha\chi|| + ||\gamma||\big) \label{eq:etabtri} \mbox {.}
\end{align}

From the concentration bound in \eqref{eq:Uconc}, we know $\eta_a$ and $\eta_b$ can be made to lie w.h.p. in a region near their expectations with large $N$, and we wish to show this region satisfies all the regularity assumptions seen in Lemma \ref{lem:covargoesdown}. Lemma \ref{lem:modeexists} states $\theta^*_\eta$ is a continuously differentiable function of $\eta$. Let it be denoted by the function $r(\eta)$. For $\eta_0 = (E[\mathbf{Y}],1)$, we see from equation \eqref{eq:themode} that $r(\eta_0) = \theta_0$.

The preimage  $r^{-1}\big(\mathcal{B}(\theta_0,\delta) \big)$ is an open set, since it is the continuous preimage of an open set. Thus there exists $\delta'$ such that $\mathcal{B}(\eta_0,\delta') \subset r^{-1}(\mathcal{B}(\theta_0,\delta/2))$.

We may now pick $\epsilon \leq \delta'/2$ and let $N'_{\delta',\tau} = \max\big(\frac{2}{\delta'}(||\gamma|| + ||\alpha\chi||),N_{\epsilon,\tau}\big)$. When $n>N'_{\delta',\tau}$, we have $\frac{1}{N}||\alpha\chi|| + \frac{1}{N}||\gamma|| \leq \delta'/2$ and \eqref{eq:Uconc}, \eqref{eq:etaatri},\eqref{eq:etabtri} together show the following:

\begin{equation}
Pr(\eta_a \not \in \mathcal{B}(\eta_0,\delta') \vee \eta_b \not \in \mathcal{B}(\eta_0,\delta')) \leq \tau \mbox{.}
\end{equation}

With high probability, $\eta_a$ and $\eta_b$ both lie in a neighborhood of $\eta_0$. Further, all $\eta$ in this neighborhood have modes $\theta^*_\eta \in \mathcal{B}(\theta_0,\delta)$, a region that assumptions (4) and (5) tell us is well-behaved. The assignment $\delta_1 = \delta'$ and $\delta_2 = \delta/2$ satisfies the conditions for Lemma \ref{lem:covargoesdown} with assumptions (2),(3),(4),(5),(6) serving to round out the rest of the regularity assumptions of Lemma \ref{lem:covargoesdown} with trivial translations.

By the construction, we have $\eta_N = N\eta_a$ and $\eta_N + \gamma = N\eta_b$. For any $\zeta$ on the line segment connecting $\eta_N$ and $\eta_N + \gamma$, we have $\zeta =  N\eta_c$ for some $\eta_c$ on the line segment connecting $\eta_a$ and $\eta_b$.

Therefore by Lemma \ref{lem:covargoesdown}, there exists a $K$ and a $C$ such that if $N > K$ we have $||cov(T(\theta)|\zeta)|| < \frac{C}{N}$. This bound can be used in Lemma \ref{lem:expfamkltaylor} with $D_N = O(1/N)$ to see 

$$
KL(\tilde{p}_N||p_N) = O(1/N)C||\gamma||
$$ 

whenever $N > max(N'_{\delta',\tau},K)$ with arbitrarily high probability $1 - \tau$. Letting $\tau$ approach 0, we can extend this to the expectation over the randomness of $\mathbf{X}$, as with probability 1 our random variables will lie in the region where this inequality holds.

\begin{equation}
\limsup_{N \rightarrow \infty} E_\mathbf{X}[KL(\tilde{p}_N||p_N)] = 0 \label{eq:fixedgammalimit}
\end{equation}

Equation \eqref{eq:fixedgammalimit} is w.r.t. to a fixed $\gamma$, but the desired result is an expectation over $\gamma$ and $\mathbf{X}$. First, let us express this expectation in terms of $\gamma$ and $\mathbf{X}$. Letting $D_N = O(1/N)$ denote the bound used in Lemma \ref{lem:expfamkltaylor} and $N$ being sufficiently large:

\begin{align}
E [ KL(\tilde{p}_N || p_N) ] =  \int E_\mathbf{X}\big[ KL(\tilde{p}_N || p_N) | \gamma\big] dPr(\gamma) \nonumber\\
\leq \int D_N ||\gamma|| dPr(\gamma) \mbox {.}\label{eq:gammaexpect}
\end{align}

The assumption that $\gamma$ comes from a distribution of finite variance ensures the right side of \eqref{eq:gammaexpect} is integrable. By an application of Fatou's Lemma, the following inequality holds:

\begin{align}
\int \limsup_{N \rightarrow \infty} E_\mathbf{X}\big[ KL(\tilde{p}_N || p_N) | \gamma\big] dPr(\gamma) \nonumber\\
\geq \limsup_{N \rightarrow \infty} \int  E_\mathbf{X}\big[ KL(\tilde{p}_N || p_N) | \gamma\big] dPr(\gamma) \mbox{.}
\end{align}

The left hand side has been shown to be zero by equations \eqref{eq:fixedgammalimit} and \eqref{eq:gammaexpect}, and the right hand side is bounded below by 0 since KL divergences are never negative. Thus this inequality suffices to show the limit is zero and prove the desired result.

\end{proof2}

\begin{Cor}
The Laplace mechanism on an exponential family satisfies the noise distribution requirements of 
Lemma \ref{lem:laplacekl} when the sensitivity of the sufficient statistics is finite and either the exponential family is minimal, or if the exponential family parameters $\theta$ are identifiable.
\end{Cor}

\begin{proof2}
If the exponential family is already minimal, this result is trivial. If it is not minimal, there exists a minimal parameterization. We wish to show adding noise to the non-minimal parameters is equivalent to adding differently distributed noise to the minimal parameterization, and this new noise distribution also satisfies the noise distribution requirements of Lemma \ref{lem:laplacekl}: the noise distribution does not depend on $N$ and it has finite covariance.

Let us explicitly construct a minimal parameterization for this family of distributions. If the exponential family is not minimal, this means the $d$ dimensions of the sufficient statistics $S(\mathbf{x})$ of the data are not fully linearly independent. Let $S(x)_j$ be the $j^{th}$ component of $S(\mathbf{x})$ and $k$ be the maximal number of linearly independent sufficient statistics, and without loss of generality assume they are the first $k$ components. Let $\tilde{S}(\mathbf{x})$ be the vector of these $k$ linearly independent components.

For $\forall j > k$, $\forall x \exists \phi_j \in \mathbb{R}^k$ such that $ S(\mathbf{x})_j =  \phi_j \cdot \tilde{S}(\mathbf{x}) + z_j$. We wish to build a minimal exponential family distribution that is identical to the original one, but is parameterized only by $\tilde{S}(\mathbf{x})$ as the sufficient statistics and some $\tilde{\theta}$ as the natural parameters. For these two distributions to be equivalent for all $x$, it suffices to have equality on the exponents.

\begin{equation}
(\theta^\intercal S(\mathbf{x}) - A(\theta) ) = (\tilde{\theta}^\intercal \tilde{S}(\mathbf{x}) - \tilde{A}(\tilde{\theta})) \label{eq:exponentmatching}
\end{equation}

Examining the difference of the two sides, we get

\begin{align}
\theta^\intercal S(x) - \tilde{\theta}^\intercal \tilde{S}(x) - A(\theta) + \tilde{A}(\tilde{\theta}) \nonumber\\
= \sum_{j=1}^k (\theta_j - \tilde{\theta}_j) S(x)_j + \sum_{j=k+1}^d \theta_j S(x)_j - A(\theta) + \tilde{A}(\tilde{\theta})\mbox{.}
\end{align}

Using the known linear dependence for $j > k$, this can be rewritten as

\begin{align}
\sum_{j=1}^k (\theta_j - \tilde{\theta}_j) S(\mathbf{x})_j + \sum_{j=k+1}^d \theta_j (\phi_j \cdot \tilde{S}(\mathbf{x}) + z_j)\nonumber\\
- A(\theta) + \tilde{A}(\tilde{\theta}) \\
= \sum_{j=1}^k (\theta_j - \tilde{\theta}_j) S(\mathbf{x})_j + \sum_{j=k+1}^d \theta_j (\phi_j \cdot \tilde{S}(\mathbf{x}))\nonumber\\
+ \sum_{j=k+1}^d \theta_j z_j - A(\theta) + \tilde{A}(\tilde{\theta}) \mbox{.} \label{eq:minimalitycollapse}
\end{align}

Now since $\tilde{S}(\mathbf{x})$ is merely the first $k$ components of $S(\mathbf{x})$, the first two sums of \eqref{eq:minimalitycollapse} are each simply dot products of $\tilde{S}(\mathbf{x})$ and can be combined as $(\theta_{[k]} - \tilde{\theta} + \sum_{j = k+1}^d \theta_j \phi_j)^\intercal \tilde{S}(\mathbf{x})$ where $\theta_{[k]}$ is the vector of the first $k$ components of $\theta$.  We can force equation \eqref{eq:exponentmatching} to hold by choosing $\tilde{\theta}$ and $\tilde{A}(\tilde{\theta})$ appropriately to set equation \eqref{eq:minimalitycollapse} to zero.

\begin{itemize}
\item $\tilde{\theta}= \theta_{[k]} + \sum_{j = k+1}^d \theta_j \phi_j$ 
\item $\tilde{A}(\tilde{\theta}) = -\sum_{j=k+1}^d \theta_j z_j + A(\theta)$
\end{itemize}

We note that this requires $\tilde{A}(\tilde{\theta})$ to truly be a function depending only on $\tilde{\theta}$, but we have written it in terms of $\theta$ instead. This is justifiable by the assumption that the natural parameters $\theta$ are identifiable, that is each distribution over $\mathbf{x}$ is associated with just one $\theta \in \Theta$. This means there is a bijection from $\theta$ and $\tilde{\theta}$, which ensures $\tilde{A}(\tilde{\theta})$ is a well-defined function.

This suffices to characterize the way the additional natural parameters affect the parameters of the equivalent minimal system. Any additive noise to a component $\theta_j$ translates linearly to additive noise on the components $\tilde{\theta}_j$, meaning the Laplace mechanism's noise distribution on the non-minimal parameter space still corresponds to some noise distribution on the minimal parameters that does not depend on the data size $N$, and it still has a finite covariance. If the minimal exponential family tends towards a KL divergence of zero, the equivalent non-minimal exponential family must as well.
\end{proof2}

\begin{theorem}
\label{thm:supplaplaceare}
Under the assumptions of Lemma \ref{lem:laplacekl}, the Laplace mechanism has an asymptotic posterior of $\mathcal{N}(\theta_0,2\mathbb{I}^{-1}/N)$ from which drawing a single sample has an asymptotic relative efficiency of 2 in estimating $\theta_0$, where $\mathbb{I}$ is the Fisher information at $\theta_0$.
\end{theorem}

\begin{proof2}

The assumptions of Lemma \ref{lem:laplacekl} match the Laplace regularity assumptions under which asymptotic normality holds, and we know that the unperturbed posterior $p_N$ converges to $\mathcal{N}(\theta^*,2\mathbb{I}^{-1}/N)$ under the Bernstein-von Mises theorem \citep{geisser1990validity}. If $\tilde{p}_N$ is the posterior of the Laplace mechanism for a fixed randomness, then we have $\lim_{N \rightarrow \infty} KL(\tilde{p}_N || p_N) = 0$ and $\tilde{p}_N$ must converge to the same distribution as $p_N$. From this it is clear that samples from $p_N$ and from $\tilde{p}_N$ both have an asymptotic relative efficiency of 2. We once again argue that if this asymptotic behavior holds for any fixed randomness of the Laplace mechanism, it also holds for the Laplace mechanism as a whole.
\end{proof2}

To show the previous results, we relied on some mathematical results involving the covariances of posteriors after observing a large amount of data. We still need to show these bounds on the covariances, which will be accomplished by adapting existing Laplace approximation methods. Before we get there, we will need one quick result about convex functions with a positive definite Hessian in order to perform the approximation:

\begin{Lem}
\label{lem:nearextremefiniteregion}

Let $f(y) : \mathbb{R}^d \rightarrow \mathbb{R}$ be a strictly convex function with minimum at $y^*$. If $\nabla^2 f(y^*)$ is positive definite and $\nabla^3 f(y)$ exists everywhere, then for any $c>0$ there exists $b>0$ such that $|f(y) - f(y^*)| \leq b$ implies $||y - y^*|| \leq c$.

\end{Lem}

\begin{proof2}

By the existence of $\nabla^3 f(y)$ and thus the continuity of $\nabla^2 f(y)$, we know there exists a positive $\delta<c$ and a $w>0$ such that $y \in B(y^*,\delta)$ implies $\nabla^2 f(y)  - w\mathbb{I}$ is positive semi-definite, where $\mathbb{I}$ is the identity matrix. (i.e. the spectral norm $||\nabla^2 f(y)|| \geq w$)

As $y^*$ is the global minimum, we know the gradient is 0 at $y^*$. Thus for $y \in B(y^*,\delta)$ this leads to a Taylor expansion of the form

\begin{align}
\label{eq:convexbound}
f(y) &= f(y^*) + (y-y^*)\frac{1}{2}\nabla^2f(y')(y-y^*)^\intercal \nonumber\\
&\geq f(y^*) + \frac{w}{2} ||y- y^*||
\end{align}

for some $y'$ on the line segment connecting $y$ and $y^*$. The inequality follows from the second derivative being positive definite on this neighborhood.

Consider the set $Q_\epsilon = \{ y \text{ s.t. } ||y - y^*|| = \epsilon \}$. By equation \eqref{eq:convexbound} we know for $y \in Q_\epsilon$ we have $|f(y) - f(y^*)| \geq \frac{w\epsilon}{2}$ if $\epsilon \leq \delta$.

For any $y \not \in B(y^*,\delta)$, there exists $t \in (0,1)$ such that $(1-t)y^* + ty \in Q_\delta$ by the continuity of the norm.

By strict convexity, we know

$$
tf(y) + (1-t)f(y^*) > f(ty + (1-t)y^*)
$$

$$
f(y) > \frac{1}{t}f(ty + (1-t)y^*) + \frac{t-1}{t}f(y^*)
$$

$$
f(y) -f(y^*) > \frac{1}{t}f(ty + (1-t)y^*) - \frac{1}{t} f(y^*)\mbox{.}
$$

If we let $t$ satisfy $(1-t)y^* + ty \in Q_\delta$ we know $t = \delta/||y-y^*|| \leq 1$. Substituting with \eqref{eq:convexbound} we get

\begin{align*}
f(y)-f(y^*) > \frac{(w/2)\delta + f(y^*)}{t} - \frac{1}{t} f(y^*) = \frac{w\delta}{2t} \geq \frac{w\delta}{2} \mbox{.}
\end{align*}

Thus if we let $b = \frac{w\delta}{2}$, we see $||y-y^*|| > c$ implies $|f(y) - f(y^*)| > b$.

The desired result then follows as the contrapositive.

\end{proof2}

Lemma \ref{lem:nearextremefiniteregion} will be used to demonstrate a regularity assumption required in the next lemma, which performs all the heavy lifting in using the Laplace approximation. Lemma \ref{lem:laplaceapprox} adapts a previous argument about Laplace approximations of a posterior. This adapted Laplace approximation argument forms the core of Lemma \ref{lem:covargoesdown}, which allows us to see the covariance of posteriors shrink as more data is observed.

\begin{Lem}
\label{lem:laplaceapprox}

Let $s(\phi,Y)$ be a function $M \times U \rightarrow \mathbb{R}$, where $M$ is the space of $\phi$ and $U$ is the space of $Y$.

For functions of the form $F_k(Y) = \int_{\phi \in M} e^{k s(\phi,Y)} d\phi$, if the following regularity assumptions hold for some $\delta_1 > 0$, $\delta_2>0$, $Y_0 \in M$:

\begin{enumerate}
\item $\phi^*_Y = \argmax_{\phi \in M} s(\phi,Y) = g(Y)$, a function of $Y$
\item $\phi^*_{Y'}$ is in the interior of $M$ for all $Y' \in \mathcal{B}(Y_0,\delta_1)$
\item $g(Y)$ is continuously differentiable over the neighborhood $\mathcal{B}(Y_0,\delta_1)$ \label{cond:approx_lipschitz}
\item $s(\phi,Y')$ has derivatives of all orders for $Y' \in \mathcal{B}(Y_0,\delta_1)$,$\phi \in \mathcal{B}(\phi^*_{Y'},\delta_2)$ and all partial derivatives up to order 7 are bounded by some constant $P$ on this neighborhood \label{cond:approx_deriv}
\item $\exists w>0$ such that $\forall Y' \in \mathcal{B}(Y_0,\delta_1),\forall \phi \in \mathcal{B}(\phi^*_{Y'},\delta_2)$ we have $det(\nabla^2_\phi s(\phi,Y)) > w$ \label{cond:approx_eigen}
\item $F_1(Y')$ exists for $Y' \in \mathcal{B}(Y_0,\delta_1)$, the integral is finite
\end{enumerate}

then there exists $C$ and $K$ such that for any $k >K$ and any $Y' \in \mathcal{B}(Y_0,\delta_1)$,  letting $\psi = kY'$,  the spectral norm $||\nabla^2_\psi \log F_k(\psi/k)|| < \frac{C}{k}$.
\end{Lem}

\begin{proof2}

Our goal here is to bound $||\nabla^2_\psi \log F_k(\psi/k)||$, which we will achieve by characterizing $F_k(\psi/k)$ and analyzing its derivatives.

We will be using standard Laplace approximation methods seen in \citep{geisser1990validity} to explore $F_k(\psi)$. To begin, we must show our assumptions satisfy the regularity assumptions for the approximation.

For a fixed $Y' \in B(Y_0,\delta)$, from condition \ref{cond:approx_eigen} we know there exists a neighborhood around $\phi^*_Y$ where $\nabla^2_\phi s(\phi,Y)$ is positive definite. For $\delta'>0$, let $Q_{\delta',Y} = \{ \phi \in M \text{ s.t. } ||\phi - \phi^*_Y|| \leq \delta'\}$. By using Lemma \ref{lem:nearextremefiniteregion} we can verify the following expression for any $\delta' \in (0,\delta)$:

\begin{equation}
\limsup_{k \rightarrow \infty} \sup_{\phi\not\in Q_{\delta',Y}}  s(\phi,Y) - s(\phi^*_Y,Y) < 0 \mbox{.} \label{cond:limsup}
\end{equation}

Note that the right hand side does not depend on $k$, and Lemma \ref{lem:nearextremefiniteregion} guarantees a non-zero bound for the right hand side for any $\delta' \in (0,\delta)$.  Equation \eqref{cond:limsup} exactly matches condition $(iii)'$ of Kass, and its intuitive meaning is that for any $\delta'$, there exists sufficiently large $k$ such that the integral $F_k$ is negligible outside the region $Q_{\delta'}$.

Conditions (4),(5),(6) also match directly the conditions given by Kass, though we note we require even higher derivatives to be bounded or present. These extra derivatives will be used later to extend the argument given by Kass to suit our purposes and give a uniform bound across a neighborhood.

Theorem 1 of \citep{geisser1990validity} gives the following result, when we set their $b$ to the constant 1:

\begin{equation}
 F_k(Y) = (2\pi)^\frac{m}{2}[\det(k\nabla^2s(\phi^*_Y,Y)]^{-\frac{1}{2}} \exp(-k s(\phi^*_Y,Y)) Z(kY) \label{eq:lapprox}
\end{equation}

\begin{align}
Z(kY) &= 1 + \frac{1}{k} \Big( \nonumber\\
&\frac{1}{72} \sum (\nabla^3_\phi s(\phi^*_Y,Y))_{(pqr)} (\nabla^3s(\phi^*_Y,Y))_{(def)}\mu^6_{pqrdef} \nonumber\\
&- \frac{1}{24} \sum (\nabla^4s(\phi^*_Y,Y))_{(defg)}\mu^4_{defg} \Big) + O(k^{-2})\mbox{,}
\end{align}

where $m$ is the dimensionality of $Y$, $\mu^6_{pqrdef}$ and $\mu^4_{defg}$ are the sixth and fourth central moments of a multivariate Gaussian with covariance matrix $(\nabla^2s(\phi^*_Y,Y))^{-1}$.  All sums are written in the Einstein summation notation. We remark that the $O(k^{-2})$ error term of this approximation also depends on $kY$.

What we are really interested in is the quantity $\nabla^2_\psi \log F_k(\psi)$ evaluated at $\psi = kY$. We take the logarithm of \eqref{eq:lapprox}:

\begin{align}
\log F_k(\psi/k) =& \log \Big ( (2\pi)^\frac{m}{2}[\det(k\nabla^2s(\phi^*_Y,Y)]^{-\frac{1}{2}} \nonumber\\
&\cdot \exp(-k s(\phi^*_Y,Y)) Z(\psi) \Big ) \nonumber\\
=& \log \big ( (2\pi)^\frac{m}{2} \big ) - \frac{1}{2}\log \big ( [\det(k\nabla^2s(\phi^*_Y,Y))] \big ) \nonumber\\
&- k s(\phi^*_Y,Y) + \log (Z(\psi)) \mbox{.} \label{eq:lapproxlog}
\end{align}

We define new functions $\tilde{s_0}, \tilde{s_1}, \tilde{s_2}$ to simplify the analysis.

\begin{equation}
\tilde{s_0}(Y) = s(\phi^*_Y,Y) = s(g(Y),Y)
\label{eq:stilde}
\end{equation}

\begin{equation}
\tilde{s_1}(Y) = \nabla_\phi s(\phi^*_Y,Y) = \nabla_\phi s(g(Y),Y)
\label{eq:stildegrad}
\end{equation}

\begin{equation}
\tilde{s_2}(Y) = \nabla^2_\phi s(\phi^*_Y,Y) = \nabla^2_\phi s(g(Y),Y)
\label{eq:stildehess}
\end{equation}

By assumptions \eqref{cond:approx_lipschitz} and \eqref{cond:approx_deriv} we know these functions are continuously differentiable on $\overline{\mathcal{B}(Y_0,\delta_1)}$ as they are the composition of continuously differentiable functions on the compact set $\overline{\mathcal{B}(Y_0,\delta_1)}$.

We next look at the first derivative of \eqref{eq:lapproxlog}.  We remark that the partial derivatives  of $\log \det(X)$ are given by $X^{-\intercal}$.

\begin{align}
\nabla_\psi  \log F_k(\psi/k) =& \nabla_\psi[- \frac{1}{2}\log \big ( [\det(k\tilde{s_2}(\psi/k)] \big )]   \nonumber\\
&- \nabla_\psi[k \tilde{s_0}(\psi/k)] + \nabla_\psi \log (Z(\psi)) \nonumber\\
=& -\frac{1}{2}(k\tilde{s_2}(\psi/k)))^{-\intercal}\frac{1}{k} \nonumber\\
&+ \tilde{s_1}(\psi/k)  + \frac{\nabla_\psi (Z(\psi))}{Z(\psi)} \label{eq:laplace_grad}
\end{align}

Now that we have an expression for $\nabla_\psi  \log F_k(\psi/k)$ , we take yet another derivative w.r.t. to $\psi$ to get our desired $\nabla^2_\psi$.

\begin{align}
\nabla^2_\psi  \log F_k(\psi/k) =&  \nabla_\psi [-\frac{1}{2}(k\tilde{s_2}(\psi/k)))^{-\intercal}\frac{1}{k} ] \nonumber\\
&{}+ \nabla_\psi [\tilde{s_1}(\psi/k)] \nonumber\\
&{}+  \nabla_\psi [\frac{\nabla_\psi (Z(\psi))}{Z(\psi)}] \label{eq:laplace_hess}
\end{align}

Let us consider each of the three terms on the right side of \eqref{eq:laplace_hess} in isolation. For the first term, we introduce yet another function $\tilde{s}_{-2}(Y)$, the composition of $\tilde{s_2}$ with the matrix inversion.

$$
\tilde{s}_{-2}(Y) = (\tilde{s_2}(Y))^{-1}
$$

With this new function in hand, we further condense the first term of \eqref{eq:laplace_hess}.

\begin{align}
\nabla_\psi [-\frac{1}{2}(k\tilde{s_2}(\psi/k)))^{-\intercal}\frac{1}{k} ] =& \nabla_\psi [-\frac{1}{2k}(\tilde{s}_{-2}(\psi/k)))\frac{1}{k} ] \nonumber\\
=& -\frac{1}{2k^3} \nabla_Y \tilde{s}_{-2}(\psi/k) \nonumber\\
=& O(k^{-3}) \label{eq:laplace_term1}
\end{align}

We previously remarked that $\tilde{s_2}$ is continuously differentiable on the compact set $\overline{\mathcal{B}(Y_0,\delta_1)}$. Condition \eqref{cond:approx_eigen} informs us that $\tilde{s_2}(Y)$ is bounded away from being a singular matrix on $\overline{\mathcal{B}(Y_0,\delta_1)}$ , so the matrix inversion is also uniformly continuous on this compact set. This means $\nabla_Y \tilde{s}_{-2}(\psi/k)$ has a finite supremum over  $\overline{\mathcal{B}(Y_0,\delta_1)}$ and thus we can say this term is $O(k^{-3})$ uniformly on this neighborhood.

Next we consider the second term of \eqref{eq:laplace_hess}.

\begin{equation}
\nabla_\psi [\tilde{s_1}(\psi/k)] = \frac{1}{k}\tilde{s_2}(\psi/k) = O(k^{-1}) \label{eq:laplace_term2}
\end{equation}

From the continuity of $\tilde{s_2}(\psi/k)$ on our compact neighborhood, we know $\tilde{s_2}(Y)$ has a finite supremum 
over the compact set $\overline{\mathcal{B}(Y_0,\delta_1)}$, which gives the uniform $O(k^{-1})$ bound.

Finally, we must consider the third term of \eqref{eq:laplace_hess}.

\begin{equation}
\nabla_\psi [\frac{\nabla_\psi (Z(\psi))}{Z(\psi)}] =  \frac{\nabla^2 (Z(\psi))}{Z(\psi)} - \frac{\nabla (Z(\psi))(\nabla (Z(\psi)))^\intercal}{Z(\psi)^2} 
\end{equation}

Recall that $Z(\psi)$ had a local $O(k^{-2})$ error term as given by \citep{geisser1990validity}. We wish to bound the derivatives of $\log F_k(\psi)$, but the local bound on this error term given by Kass does not bound its derivatives. However, a slight modification of the argument of \citep{geisser1990validity} shows that our added assumptions about the higher order derivatives are sufficient to control the behavior of this error term. The following expression is their equation (2.2), translated to our setting:

\begin{align}
 \exp(-k s(\phi,Y)) = & &\nonumber\\
 \exp(-k s(\phi^*_Y,Y) )\exp(\frac{1}{2}\nabla^2s(\phi^*_Y,Y)u^2) W(\phi,Y) \\
W(\phi,Y) = 1 - \frac{1}{6}k^{-1/2}\nabla^3s(\phi^*_Y,Y)u^3 & &\nonumber\\
+ \frac{1}{72}k^{-1}(\nabla^3s(\phi^*_Y,Y))^2u^6 \nonumber\\
- \frac{1}{24}k^{-1}\nabla^4s(\phi^*_Y,Y)u^4 \nonumber\\
-\frac{1}{120}k^{-3/2}\nabla^5s(\phi^*_Y,Y)u^5 \nonumber\\
+ \frac{1}{72}k^{-3/2}\nabla^3A(s(\phi^*_Y,Y))\nabla^4s(\phi^*_Y,Y)u^7 \nonumber\\
+ G(\phi,\phi^*_Y,Y)\mbox{,}
\end{align}

where $G(\phi,\phi^*_Y,Y)$ is the fifth-order Taylor expansion error term (i.e. it depends on the sixth-order partial derivatives at some $\phi'$ between $\phi$ and $\phi^*_Y$).

We may continue this Taylor expansion another degree further to bound the variation of $G(\phi,\phi^*_Y,Y)$ for $\phi \in \mathcal{B}(\phi^*_Y,\delta_2)$. We will consider $Z(\psi)$, $\nabla_\psi Z(\psi)$, and $\nabla_\psi^2 Z(\psi)$ as three separate functions, each permitting a higher order Taylor expansion. Each will have their own respective error term depending on the seventh-order partial derivatives at some $\phi'$, but we note that $\phi'$ is not necessarily the same for each of them.

The argument of  \citep{geisser1990validity} already shows how the terms composing their $O(k^{-2})$ error term can be bounded in terms of $\nabla^6_\phi S(\phi^*_Y,Y)$. It is trivial to show an analogous result for our higher order approximations. This allows us to extend our approximation of $Z(\psi)$ and its derivatives uniformly to the neighborhood $\mathcal{B}(\phi^*_Y,\delta_2)$. The newly introduced extra approximation terms are $O(k^{-v})$ with $v \geq 2$, and so our uniform bounds are still simply $O(k^{-2})$, though with a larger constant now.

Let $k$ be sufficiently large, and let $Q,R,S$ be positive constants satisfying $0 < Q < ||Z(\psi)||$, $R > k||\nabla_\psi Z(\psi)||$, $S > k||\nabla^2_\psi Z(\psi)||$ for all $\psi$ in $\{\psi | \psi/k \in B(Y_0,\delta)$. We remark that $Q$ exists by virtue of $Z = 1 + O(k^{-1}) + O(k^{-2})$. $R$ and $S$ similarly exist by $||\nabla_\psi Z(\psi)||$ and  $||\nabla^2_\psi Z(\psi)||$ both being $O(k^{-1})$ with no constant term in front.

$$
\nabla_\psi [\frac{\nabla_\psi (Z(\psi))}{Z(\psi)}] \leq  \frac{S}{kQ} - \frac{R^2}{k^2Q^2} \text{for all } Y' \in B(Y_0,\delta)
$$

This right hand side is clearly $O(k^{-1})$, and we have uniform bounds across our neighborhood.

\begin{equation}
\nabla_\psi [\frac{\nabla_\psi (Z(\psi))}{Z(\psi)}] = O(k^{-1}) \label{eq:laplace_term3}
\end{equation}

Combining the results of \eqref{eq:laplace_term1},  \eqref{eq:laplace_term2},  \eqref{eq:laplace_term3} with their sum in \eqref{eq:laplace_hess}, we get this result:

\begin{equation}
||\nabla^2_\psi  \log F_k(\psi/k)|| = O(k^{-1})\mbox{.}
\end{equation}

This uniform asymptotic bound then ensures we have the intended result: $\exists C,K$ such that $\forall Y \in \mathcal{B}(Y_0,\delta_1)$ when $k>K$ and $\psi = kY$ we have $||\nabla^2_\psi  \log F_k(\psi/k)|| \leq C/k$

\end{proof2}